\documentclass[acmtog, authorversion]{acmart}
\pdfoutput=1

\usepackage{graphicx}
\usepackage{amsmath}
\usepackage{multirow}
\usepackage{booktabs}
\usepackage{lipsum}
\usepackage{colortbl}

\newcommand{\highlightBest}[1]{\colorbox[HTML]{FD9C9B}{#1}}
\newcommand{\bestCellColor}[1]{\cellcolor[HTML]{FD9C9B}#1}

\newcommand{\highlightSecondBest}[1]{\colorbox[HTML]{FEC998}{#1}}
\newcommand{\secondBestCellColor}[1]{\cellcolor[HTML]{FEC998}#1}

\AtBeginDocument{%
  \providecommand\BibTeX{{%
    \normalfont B\kern-0.5em{\scshape i\kern-0.25em b}\kern-0.8em\TeX}}}

\setcopyright{acmlicensed}
\acmJournal{TOG}
\acmYear{2023} \acmVolume{42} \acmNumber{4} \acmArticle{1} \acmMonth{8} \acmPrice{15.00}\acmDOI{10.1145/3592415}

\newcommand{\methodname}{HumanRF}
\newcommand{\datasetname}{\mbox{ActorsHQ}}
\newcommand{\OURS}{\methodname}

\newlength{\mrgone}

\graphicspath{{figures/}}

\acmSubmissionID{560}

\newif\ifArxivVersion

\newcommand{\ray}{\boldsymbol{r}}
\newcommand{\colorGt}{C}
\newcommand{\colorPred}{\hat{C}}
\newcommand{\raySpace}{\mathcal{R}}
\newcommand{\maskGt}{M}
\newcommand{\occGrid}{O}
\newcommand{\maskPred}{\hat{M}}

\newcommand{\frameIndex}{t}
\newcommand{\keyframeIndex}{k}
\newcommand{\keyframeSize}{N}
\newcommand{\loss}{\mathcal{L}}
\newcommand{\bceLossWeight}{\beta}

\newcommand{\volrenDistance}{\alpha}
\newcommand{\volrenDensity}{\sigma}
\newcommand{\volrenRadiance}{L}

\newcommand{\occupancy}{\delta}
\newcommand{\expansion}{\phi}

\newcommand{\keyframeTimeSpace}{\mathcal{T}}
\newcommand{\spatialSpace}{\mathcal{P}}

\newcommand{\reprPoint}{\mathbf{p}}
\newcommand{\reprTime}{t}
\newcommand{\reprDirection}{\mathbf{d}}
\newcommand{\featureTensor}{T}
\newcommand{\featureDim}{m}
\newcommand{\geometryFeatures}{F}

\newcommand{\R}{\mathbb R}
\DeclareMathOperator{\MLP}{MLP}
\DeclareMathOperator{\SH}{SH}
\newcommand{\densityMLP}{\MLP_{\volrenDensity}}
\newcommand{\radianceMLP}{\MLP_{\volrenRadiance}}

\begin{document}

\title{\OURS: High-Fidelity Neural Radiance Fields for Humans in Motion}

\author{Mustafa Işık}
\email{mustafa.isik@synthesia.io}
\affiliation{
  \institution{Synthesia}
  \city{Munich}
  \country{Germany}
}
\author{Martin Rünz}
\email{martin@synthesia.io}
\affiliation{
  \institution{Synthesia}
  \city{Munich}
  \country{Germany}
}
\author{Markos Georgopoulos}
\email{markos@synthesia.io}
\affiliation{
  \institution{Synthesia}
  \city{London}
  \country{United Kingdom}
}
\author{Taras Khakhulin}
\email{taras.khakhulin@synthesia.io}
\affiliation{
  \institution{Synthesia}
  \city{London}
  \country{United Kingdom}
}
\author{Jonathan Starck}
\email{jon@synthesia.io}
\affiliation{
  \institution{Synthesia}
  \city{London}
  \country{United Kingdom}
}
\author{Lourdes Agapito}
\email{l.agapito@cs.ucl.ac.uk}
\affiliation{
  \institution{University College London}
  \city{London}
  \country{United Kingdom}
}
\author{Matthias Nießner}
\email{niessner@tum.de}
\affiliation{
  \institution{Technical University of Munich}
  \city{Munich}
  \country{Germany}
}

\renewcommand{\shortauthors}{Işık et al.}

\begin{abstract}
Representing human performance at high-fidelity is an essential building block in diverse applications, such as film production, computer games or videoconferencing. 
To close the gap to production-level quality, we introduce \methodname{}\footnote{Project website: \href{https://synthesiaresearch.github.io/humanrf}{\color{magenta}{synthesiaresearch.github.io/humanrf}}}, a 4D dynamic neural scene representation that captures full-body appearance in motion from multi-view video input, and enables playback from novel, unseen viewpoints. 
Our novel representation acts as a dynamic video encoding that captures fine details at high compression rates by factorizing space-time into a temporal matrix-vector decomposition.
This allows us to obtain temporally coherent reconstructions of human actors for long sequences, while representing high-resolution details even in the context of challenging motion.
While most research focuses on synthesizing at resolutions of 4MP or lower, we address the challenge of operating at 12MP. 
To this end, we introduce \datasetname{}, a novel multi-view dataset that provides 12MP footage from 160 cameras for 16 sequences with high-fidelity, per-frame mesh reconstructions\footnote{\datasetname{} dataset is publicly available under \href{https://www.actors-hq.com}{\color{magenta}{www.actors-hq.com}} including all raw RGB frames and per-frame reconstructed 3D meshes.}.
We demonstrate challenges that emerge from using such high-resolution data and show that our newly introduced \methodname{} effectively leverages this data, making a significant step towards production-level quality novel view synthesis.
\end{abstract}



\begin{teaserfigure}
  \includegraphics[width=\textwidth]{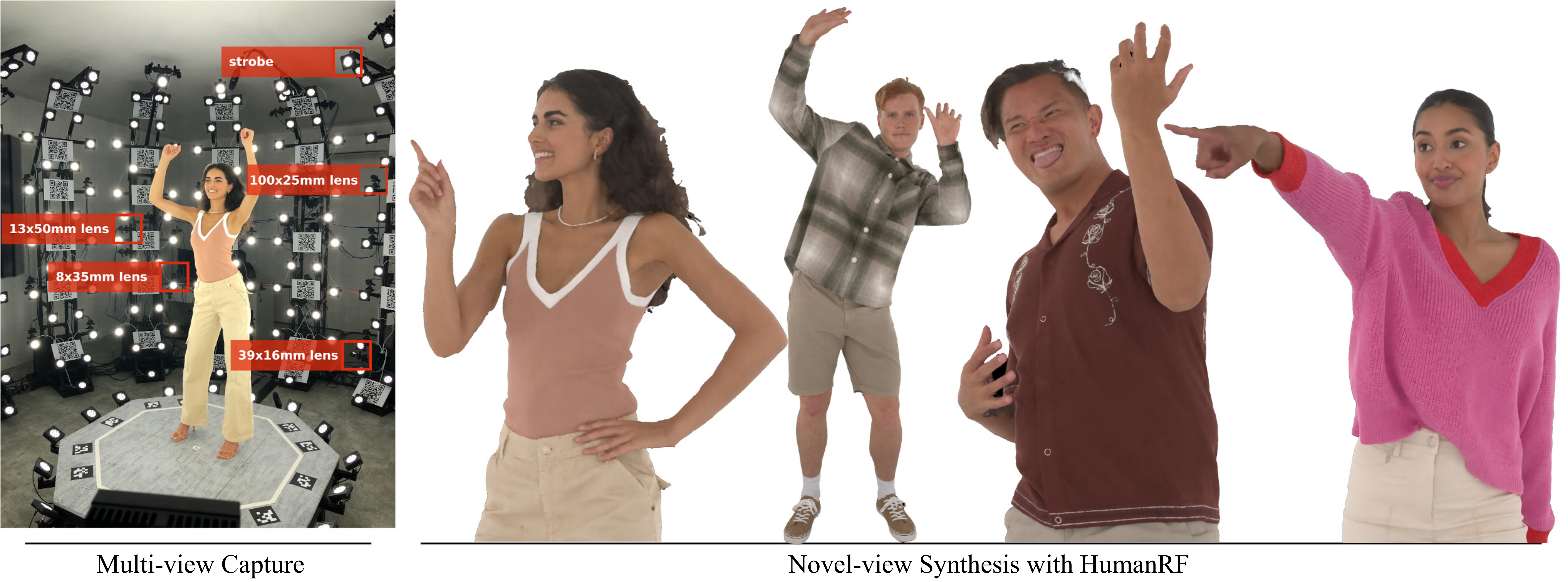}
  \caption{We introduce a novel multi-view dataset of humans in motion captured with a rig of 160 cameras, recording footage of 12MP each (left image). From an input multi-view recording of a specific person, our \OURS{} method reconstructs a spatio-temporal radiance field which captures appearance and motion of the actor. From this representation, we can then synthesize highly-realistic images from unseen, novel view points (right images).}
  \label{fig:teaser}
\end{teaserfigure}

\maketitle

\section{Introduction}

Photo-realistic image synthesis of virtual environments has been one of the core challenges in computer graphics research for decades.
Traditionally, the underlying 3D assets have been created by artists with heavy manual labor; however, recently, significant effort has been devoted to reconstructing the 3D representations from real-world observations.
In particular, novel view synthesis of recorded humans has been the center of attention in numerous  applications, ranging from movie and game production to immersive telepresence.

Yet, reconstructing photo-realistic digital humans from real-world captures involves significant technical challenges.
The diverse granularity of fine-scale detail -- e.g., on faces, hair, clothing -- makes the reconstruction difficult to scale, while the margin for error is low due to the acute ability of the human visual system to perceive even the smallest inconsistencies in synthesized images.
From a methodological standpoint, the main challenge lies in jointly reconstructing appearance and motion in realistic settings due to the large number of degrees of freedom that needs to be encoded.
In particular, modeling fast and complex motions while obtaining photo-realistic results at a sufficient resolution remains an open problem in production.

In recent years, we have seen tremendous progress in addressing these challenges.
More specifically, \citet{mildenhall2020nerf} reconstructs a 3D neural radiance field (NeRF) constrained by a multi-view volumetric rendering loss.
The resulting 3D field is encoded in a multi-layer perceptron (MLP) which then enables novel-view synthesis.
While NeRF originally focused on static scenes, recent works handle dynamic scenes implicitly via time conditioning~\cite{li2022neural} or explicitly via deformation fields~\cite{park2021nerfies,park2021hypernerf}.
These dynamic methods show impressive results; but, they still struggle to handle longer sequences with complex motion -- especially for humans.
In the mean time, obtaining high-quality output renderings requires high-resolution training data which is both difficult to capture and utilize in the subsequent radiance field reconstructions.

In this work, we propose to address these shortcomings of dynamic NeRF methods in the context of capturing moving humans.
Therefore, we first introduce \datasetname{}, a new high-fidelity dataset of clothed humans in motion tailored for photo-realistic novel view synthesis.
The dataset features multi-view recordings of 160 synchronized cameras that simultaneously capture individual video streams of 12MP each, as illustrated in Fig.~\ref{fig:teaser}.
Leveraging our newly captured data, we propose a new scene representation that lifts Instant-NGP~\cite{mueller2022instant} hash encodings to the temporal domain by incorporating the time dimension in conjunction with a low-rank space-time tensor decomposition of the feature grid. 
We further split a sequence into segments, which allows representing very long sequences as only few of the segments need to reside in GPU memory during a training iteration -- something existing methods struggle with due to using a single representation for an entire sequence.
Finally, we demonstrate the effectiveness of our new representation on our newly introduced dataset where we significantly improve over existing state-of-the-art methods. Concretely, our contributions are as follows:
\begin{itemize}
    \item We propose a new spatio-temporal decomposition that can efficiently reconstruct a dynamic radiance field representation from multi-view inputs, based on a low-rank decomposition.
    \item Additionally, we introduce an adaptive splitting scheme which divides a sequence into segments allowing us to capture arbitrarily long sequences.
    \item We further introduce \datasetname, a high-fidelity dataset, featuring footage of 8 actors from 160 cameras that record at a resolution of 12MP each.
\end{itemize}

\begin{figure*}
    \includegraphics[width=\textwidth]{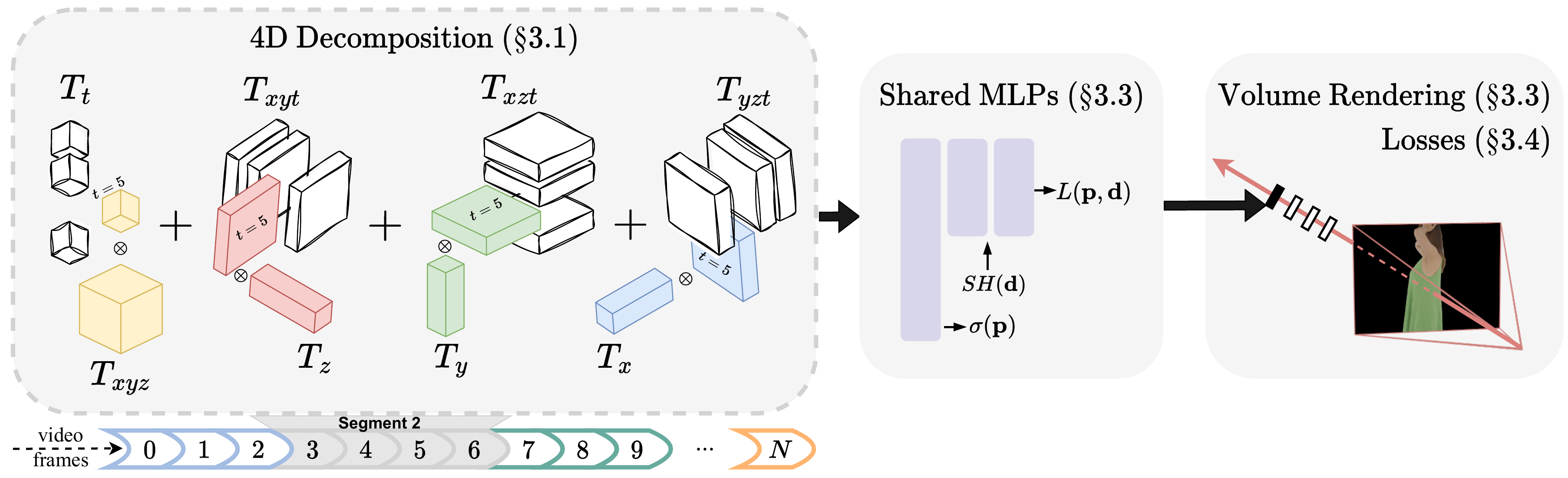}
    \caption{\textbf{Overview of \methodname{}}: 
    Prior to training, our method starts by splitting the temporal domain into 4D segments with similar union occupancy in 3D (\S\ref{sec:adaptive_temporal_partitioning}). Each segment is modeled by a 4D feature grid which is compactly represented by utilizing tensor decomposition and hash grids (\S\ref{sec:feature_grid_decomposition}). During training, we sample a batch of rays across different time frames and cameras. After each pixel color is predicted via volume rendering (\S\ref{sec:shared_mlps_and_rendering}), we enforce photometric constraints and regularize ray marching weights via foreground masks (\S\ref{sec:losses}).
    }
    \label{fig:method}
\end{figure*}

\section{Related Work}
\methodname{} leverages hybrid implicit and volumetric representations to reconstruct free-viewpoint videos. 
In this section, we discuss related work on neural representations for static and dynamic scenes and for human performance capture.

\subsection{3D Neural Representations}
3D reconstruction is a long-standing problem that has been redefined with the advent of deep learning-based approaches.
In particular, coordinate-based networks have become a popular choice for implicit 3D scene representations such as radiance~\cite{mildenhall2020nerf}, signed distance~\cite{park2019deepsdf}, or occupancy~\cite{mescheder2019occupancy} fields. 
In the pioneering work of~\citet{mildenhall2020nerf}, an MLP is trained to encode a radiance field reconstructed from a set of input RGB images. 
Alternatively, some methods utilize explicit data structures, such as sparse grids~\cite{fridovich2022plenoxels}, to achieve fast training and inference at the expense of a larger memory footprint.
TensoRF~\cite{tensorf} addresses memory inefficiencies by using a low-rank tensor decomposition while~\citet{mueller2022instant} propose using hash data structures accompanied with small MLPs.
We elevate the ideas from TensoRF into spatio-temporal domain by representing the feature grids via 4D decomposition using four 3D hash grids and four 1D dense grids.

\subsection{4D Dynamic Representations} 

The creation of free-viewpoint videos has been widely studied due to its numerous applications. 
The seminal work of~\citet{kanade1997virtualized} allows the reconstruction of shapes and textures using a multi-camera dome. 
Similarly, later efforts \cite{Carranza2003FreeviewpointVO,starck2007surface} leveraged multiple cameras for free-viewpoint human rendering. 
More recently, the breakthrough work of~\citet{Collet2015HighqualitySF} proposes to track textured meshes in order to create streamable 3D videos.
~\citet{broxton2020immersive} presents a layered mesh representation to reconstruct and compress video from a multi-camera rig.
More recently, deep learning-based approaches have been proposed for deformable 3D scenes. 
Neural Volumes~\cite{neural_volumes} use an encoder-decoder architecture to optimize a 3D volume from 2D images. 
Similarly, instead of decoding a 3D volume,~\citet{lombardi2021mixture} propose to decode a mixture of volumetric primitives that are attached to a guide mesh. 

A plethora of efforts has been dedicated to extending the success of NeRF into the temporal domain using implicit representations. 
~\citet{li2022neural} extends NeRF with time-conditioning and introduces a keyframe-based training strategy.
Alternatively,~\citet{pumarola2021d, park2021nerfies, park2021hypernerf} introduce a separate MLP to predict scene deformations for multi-view and monocular videos, respectively. In similar vain,~\citet{li2021neural} leverages 2D flow supervision to model a dynamic scene. Similarly to static scenes, the slow convergence of such methods has been addressed using explicit~\cite{liu2022devrf} and hybrid~\cite{fang2022fast,guo2022neural} representations. Orthogonal to these approaches,~\citet{wang2022fourier} fuses a set of static PlenOctrees~\cite{yu2021plenoctrees} into a dynamic representation using DFT to achieve real-time inference.

Concurrently to our work, ~\citet{song2022nerfplayer} decompose the 4D space into static, deforming and newly appeared regions, while~\citet{shao2022tensor4d,hexplanes,kplanes} also propose to represent 4D scenes using low-rank decompositions with 2D tensors.
Unlike these methods, our method uses 3D and 1D tensors. In \S\ref{sec:ablation_feature_grid}, we show our 3D-1D scheme is significantly better than its 2D-2D counterpart for rapid motions. Additionally, our method partitions a sequence into segments, which enables training at scale on modern GPUs without sacrificing quality.

\subsection{Neural Human Performance Capture}
Our goal is closely related to neural radiance field-based methods that specialize in rendering humans. 
This is often achieved by learning a canonical representation that is forward warped to a target frame~\cite{chen2021snarf,wang2022arah} or backward sampled from the observation space~\cite{liu2021neural,xu2022surface}. 
This deformation can be guided by a learned template model such as SMPL~\cite{Loper2015SMPLAS} or a sparse skeleton.
For instance, utilizing sparse landmarks,~\citet{su2021nerf, noguchi2021neural} reparametrize radiance fields relative to the pose of the skeleton. 
TAVA~\cite{li2022tava} optimizes a canonical shape and forward-skinning weights based on skeleton pose.
Neural Body~\cite{peng2021neural} employs a SMPL model to optimize a latent representation for each mesh vertex, while~\citet{liu2021neural} learns a backwards warping into the canonical pose.
While these methods achieve impressive results, they also suffer from the innate limitations of template-based approaches; 
i.e., their approximate geometry (e.g., from SMPL) or ambiguous pose conditioning (e.g., skeletal joints) often poses challenges in novel view synthesis.
This becomes particularly problematic for fine-scale deformations such as dynamic cloth animations or local detail in the face which cannot be represented by existing geometric template proxies.
To address these challenges, recent efforts opt for template-free approaches. 
For instance,~\citet{zhang2022neuvv} train a time-conditioned network to predict hyper-spherical harmonics for free-viewpoint human rendering while~\citet{Zhao2022HumanPM} propose a static-to-dynamic approach where per-frame neural surface reconstruction is combined with a hybrid neural tracker to generate neural animated human meshes.
In our work, we also propose a template-free approach since we are aiming for the highest visual quality.

\begin{figure*}
    \includegraphics[width=\linewidth]{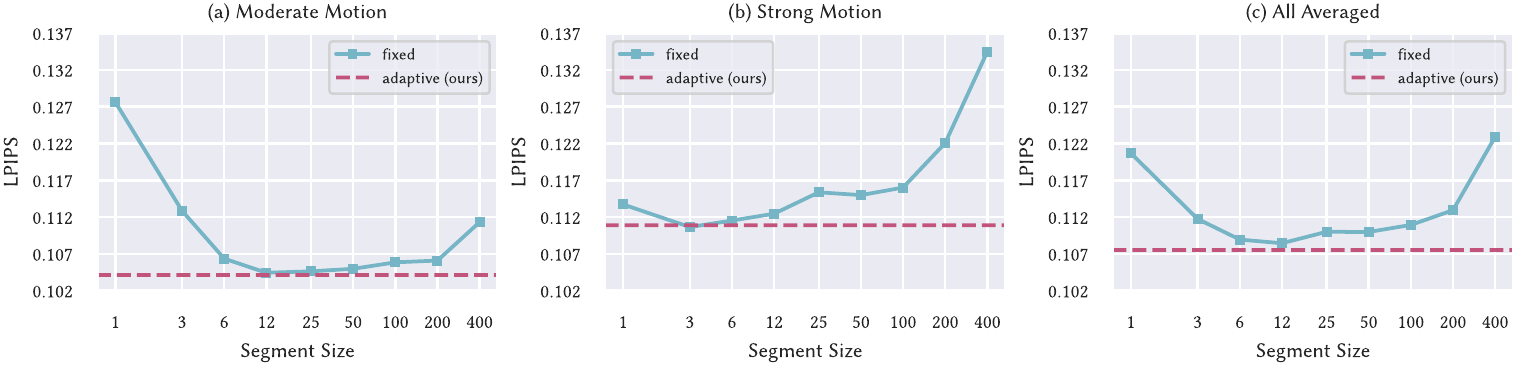}
    \vspace{-0.75cm}
    \caption{ \textbf{Fixed-segment size vs. adaptive partitioning} Using a single 4D representation for an entire sequence (segment size 400) or using a 3D hash grid per frame (segment size 1), give poor results. We observe that finding the middle ground (segment sizes from 3 to 100) leads to better results (a, b, c). Sequences with moderate motions favor larger segment sizes whereas those with stronger motions favor smaller ones (b). Our Adaptive Temporal Partitioning scheme (\S\ref{sec:adaptive_temporal_partitioning}) avoids the costly hyper-parameter search for the optimal, global segment size, and leads to results close to those of optimal segment sizes (a, b). On average, our adaptive method is better than using any fixed segment size (c). These experiments are performed on 400-frame sequences using shared MLPs. The total number of parameters is kept approximately the same while varying the segment size.}

    \label{fig:segment_size}
\end{figure*}

\section{Method}
Given a set of input videos of a human actor in motion, captured in a multi-view camera setting, our goal is to enable temporally consistent, high-fidelity novel view synthesis. 
To that end, we learn a 4D scene representation using differentiable volumetric rendering \cite{mildenhall2020nerf, neural_volumes}, supervised via multi-view 2D photometric and mask losses that minimize the discrepancy between the rendered images and the set of input RGB images and foreground masks. 
To enable efficient photo-realistic neural rendering of arbitrarily long multi-view data, we use sparse feature hash-grids in combination with shallow multilayer perceptrons (MLPs) \cite{mueller2022instant, SunSC22, hybrid_nerf}.

The core idea of \methodname{} -- as illustrated in Fig.~\ref{fig:method} -- is to partition the time domain into optimally distributed temporal segments, and to represent each segment by a compact 4D feature grid (\S\ref{sec:feature_grid_decomposition}). 
For this purpose, we propose an extension to the TensoRF vector-matrix decomposition of~\citet{tensorf} -- designed for static 3D scenes -- that can support time-varying 4D feature grids.
Our adaptive temporal partitioning (\S\ref{sec:adaptive_temporal_partitioning}) ensures that the total 3D space volume covered by each individual temporal segment is of similar size, which helps our method achieve superior representation power, regardless of the temporal context. 
Furthermore, we use shallow MLPs to transform features into density and view-dependent radiance to be used in the volumetric rendering framework (\S\ref{sec:shared_mlps_and_rendering}). Through sharing information across the temporal domain via both shared MLPs and 4D decomposition, our results are temporally consistent. We refer to the accompanying videos regarding temporal stability.
We supervise our differentiable rendering pipeline with 2D-only losses that measure the errors between the rendered and input RGB images and foreground masks (\S\ref{sec:losses}).

\subsection{4D Feature Grid Decomposition}
\label{sec:feature_grid_decomposition}
Our method models a dynamic 3D scene by combining optimally partitioned 4D segments. 
Each segment $\keyframeIndex$ has its own trainable 4D feature grid $\featureTensor^{(\keyframeIndex)}_{xyzt}: \R^4 \mapsto \R^{\featureDim}$ which encodes a set of $\keyframeSize_{\keyframeIndex}$ consecutive frames ${\keyframeTimeSpace}^{(\keyframeIndex)}\in\{\reprTime_s, \reprTime_{s+1}, \reprTime_{s+2}, ..., \reprTime_{s+\keyframeSize_{\keyframeIndex}-1}\}$. 
Previous works~\cite{ccnerf, tensorf, mueller2022instant} have shown that dense 3D data exhibits redundancies and can be represented more compactly. 
We make the same argument for spatio-temporal data, and define our 4D feature grid as a decomposition of four 3D and four 1D feature grids ($\keyframeIndex$ is dropped for brevity below):
\begin{equation}
\label{eq:decomposition}
\begin{split}
\featureTensor_{xyzt}(\reprPoint_{xyzt}) = \featureTensor_{xyz}(\reprPoint_{xyz}) & \odot \featureTensor_t(\reprPoint_{t}) \\
+ \featureTensor_{xyt}(\reprPoint_{xyt}) & \odot \featureTensor_z(\reprPoint_{z}) \\
+ \featureTensor_{xzt}(\reprPoint_{xzt}) & \odot \featureTensor_y(\reprPoint_{y}) \\
+ \featureTensor_{yzt}(\reprPoint_{yzt}) & \odot \featureTensor_x(\reprPoint_{x})
\end{split} \ ,
\end{equation}
where $\odot$ denotes the hadamard product, and $\reprPoint_{xyzt}\in\R^4$ is the queried point. We represent each 3D grid ($\featureTensor_{xyz}, \featureTensor_{xyt}, \featureTensor_{xzt}, \featureTensor_{yzt}: \R^3 \mapsto \R^{\featureDim}$) with a multi-resolution hash grid \cite{mueller2022instant} -- which has proven to be more efficient than using dense 3D grids -- and each 1D grid ($\featureTensor_t, \featureTensor_z, \featureTensor_y, \featureTensor_x: \R \mapsto \R^{\featureDim}$) with dense array of vectors. In Fig. \ref{fig:compression_vs_psnr}, we show that this compact representation lets our method surpass the quality of per-frame Instant-NGP while using only a fraction of the number of trainable parameters.

\subsection{Adaptive Temporal Partitioning}
\label{sec:adaptive_temporal_partitioning}
Using a single 4D feature grid for an entire sequence becomes impractical for longer sequences. 
Figure \ref{fig:segment_size} shows that representing a long sequence with a single 4D segment performs significantly worse than using multiple fixed-sized segments, especially when considering the total hash capacities are roughly the same.
Therefore, partitioning the sequence plays a critical role in our representation. 
Figure \ref{fig:segment_size} also highlights the impact of motion complexity on the optimal segment size (stronger deformations require shorter segments). 
To mitigate the prohibitive cost of hyper-parameter search for the optimal segment size, we present a greedy algorithm to adaptively select the sizes of segments prior to training. 
Unlike fixed-size temporal partitioning, our method does not require a unique segment size which would be less suited to very long sequences.

\paragraph{Occupancy Grids}
To reason about temporal changes, we analyze per-frame occupancy grids $\occGrid^{\frameIndex}: \spatialSpace\subset\R^3 \mapsto \{0,1\}$ that we compute by carving the free space \cite{space_carving} using foreground masks. We define several terms that will be relevant hereafter. 
First, we define the occupancy grid of a set of frames $\keyframeTimeSpace$ by the logical union of their occupancy grids:
\begin{equation}
\label{eq:occupancy_grid_set}
\occGrid^{\keyframeTimeSpace}(\reprPoint)=\bigvee_{\frameIndex_j\in\keyframeTimeSpace}\occGrid^{\frameIndex_j}(\reprPoint) \ .
\end{equation}
Second, for a set of frames $\keyframeTimeSpace$, \emph{total occupancy} is defined as the number of occupied voxels:
\begin{equation}
\label{eq:total_occupancy}
\occupancy({\keyframeTimeSpace})=\sum_{\reprPoint_j\in\spatialSpace}\occGrid^{\keyframeTimeSpace}(\reprPoint_j) \ .
\end{equation}
Finally, for a set of $\keyframeSize$ consecutive frames $\keyframeTimeSpace=\left\{\frameIndex_0, \frameIndex_1, \frameIndex_2, ..., \frameIndex_{\keyframeSize-1}\right\}$, we define the \emph{expansion factor} as:
\begin{equation}
\label{eq:expansion_factor}
\expansion({\keyframeTimeSpace})=\frac{\occupancy({\keyframeTimeSpace})}{\occupancy(\left\{\frameIndex_0\right\})} \ ,
\end{equation}
which practically indicates how much the union occupancy grid is enlarged from $\frameIndex_0$ onwards, and positively correlates with the motion complexity.

\paragraph{Criteria for spawning new segments}
Given a fixed budget of total number of trainable parameters, our objective is to keep the expansion factor (Equation~\eqref{eq:expansion_factor}) similar for each segment.
To this end, we iterate over each frame with a greedy heuristic and spawn a new segment when the expansion factor exceeds a certain threshold. 
This ensures that each segment represents a similar amount of volume in 3D space, which leads to a fair distribution of the total representation workload.
This can also be regarded as maximizing the efficiency of the 4D decomposition models by adjusting the temporal context of each segment such that the temporal sharing is encouraged for smaller movements and discouraged for larger ones.
In Figure \ref{fig:expansion_factor_threshold}, we experiment with several threshold values for the expansion factor, and set it to $1.25$ for all our experiments.

\subsection{Shared MLPs and Volume Rendering}
\label{sec:shared_mlps_and_rendering}
Similarly to previous work \cite{mildenhall2020nerf}, we describe the distribution of the radiance in a scene at time instance $\reprTime$ using the volumetric rendering formulation with emission and absorption:
\begin{equation}
\label{eq:volume_rendering}
    \colorPred(\ray, \reprTime)=\int_{\volrenDistance_{min}}^{\volrenDistance_{max}}T(\volrenDistance)\volrenDensity(\ray(\volrenDistance), \reprTime)\volrenRadiance(\ray(\volrenDistance), \reprDirection, \reprTime)d\volrenDistance \ ,
\end{equation}
where $T(\volrenDistance)$ is the transmittance, $\ray(\volrenDistance)$ is the point on the ray $\ray$ at distance $\volrenDistance$, $\volrenDensity(\ray(\volrenDistance), \reprTime)$ denotes the volumetric density, and $\volrenRadiance(\ray(\volrenDistance), \reprDirection, \reprTime)$ indicates the radiance emitted along the direction $\reprDirection$. 
We solve this integral numerically using quadrature \cite{max1995optical}. 
Similarly to \citet{mueller2022instant}, we use two shallow MLPs to model density and view-dependent radiance. 
First, we leverage a 3-layer network, $\densityMLP: \R^{32} \mapsto \R^{16}$, to generate density $\volrenDensity(\reprPoint, \reprTime) \in \R$ and geometry features $\geometryFeatures(\reprPoint, \reprTime) \in \R^{15}$ for any point $\reprPoint \in \R^3$ in time $\reprTime$ of segment $\keyframeIndex$
\begin{equation}
\label{eq:density_mlp}
    \{\volrenDensity(\reprPoint, \reprTime), \geometryFeatures(\reprPoint, \reprTime)\} = \densityMLP(\featureTensor_{xyzt}^{(\keyframeIndex)}(\reprPoint, \reprTime))
     \ .
\end{equation}
Then, we employ a 4-layer network $\radianceMLP: \R^{31} \mapsto \R^{3}$ to produce view-dependent RGB radiance values:
\begin{equation}
\label{eq:radianceMLP}
    \volrenRadiance(\reprPoint, \reprDirection, \reprTime)\ = \radianceMLP(\SH(\reprDirection), \geometryFeatures(\reprPoint, \reprTime))
     \ ,
\end{equation}
where $\SH(\reprDirection)\in\R^{16}$ is the encoding of the viewing direction $\reprDirection$ formed by using the first 4 bands of the spherical harmonics.
Although each spatio-temporal segment has its own trainable 4D feature grid, these two MLPs are shared by an entire sequence.

\subsection{Losses}
\label{sec:losses}
We utilize RGB images and masks to guide the training. 
First, we enforce the Huber loss \cite{collins1976robust} between the ground truth color $\colorGt(\ray, \reprTime)$ and the predicted pixel color $\colorPred(\ray, \reprTime)$ (Equation \eqref{eq:volume_rendering}): 
\begin{equation}
\label{eq:photometric_loss}
    \loss_{pho}=
    \frac{1}{|\raySpace|}
    \sum_{\ray \in \raySpace}
    \begin{cases}
    \frac{1}{2} l^2,& \text{if } l \leq \delta\\
    \delta\cdot(l - \frac{1}{2}\delta),              & \text{otherwise}
    \end{cases} \ ,
\end{equation}
where $l=|\colorGt(\ray, \reprTime) - \colorPred(\ray, \reprTime)|$. 
This loss is averaged over 3 color channels and we set $\delta=0.01$ in all of our experiments.

In addition to background removal and occupancy grid computation, we use foreground masks to regularize volumetric occupancy similarly to \citet{yariv2020multiview}. 
More specifically, we use the binary cross entropy loss between the ground truth mask $\maskGt(\ray)$ and the accumulated volume rendering weight $\maskPred(\ray)$:
\begin{equation}
\label{eq:bce_loss}
    \loss_{bce}=\frac{1}{|\raySpace|}\sum_{\ray \in \raySpace} \left[\maskGt(\ray)\log(\maskPred(\ray)) +
    (1-\maskGt(\ray))\log((\maskPred(\ray))\right] ,
\end{equation}
where $M(\ray)=1$ and $M(\ray)=0$ denote the pixels on the foreground and the background, respectively, and $\maskPred(\ray)$ is defined as follows,
\begin{equation}
\label{eq:accumulated_raymarching_weight}
    \maskPred(\ray)=\int_{\volrenDistance_{min}}^{\volrenDistance_{max}}T(\volrenDistance)\volrenDensity(\ray(\volrenDistance))d\volrenDistance \ .
\end{equation}
This loss helps to prune the empty space early in the training, which leads to significant speed up in the training iterations. 
Our final loss term is defined as
\begin{equation}
\label{eq:final_loss}
    \loss=\loss_{pho} + \bceLossWeight\loss_{bce} \ ,
\end{equation}
where we set $\bceLossWeight=10^{-3}$ for all of our experiments. 
We refer the reader to the supplemental material for additional training details.

\begin{figure}
    \setlength{\tabcolsep}{0pt}
    \centering
    \begin{tabular}{cc}
        \includegraphics[width=0.5\linewidth]{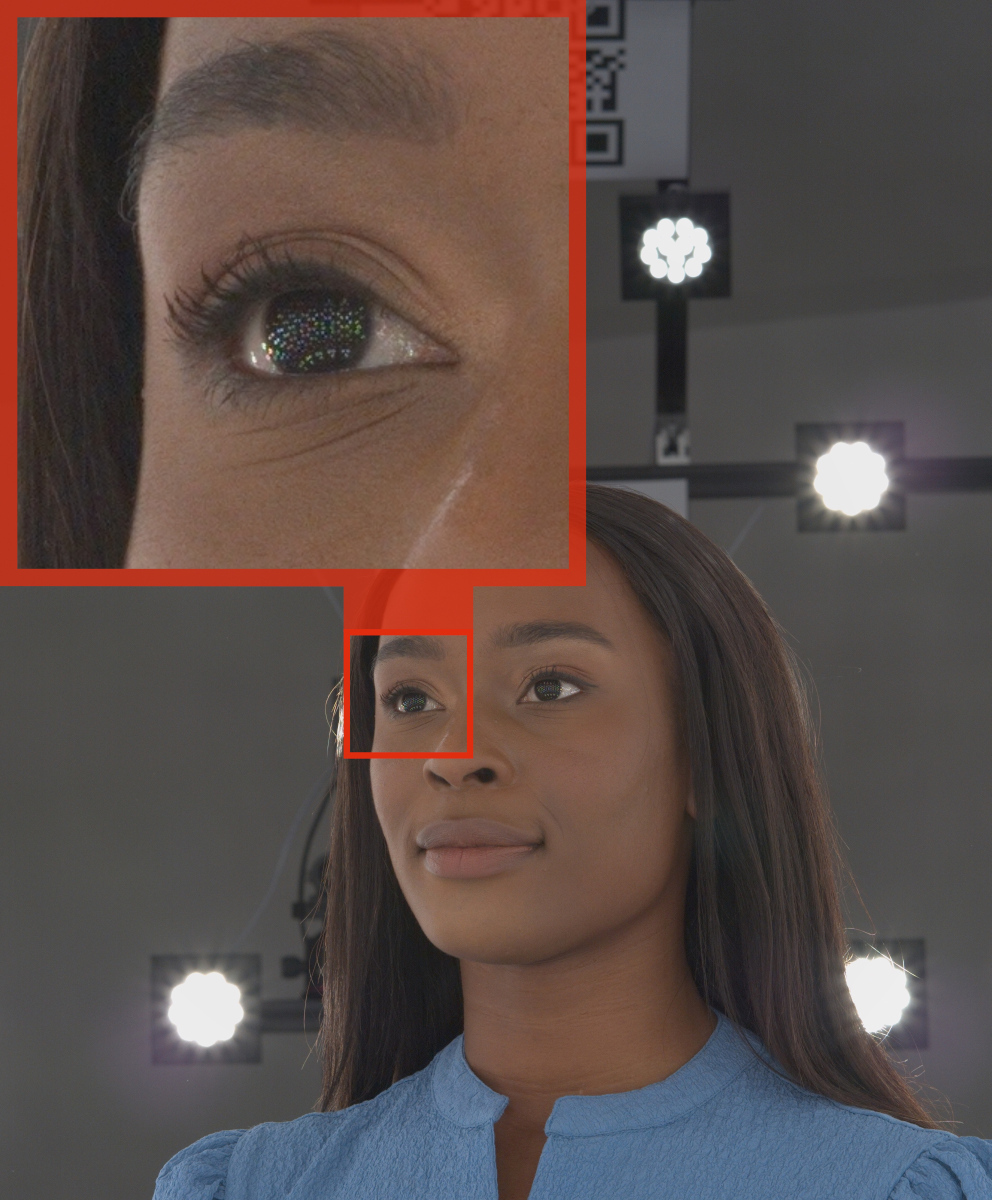} &
        \includegraphics[width=0.5\linewidth]{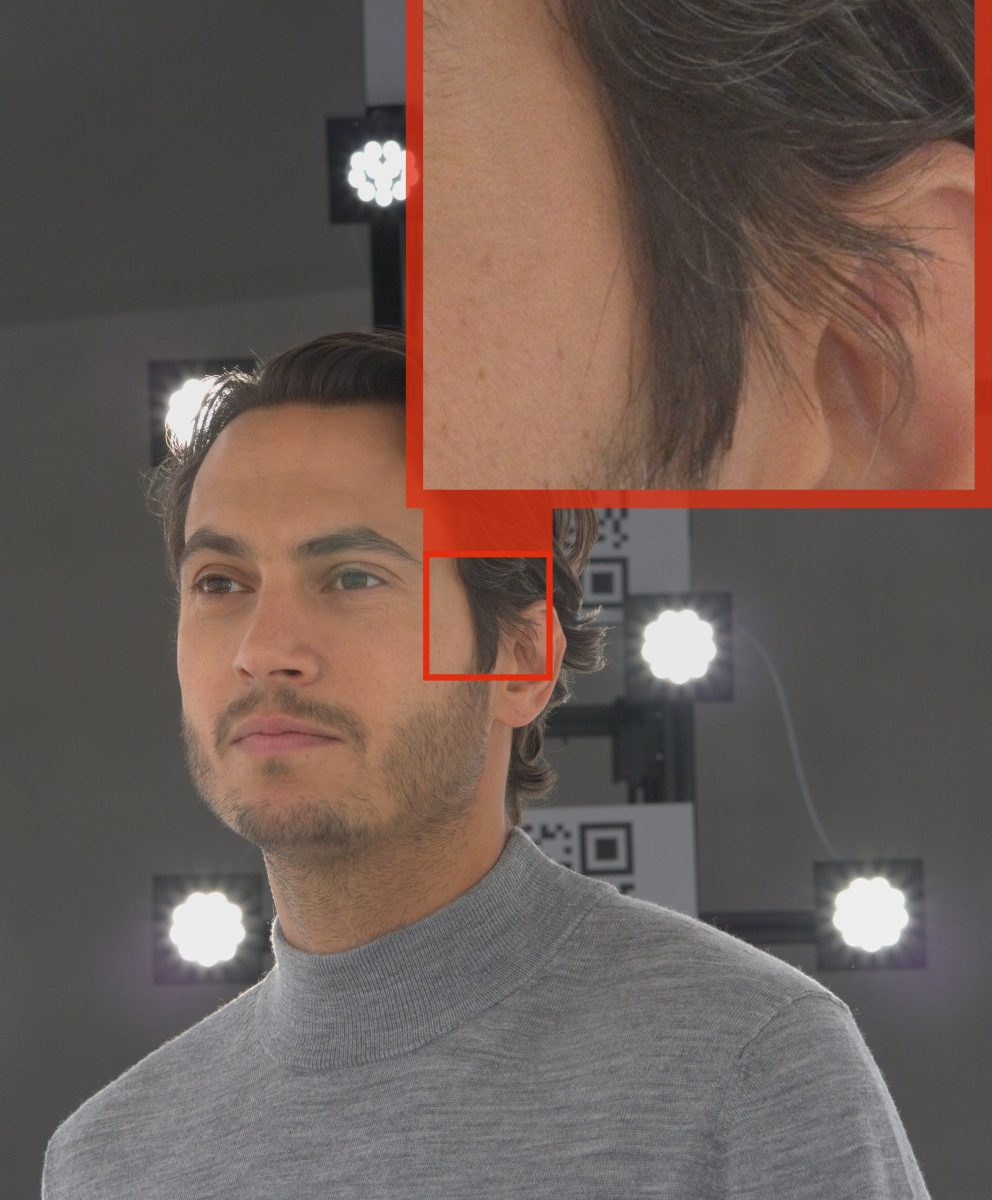} \\
    \end{tabular}
    \caption{\textbf{Dataset resolution} We show closeups for two actors in our \datasetname{} dataset. The cameras record at 12MP each, thus enabling the capture of eyelashes, wrinkles, and hair strands.}
    \label{fig:closeups}
\end{figure}
\begin{figure*}
    \includegraphics[width=\linewidth]{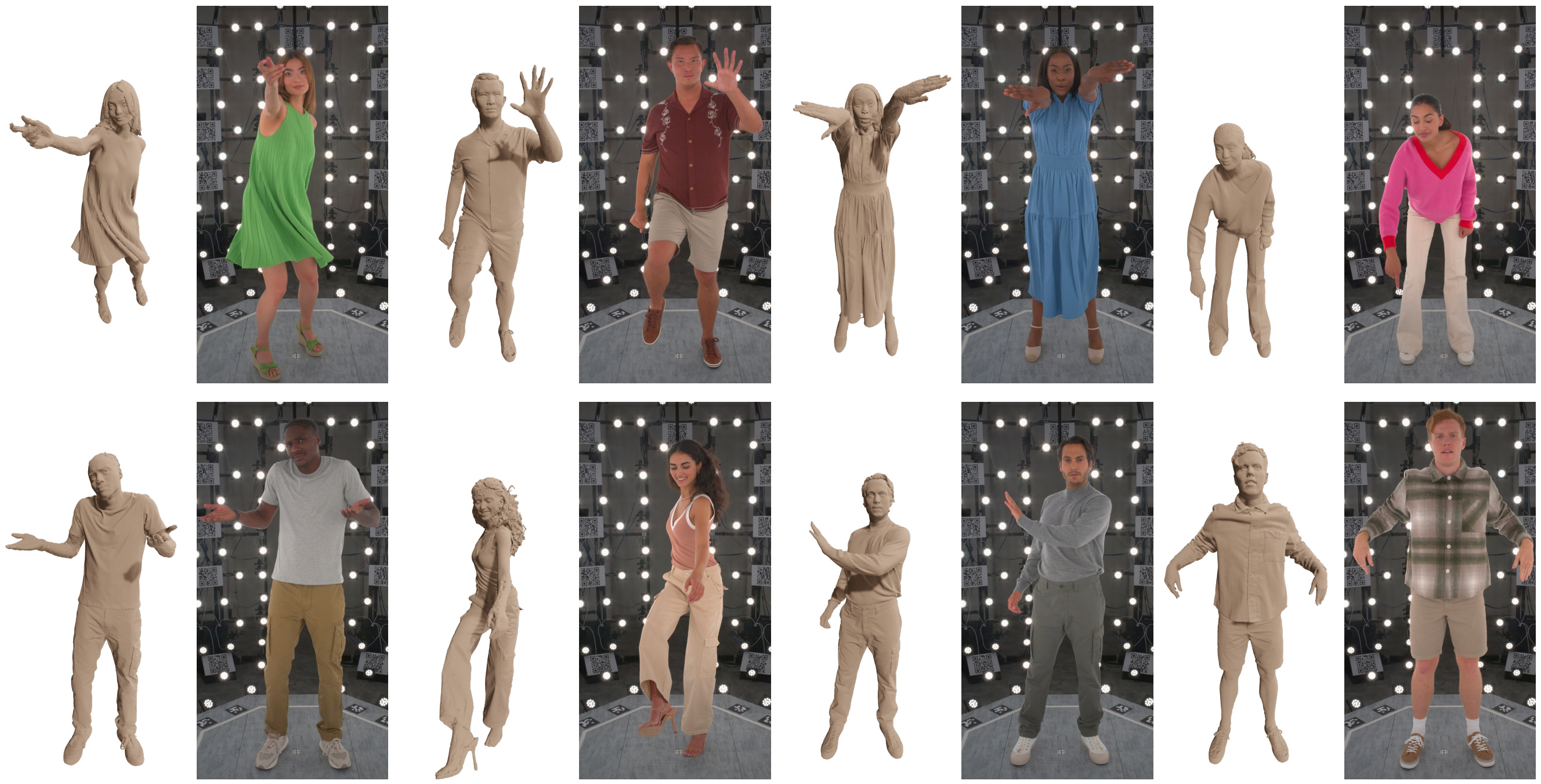}
    \caption{
    \textbf{Actors and meshes.} Our \datasetname{} dataset contains 8 actors with casual clothing such as skirts or shorts. Each sequence is captured by 160 camera, each recording at 12MP. In addition to the recorded images, we also provide high-quality, per-frame mesh reconstructions with approximately 500k vertices.
    }
    \label{fig:actor-meshes}
\end{figure*}

\section{Dataset}\label{sec:dataset}
Our dataset, \datasetname{}, consists of $39,765$ frames of dynamic human motion captured using multi-view video. 
We used a proprietary multi-camera capture system combined with an LED array for global illumination. 
The camera system comprises 160 12MP Ximea cameras operating at 25fps. Close-up details that are captured at this resolution are highlighted in Fig.~\ref{fig:closeups}.
The lighting system provides a programmable lighting array of 420 LEDs that are time-synchronized to the camera shutter. 
All cameras were set to a shutter speed of 650us to minimize motion blur for fast actions. We additionally reconstruct each frame independently using state-of-the-art multi-view stereo from RealityCapture~\cite{RealityCapture} with approximately 500k faces per frame (Fig.~\ref{fig:actor-meshes}). The camera array was configured to cover a capture volume of 1.6m diameter and 2.2m height, enabling actors to perform a range of motions at the center.

The dataset comprises 4 female and 4 male actors. 
Each actor performed two 100 second motion sequences of choreographed actions wearing everyday clothing. 
The actors wore either short or long upper and lower body clothing to provide variation in cloth dynamics. 
In the first sequence, each actor followed the same set of 32 actions that activate key joint rotations for shoulders, arms, legs, and torso as well as combined joint activations. 
The actors were directed to return to a resting A-pose between sets of actions. 
In the second sequence, the actors performed 20 randomly selected everyday actions designed to produce more exaggerated body poses such as sports, dance, celebration, and gestures. 
The actors were directed to move continuously throughout the capture to provide more exaggerated body dynamics in motion. 
A comparison of the \datasetname{} with other standard benchmarks is tabulated in Table~\ref{tab:datasets-comparison}.

\begin{table}
\caption{\textbf{Comparison of multi-view human video datasets.} Our new dataset features longer sequences with more cameras at a higher resolution.}
\centering
\resizebox{1.0\columnwidth}{!}{
\begin{tabular}{ccccc}
\toprule 
    Dataset & $\#$ID & $\#$Frames & Resolution & $\#$Cameras \\
\toprule
    Human3.6M~\cite{ionescu2013human3} & 11 & 581k & 1MP & 4 \\
    MPI-INF-3DHP~\cite{mehta2017monocular} & 8 & $>$1.3M & 4MP & 14  \\
    ZJU-Mocap~\cite{peng2021neural} & 9 & $<$ 2,700 & 1MP & 21  \\
    DynaCap~\cite{habermann2021real} & 5 & 27k &  1.2MP & 50-101 \\
    THUman \cite{zheng2019deephuman} & 500 & 500k & 0.4MP & 4 \\
    THUman4 \cite{zheng2022structured} & 3 & <15k & 1.4MP & 24 \\
\midrule
    \datasetname{} (ours) & 8 & 39,765 & \textbf{12MP} & \textbf{160} \\
\end{tabular}
}

\label{tab:datasets-comparison}
\vspace{10pt}
\end{table}

\section{Evaluation}

To demonstrate the ability of \methodname{} to represent long sequences and fine details at 12MP, we perform extensive quantitative and qualitative experiments with our \datasetname{} dataset.
As \methodname{} is a temporal method, we also highly recommend watching the supplementary videos.

\begin{figure*}
    \setlength{\tabcolsep}{1pt}
    \setlength{\mrgone}{0.15\textwidth}
    \centering
    \begin{tabular}{ccccccc}

        \includegraphics[width=\mrgone]{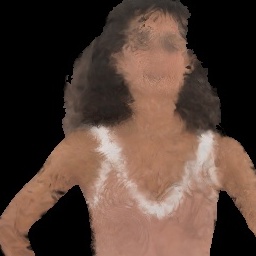}&
        \includegraphics[width=\mrgone]{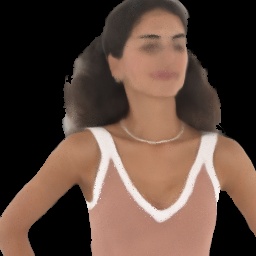}&
        \includegraphics[width=\mrgone]{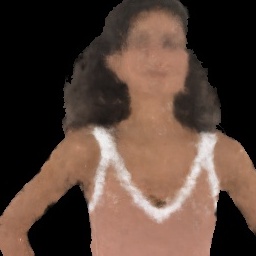}&
        \includegraphics[width=\mrgone]{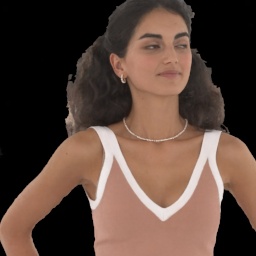}&
        \includegraphics[width=\mrgone]{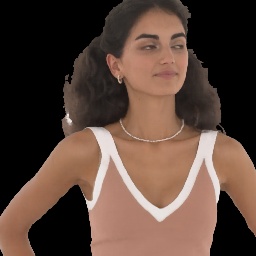}&
        \includegraphics[width=\mrgone]{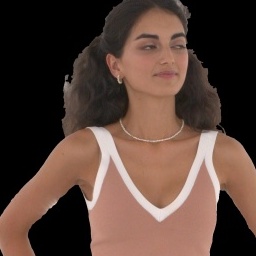}\\

        \includegraphics[width=\mrgone]{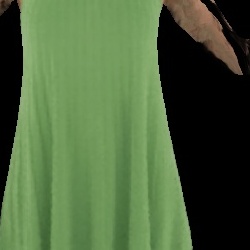}&
        \includegraphics[width=\mrgone]{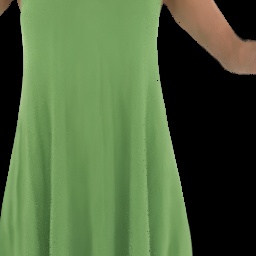}&
        \includegraphics[width=\mrgone]{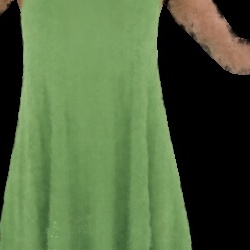}&
        \includegraphics[width=\mrgone]{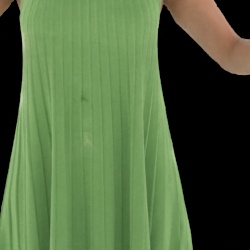}&
        \includegraphics[width=\mrgone]{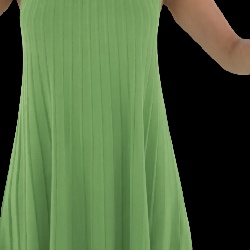}&
        \includegraphics[width=\mrgone]{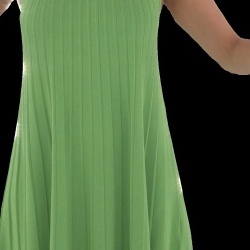}\\

        \includegraphics[width=\mrgone]{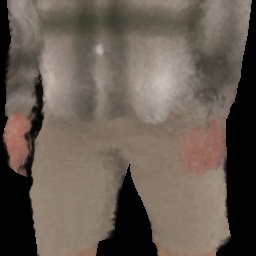}&
        \includegraphics[width=\mrgone]{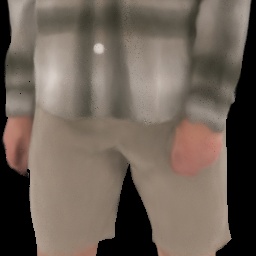}&
        \includegraphics[width=\mrgone]{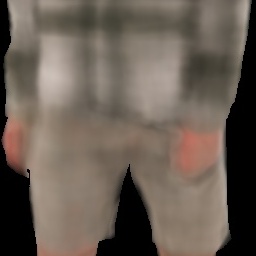}&
        \includegraphics[width=\mrgone]{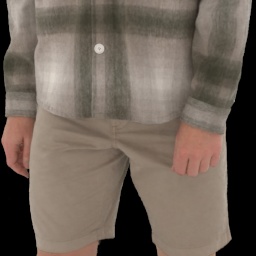}&
        \includegraphics[width=\mrgone]{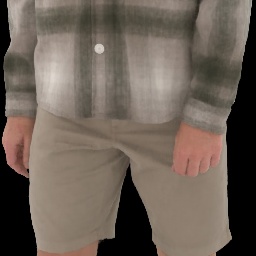}&
        \includegraphics[width=\mrgone]{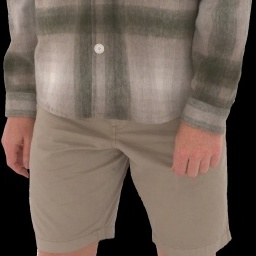}\\

        \includegraphics[width=\mrgone]{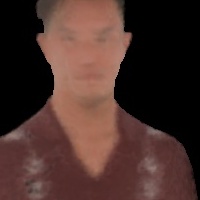}&
        \includegraphics[width=\mrgone]{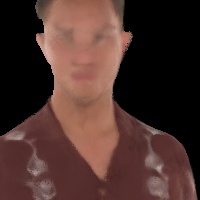}&
        \includegraphics[width=\mrgone]{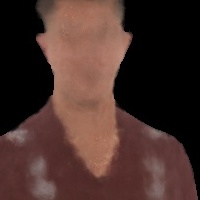}&
        \includegraphics[width=\mrgone]{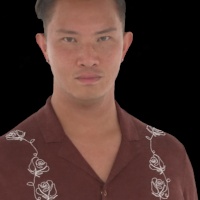}&
        \includegraphics[width=\mrgone]{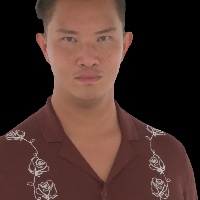}&
        \includegraphics[width=\mrgone]{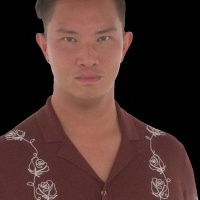}\\

        \includegraphics[width=\mrgone]{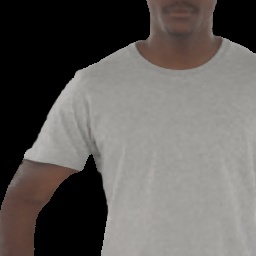}&
        \includegraphics[width=\mrgone]{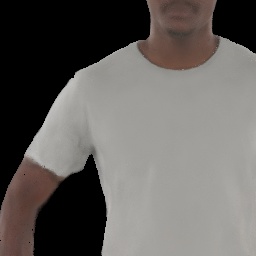}&
        \includegraphics[width=\mrgone]{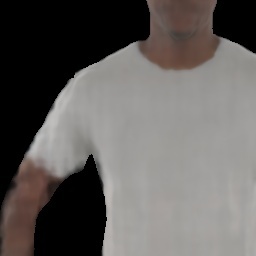}&
        \includegraphics[width=\mrgone]{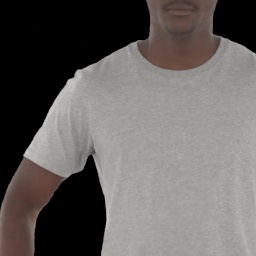}&
        \includegraphics[width=\mrgone]{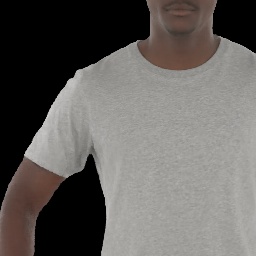}&
        \includegraphics[width=\mrgone]{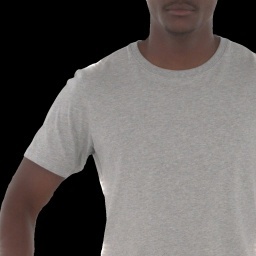}\\

        \includegraphics[width=\mrgone]{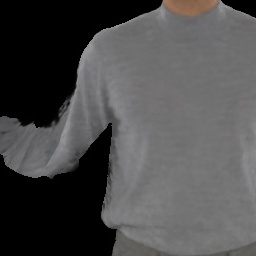}&
        \includegraphics[width=\mrgone]{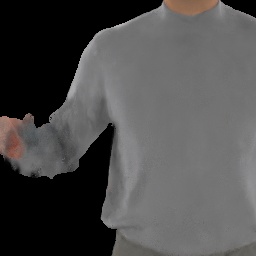}&
        \includegraphics[width=\mrgone]{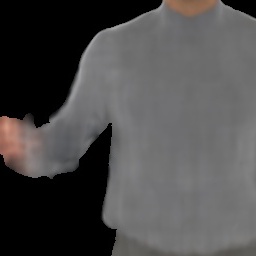}&
        \includegraphics[width=\mrgone]{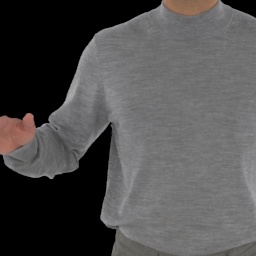}&
        \includegraphics[width=\mrgone]{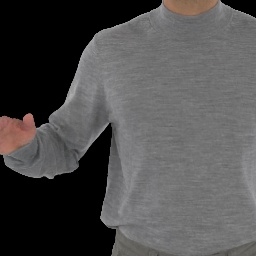}&
        \includegraphics[width=\mrgone]{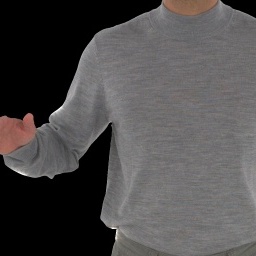}\\
        
        \includegraphics[width=\mrgone]{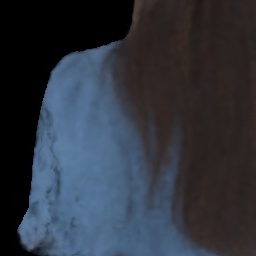}&
        \includegraphics[width=\mrgone]{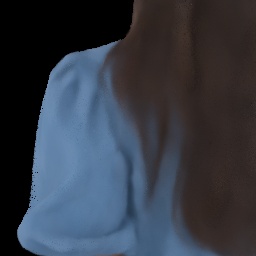}&
        \includegraphics[width=\mrgone]{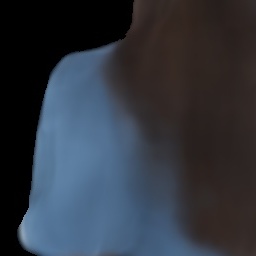}&
        \includegraphics[width=\mrgone]{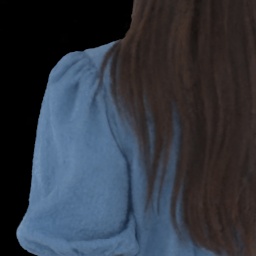}&
        \includegraphics[width=\mrgone]{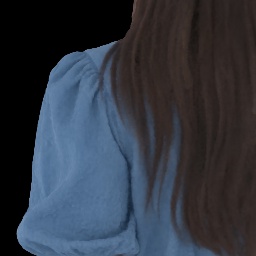}&
        \includegraphics[width=\mrgone]{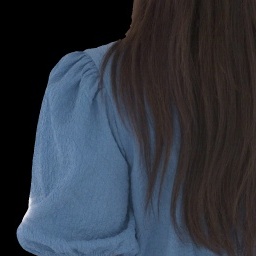}\\
        
         NDVG &
         HyperNeRF &
         TiNeuVox &
         NGP &
         Ours &
         GT &
    \end{tabular}
    \caption{\textbf{Qualitative comparison.} The synthesis quality of \methodname{} is visually compared to the 4 baselines NDVG~\cite{ndvg}, HyperNeRF~\cite{park2021hypernerf}, TiNeuVox~\cite{TiNeuVox} and per-frame NGP~\cite{mueller2022instant} using a sequence of 100 frames. While deformation-based baselines tend to produce blurry results and can fail to capture rapid motions, NGP and ours are able to generate crisp images that are close to groundtruth.
    }
    
    \label{fig:qualitative-comparison}
\end{figure*}

We compare our method against six state-of-the-art baselines. There are three deformation-based approaches for general scenes: NDVG \cite{ndvg}, HyperNeRF~\cite{park2021hypernerf}, and TiNeuVox~\cite{TiNeuVox} along with two human-specific methods: Neural Body~\cite{peng2021neural} and TAVA~\cite{li2022tava}. As an additional baseline, we train Instant-NGP~\cite{mueller2022instant} independently on each frame. For all  baselines, we use the official implementations that are publicly available and tune hyper-parameters to achieve best possible results. A visual comparison between the baselines and our method can be found on Fig.~\ref{fig:qualitative-humanbaselines} and Fig.~\ref{fig:qualitative-comparison}.

\subsection{Evaluation Protocol}
In all experiments, we use the same set of 124 training cameras, 10 validation cameras and 14 test cameras. For one frontal test camera we render a video that is used to compute the VMAF~\cite{li2016toward} score, and we alternate through the remaining test cameras to compute PSNR, LPIPS~\cite{zhang2018perceptual} and SSIM~\cite{wang2004image}. The numerical results are averaged over 8 actors for all the experiments. As some baselines fail to produce reasonable results at full resolution, we compare on $4\times$ downscaled data (per axis), producing better relative performance compared to our results. To test the performance of \methodname{} on high-resolution data, we perform additional experiments in full resolution (12MP) in \S\ref{sec:resolution}. More details on the protocol can be found in the supplementary material.

\subsection{Quality vs Number of Frames}

\begin{figure}
    \setlength{\tabcolsep}{0.3pt}
    \setlength{\mrgone}{0.12\textwidth}
    \centering
    \begin{tabular}{ccccccc}
        
        \includegraphics[width=\mrgone]{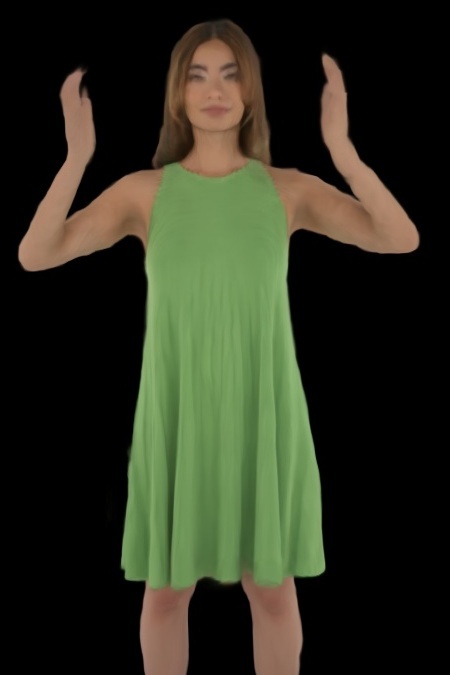}&
        \includegraphics[width=\mrgone]{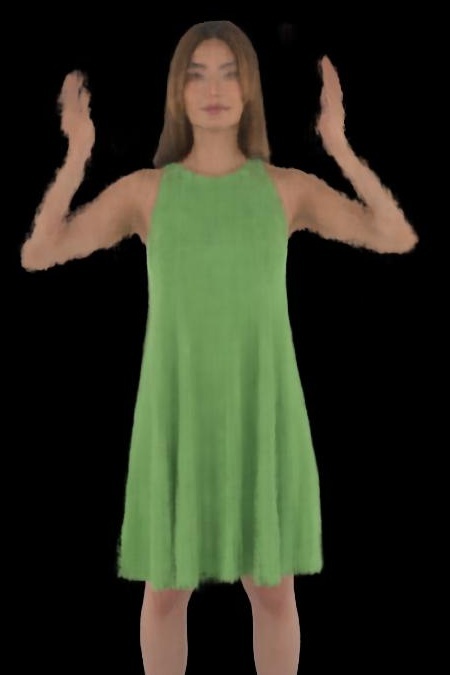}&
        \includegraphics[width=\mrgone]{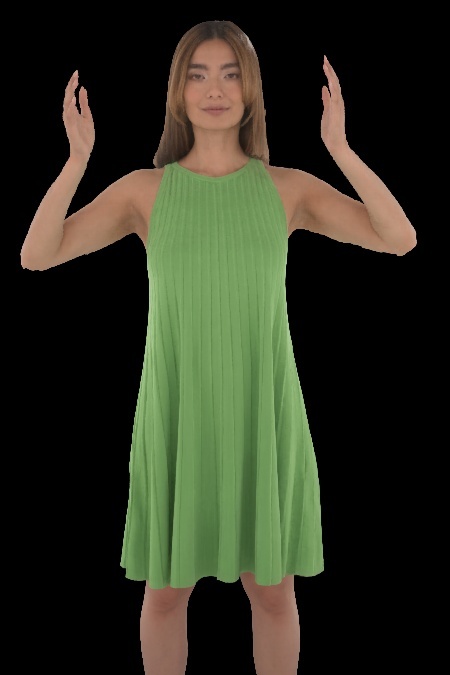}&
        \includegraphics[width=\mrgone]{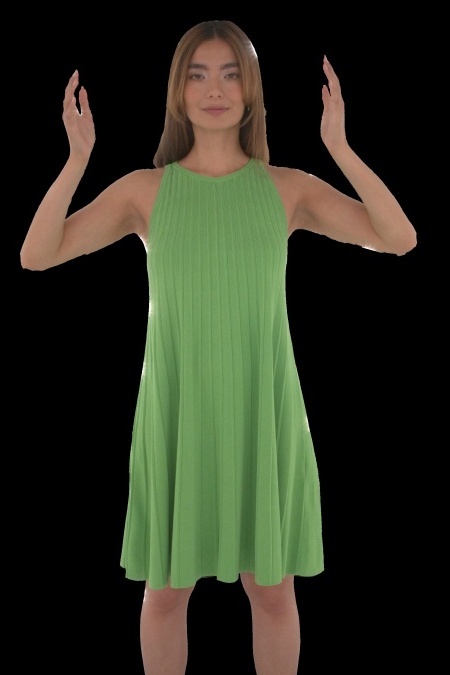}\\

        \includegraphics[width=\mrgone]{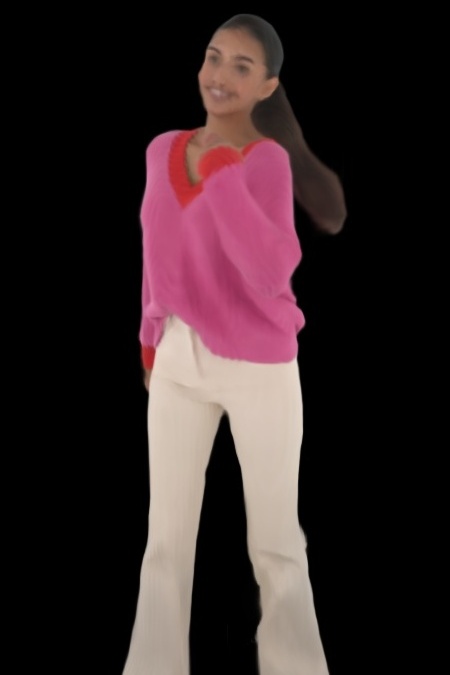}&
        \includegraphics[width=\mrgone]{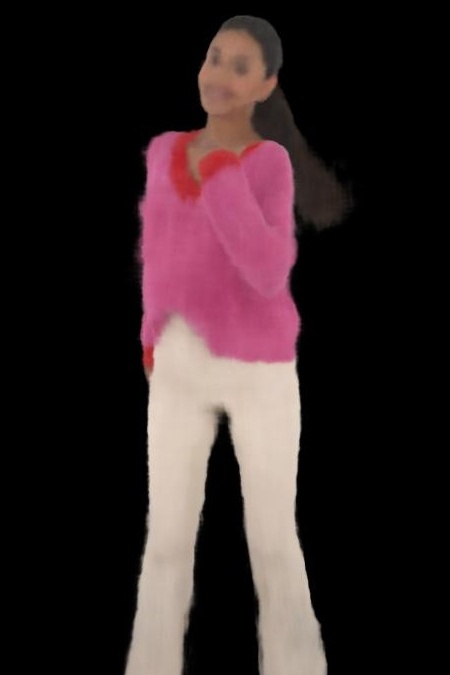}&
        \includegraphics[width=\mrgone]{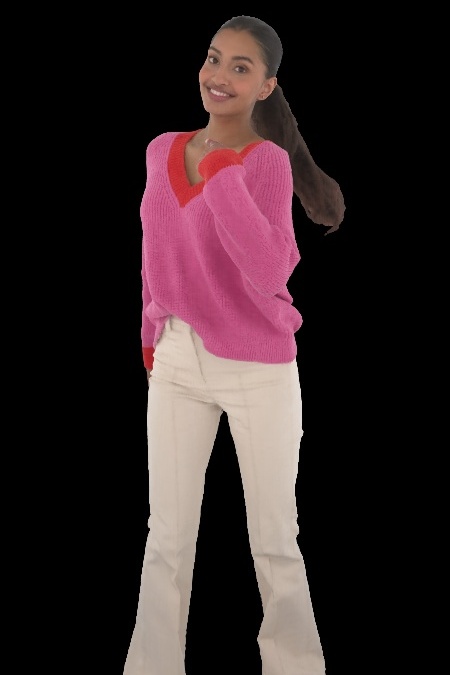}&
        \includegraphics[width=\mrgone]{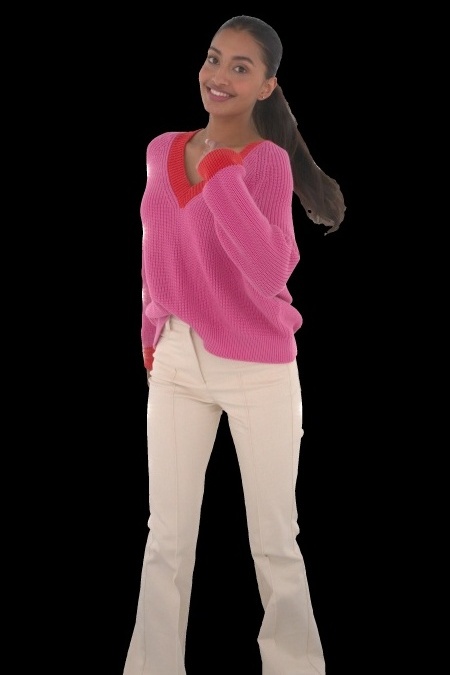}\\
         Neural Body &
         TAVA &
         Ours &
         GT &
    \end{tabular}
    \caption{\textbf{Comparison with human-specific methods.} 
    Although \citet{peng2021neural, li2022tava} incorporate geometry and pose information, they fail to capture fine details, and produce blurrier results compared to \methodname{}.
    }
    
    \label{fig:qualitative-humanbaselines}
\end{figure}

In Table~\ref{tab:frames-vs-quality}, we analyze each method for various sequence lengths using per-frame metrics PSNR, LPIPS, and SSIM, as well as the temporal metric VMAF, which measures perceptual quality of the generated videos and correlates  well with temporal consistency. Due to our efficient 4D feature grid structure and its ability to scale to arbitrarily long sequences via temporal partitioning, \methodname{} consistently outperforms the baselines. Existing methods that predict a deformation field struggle to represent long sequences with complex motion. This is mainly due to rapid topological changes and limited representation power for deformations, and the effect is typically reflected in both the per-frame and  temporal metrics. Previous works ~\cite{DynIBaR,shao2022tensor4d} have also observed similar disadvantages on complex or fast-changing scenes for deformation-based approaches. Although human-specific baselines perform better than deformation-based ones on average, they still lack visual details, and tend to produce blurry results compared to \methodname{} as illustrated in Fig.~\ref{fig:qualitative-humanbaselines}. On the other hand, per-frame Instant-NGP excels on per-frame metrics, but it lacks temporal stability, and uses $20\times$ more trainable parameters compared to our method (Fig. \ref{fig:compression_vs_psnr}).

\begin{table}[t!]
\caption{\textbf{Numerical evaluation on \datasetname{}}. We demonstrate results using the standard visual metrics and VMAF to measure perceptual video quality. \methodname{} outperforms baselines on all sequences. Instant-NGP -- trained per frame separately -- demonstrates better LPIPS, but struggles in terms of temporal consistency and memory footprint (Fig. \ref{fig:compression_vs_psnr}). The \highlightBest{best} and the \highlightSecondBest{second best} results are highlighted.}
 \centering
 \resizebox{\linewidth}{!}{
\begin{tabular}{clcccccc}                                                                                                                                          
\toprule                                                                  
\multicolumn{1}{l}{Method} & Metric & 20 & 50 & 100 & 250 & 500 & 1000 \\                                                                                                          
\midrule
\multirow{4}{*}{Ours}& $\downarrow$ LPIPS & \bestCellColor{0.095} & \secondBestCellColor{0.100} & \secondBestCellColor{0.097} & \secondBestCellColor{0.100} & \secondBestCellColor{0.102} & \secondBestCellColor{0.107} \\
& $\uparrow$ PSNR & \bestCellColor{30.30} & \bestCellColor{30.05} & \bestCellColor{29.83} & \bestCellColor{29.26} & \bestCellColor{29.34} & \bestCellColor{29.05} \\
& $\uparrow$ SSIM & \bestCellColor{0.918} & \bestCellColor{0.918} & \bestCellColor{0.921} & \bestCellColor{0.920} & \bestCellColor{0.919} & \bestCellColor{0.913} \\
& $\uparrow$ VMAF & \bestCellColor{83.67} & \bestCellColor{84.43} & \bestCellColor{85.62} & \bestCellColor{85.28} & \bestCellColor{85.33} & \bestCellColor{85.74} \\
\midrule
\multirow{4}{*}{Instant-NGP}& $\downarrow$ LPIPS & \secondBestCellColor{0.095} & \bestCellColor{0.092} & \bestCellColor{0.094} & \bestCellColor{0.093} & \bestCellColor{0.093} & \bestCellColor{0.093} \\
& $\uparrow$ PSNR & \secondBestCellColor{29.45} & \secondBestCellColor{28.78} & \secondBestCellColor{28.96} & \secondBestCellColor{28.77} & \secondBestCellColor{28.73} & \secondBestCellColor{28.85} \\
& $\uparrow$ SSIM & \secondBestCellColor{0.881} & \secondBestCellColor{0.898} & \secondBestCellColor{0.902} & \secondBestCellColor{0.904} & \secondBestCellColor{0.904} & \secondBestCellColor{0.905} \\
& $\uparrow$ VMAF & \secondBestCellColor{74.15} & \secondBestCellColor{73.23} & \secondBestCellColor{76.70} & \secondBestCellColor{76.77} & \secondBestCellColor{77.28} & \secondBestCellColor{77.60} \\

\midrule

\multirow{4}{*}{TiNeuVox}& $\downarrow$ LPIPS & 0.312 & 0.305 & 0.327 & 0.346 & 0.348 & 0.371 \\
& $\uparrow$ PSNR & 26.07 & 24.14 & 23.11 & 21.82 & 20.94 & 19.80 \\
& $\uparrow$ SSIM & 0.792 & 0.800 & 0.794 & 0.792 & 0.786 & 0.772 \\
& $\uparrow$ VMAF & 56.26 & 47.31 & 40.27 & 31.11 & 24.74 & 18.11 \\

\midrule

\multirow{4}{*}{NDVG}& $\downarrow$ LPIPS & 0.268 & 0.275 & 0.300 & 0.338 & 0.367 & 0.391 \\
& $\uparrow$ PSNR & 26.60 & 23.65 & 22.16 & 19.75 & 17.93 & 16.17 \\
& $\uparrow$ SSIM & 0.823 & 0.811 & 0.793 & 0.765 & 0.741 & 0.716 \\
& $\uparrow$ VMAF & 63.26 & 50.53 & 38.13 & 21.88 & 12.76 & 6.183 \\

\midrule

\multirow{4}{*}{HyperNeRF}& $\downarrow$ LPIPS & 0.250 & 0.235 & 0.251 & 0.270 & 0.302 & 0.325 \\
& $\uparrow$ PSNR & 25.70 & 25.23 & 24.72 & 23.82 & 22.58 & 21.77 \\
& $\uparrow$ SSIM & 0.820 & 0.832 & 0.826 & 0.817 & 0.801 & 0.790 \\
& $\uparrow$ VMAF & 73.08 & 73.05 & 67.17 & 57.51 & 44.84 & 37.01 \\

\midrule

\multirow{4}{*}{Neural Body}& $\downarrow$ LPIPS & 0.305 & 0.308 & 0.310 & 0.318 & 0.340 & 0.367 \\
& $\uparrow$ PSNR & 27.03 & 25.16 & 26.92 & 24.66 & 24.35 & 25.58 \\
& $\uparrow$ SSIM & 0.806 & 0.807 & 0.805 & 0.805 & 0.793 & 0.767 \\
& $\uparrow$ VMAF & 46.72 & 45.27 & 41.83 & 38.71 & 32.13 & 26.95 \\

\midrule

\multirow{4}{*}{TAVA}& $\downarrow$ LPIPS & 0.270 & 0.277 & 0.295 & 0.344 & 0.388 & 0.429 \\
& $\uparrow$ PSNR & 27.44 & 25.75 & 25.05 & 23.61 & 22.40 & 21.50 \\
& $\uparrow$ SSIM & 0.820 & 0.821 & 0.816 & 0.792 & 0.765 & 0.740 \\
& $\uparrow$ VMAF & 66.92 & 58.66 & 54.28 & 38.40 & 23.45 & 13.34 \\

\bottomrule

\end{tabular}
}
\label{tab:frames-vs-quality}
\end{table}

\subsection{Design Choices}
In this section, we validate the effectiveness of our approach, and clarify how we select several hyper parameters. More specifically, we perform ablation studies regarding the 4D feature grid representation (\S\ref{sec:ablation_feature_grid}). We discuss choosing the optimal grid resolution, feature dimensionality and hash size (\S\ref{sec:gridres_fdim}), and argue how segment sizes and expansion factor thresholds are determined (\S\ref{sec:model_size}).

\subsubsection{4D Feature Grid}
\label{sec:ablation_feature_grid}
To validate our choice of feature grid representation, we run a comparison against two ablations of our method where we only alter the 4D feature grid and use a single segment over 100-frame sequences. The first variant simply concatenates 3D spatial coordinates with time to form 4D input to a hash grid -- we dub this as \emph{tNGP}. The second variant also utilizes a 4D decomposition by using six multi-resolution 2D dense grids inspired by the concurrent work \cite{hexplanes, kplanes} -- which we dub as \emph{Hex4D}. Please refer to supplemental to see how Hex4D is formulated. Unlike Hex4D, our method uses four multi-resolution 3D hash grids and four 1D dense grids. Table \ref{tab:feature_grids} indicates that using a decomposition model (Hex4D and ours) for the feature grid is superior when the motion is moderate as the temporal context can be efficiently compressed into lower-rank tensors. Although Hex4D loses its advantage over tNGP for stronger motion, our method consistently outperforms both ablations in both cases.

\subsubsection{Grid Resolution and Feature Dimensionality}
\label{sec:gridres_fdim}
\begin{figure}
    \includegraphics[width=\linewidth]{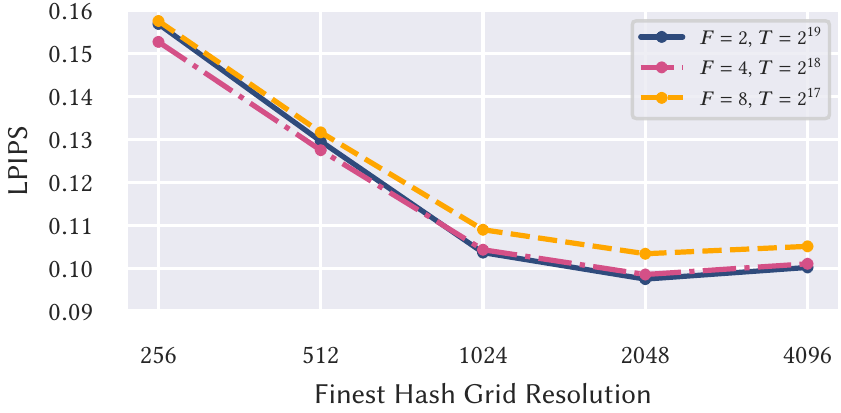}
    \caption{ \textbf{Optimal grid resolution and feature dimensionality.}
    To determine the ideal parameters for our feature grid, we search through several finest grid resolutions ($K_\text{max}$) and feature dimensionalities ($F$) while fixing the number of parameters. As a result, we use $K_\text{max}=2048$ and $F=2$.}
    \label{fig:gridres_fdim}
\end{figure}

In order to determine the optimal grid resolution and feature dimensionality of the 4D feature grid, we perform a parameter search in two dimensions: finest grid resolution ($K_\text{max}$) and per-level feature dimensionality ($F$). To facilitate the search, we perform the experiments on $4\times$ downscaled data over 100-frame sequences, and we fix the coarsest resolution in our multi-resolution grids to $K_\text{min}=32$ and number of resolution levels to $L=16$. To narrow down the search even further, we fix the total number of trainable parameters per hash grid as $T \cdot L\cdot F = 2^{24}$, where $T$ denotes per-level hash size. Finally, we restrict hash size to be $T\leq2^{19}$, because further increase leads to performance penalty, which is also reported by \citet{mueller2022instant}. From the results presented in Fig.~\ref{fig:gridres_fdim}, we pick $K_\text{max}=2048$ and $F=2$. For experiments carried out in full resolution (see \S\ref{sec:resolution} and Fig.~\ref{fig:resolution_insets}), we set $K_\text{max}=8192$ and use $L=24$ to maintain the model capacity so that the finer details can be reconstructed.

\begin{table}
\caption{\textbf{Comparison between different feature grid representations.} Hex4D outperforms tNGP for scenes with moderate motion due to its ability to compress, however, its quality heavily degrades for rapid motion. On the other hand, our method benefits from its compact nature while not sacrificing the quality as much as Hex4D, and it consistently outperforms both ablations. We use default multi-resolution grid parameters for these ablations (see \S\ref{sec:gridres_fdim}). The \highlightBest{best} and the \highlightSecondBest{second best} results are highlighted.}
\begin{tabular*}{\linewidth}{l@{\extracolsep{\fill}}cccccc}
    \toprule
    \multirow{2}{*}{Metric} &
      \multicolumn{3}{c}{Moderate Motion} &
      \multicolumn{3}{c}{Strong Motion} \\
      \cmidrule(lr){2-4} \cmidrule(lr){5-7} 
      & {Hex4D} & {tNGP} & {Ours} & {Hex4D} & {tNGP} & {Ours} \\
      \midrule
    $\downarrow$ LPIPS & \secondBestCellColor{0.105} & 0.129 & \bestCellColor{0.090} & 0.184 & \secondBestCellColor{0.126} & \bestCellColor{0.110} \\
    $\uparrow$ PSNR & \secondBestCellColor{29.89} & 29.27 & \bestCellColor{30.79} & 26.04 & \secondBestCellColor{28.10} & \bestCellColor{28.87} \\
    $\uparrow$ SSIM & \secondBestCellColor{0.915} & 0.906 & \bestCellColor{0.931} & 0.851 & \secondBestCellColor{0.902} & \bestCellColor{0.906} \\
    $\uparrow$ VMAF & 77.50 & \secondBestCellColor{79.18} & \bestCellColor{81.15} & 75.22 & \secondBestCellColor{87.76} & \bestCellColor{89.00} \\
    \bottomrule
  \end{tabular*}
  \label{tab:feature_grids}
\end{table}

\subsubsection{Model Size}
\label{sec:model_size}
\begin{figure}
    \includegraphics[width=\linewidth]{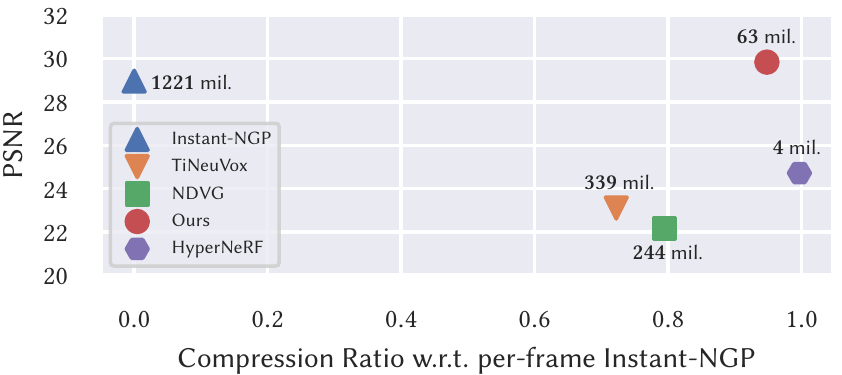}
    \caption{\textbf{Compression ratio vs. PSNR on 100-frame sequence.} We demonstrate number of parameters for each baseline (bold numbers) and the compression ratio with respect to per-frame Instant-NGP. The resulted ratio for \methodname{} is near the most compact one with the best PSNR quality. Here, we define compression ratio for a method \emph{M} as $(1 - \frac{P_\text{M}}{P_\text{I-NGP}})$ where $P_\text{M}$ denotes the number of trainable parameters.}
    \label{fig:compression_vs_psnr}
\end{figure}

Our method uses 4D spatio-temporal segments with various lengths to represent arbitrarily long sequences. Adaptive temporal partitioning (\S\ref{sec:adaptive_temporal_partitioning}) tries to keep the number of trainable parameters per frame approximately the same in order not to lose its representation power. For this reason, the number of trainable parameters scales linearly with the sequence length. Nonetheless, our method remains more compact than most of the baselines. In Fig. \ref{fig:compression_vs_psnr}, we demonstrate that \methodname{} uses only $5.2\%$ of the parameters compared to per-frame Instant-NGP while outperforming it.

\begin{figure}
    \includegraphics[width=\linewidth]{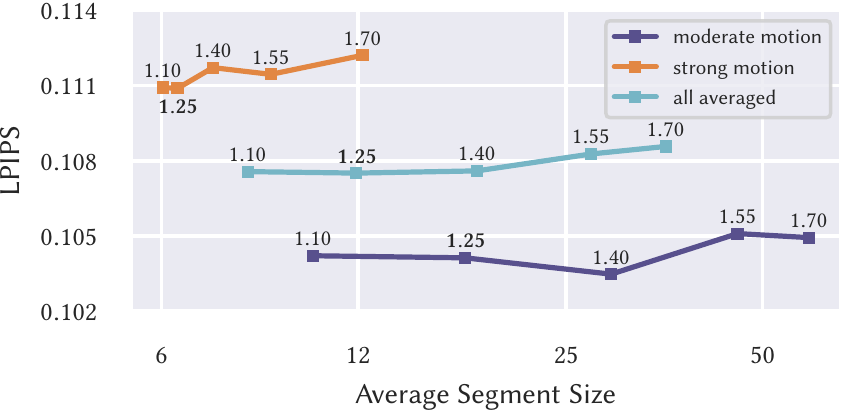}
    \caption{ \textbf{Impact of the expansion factor thresholds and motion complexity on average segment size and quality.} Larger threshold values (indicated by numbers) lead to larger segments on average. Unlike using fixed-size segments, we do not observe a striking difference in quality when the average segment size changes (see Fig.~\ref{fig:segment_size} for a comparison). Furthermore, rapid motions increase the frequency of spawning new segments, and hence lead to smaller segment sizes.}
    \label{fig:expansion_factor_threshold}
\end{figure}

\paragraph{Predefined segment sizes} During adaptive temporal partitioning, we choose segment sizes from a pool of predefined lengths where each one has a hash capacity proportional to its size. For our experiments, we specifically use the segment sizes $6, 12, 25, 50, 100$ with per-level hash sizes $2^{15}, 2^{16}, 2^{17}, 2^{18}, 2^{19}$, respectively, using the \emph{Tiny CUDA neural networks} framework \cite{tiny-cuda-nn}. Moreover, we demonstrate the effect of motion complexity and expansion factor threshold on average segment size in Fig. \ref{fig:expansion_factor_threshold}.

\subsection{Input Resolution}
\label{sec:resolution}
\begin{figure}
    \includegraphics[width=\linewidth]{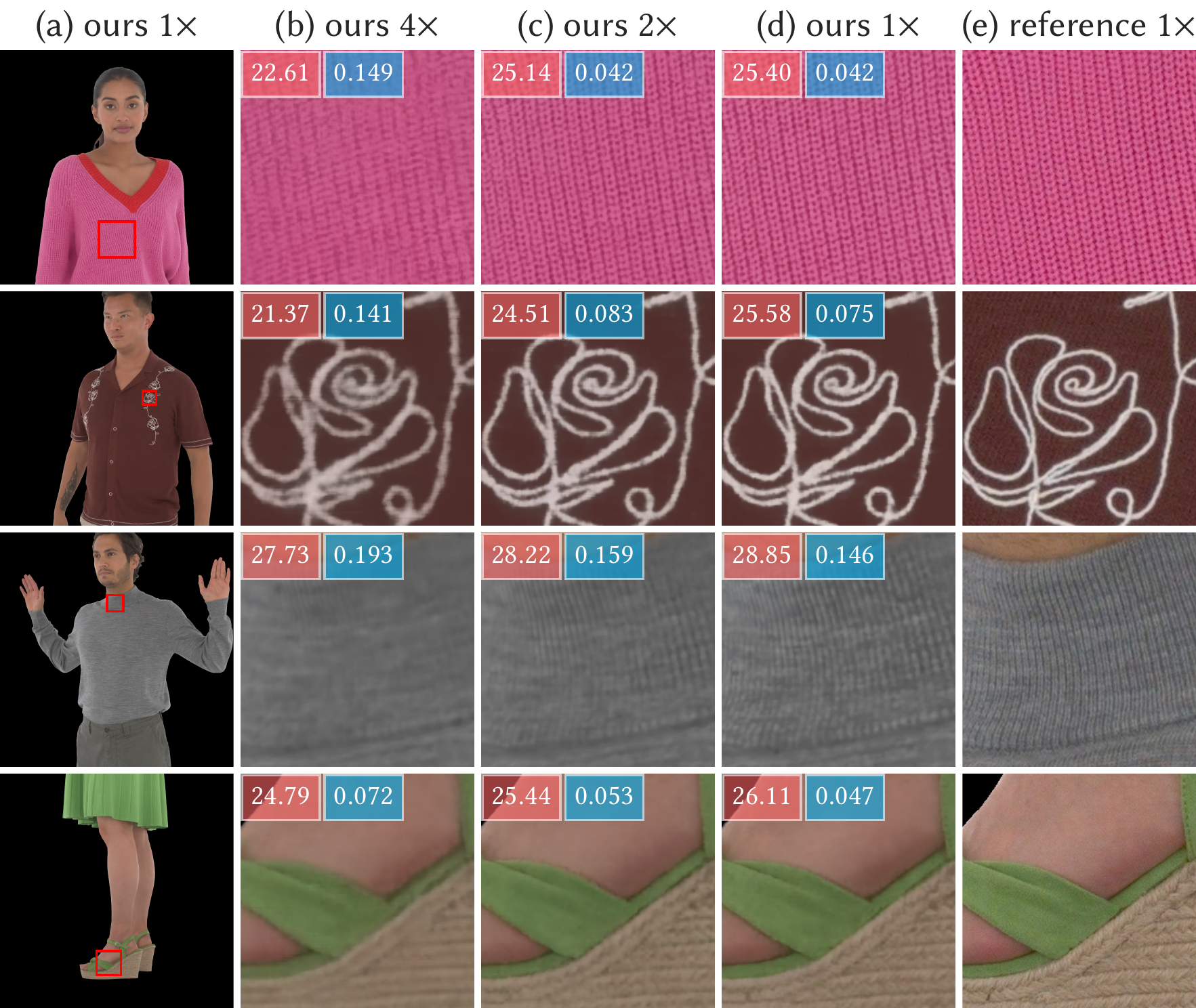}
    \caption{\textbf{Impact of input resolution on the full-resolution results.}
    We illustrate the significance of high resolution data by training \methodname{} on $4\times$ downscaled (b), $2\times$ downscaled (c) and full-resolution (a, c) input to generate full-resolution results. We observe striking differences in capturing finer details as the resolution increases, which is also reflected in highlighted \colorbox[HTML]{e68182}{$\uparrow$PSNR} and \colorbox[HTML]{3cb6e6}{$\downarrow$LPIPS} values.
    }
    \label{fig:resolution_insets}
\end{figure}

Previous publicly available datasets provide images at the resolution of 4MP or less, but our method is designed to capture details beyond this resolution. Fig.~\ref{fig:resolution_insets} illustrates the impact of using downscaled training data on the rendering quality at full resolution. We observe that \methodname{} can recover finer details as the input resolution increases -- see supplemental for numerical results. While it seems natural that scores improve with increasing resolution of training data, we observed that some baselines struggle to represent high-fidelity data and deteriorate instead.

\subsection{Dynamic Furry Animal Dataset}
Although \methodname{} is tailored to \datasetname{}, it is a template-free method which is not necessarily restricted to humans. In fact, our method can be applied to any scene with a foreground object with masks. To demonstrate this ability, we run our method on Dynamic Furry Animal (DFA) \cite{artemis} which is a multi-view dataset of furry animals in motion. In Table \ref{tab:dfa}, we demonstrate that our method can be applied to non-human scenes and it can still surpass the quality of the state of the art. For these experiments, we use default settings (\S\ref{sec:gridres_fdim}) except by setting $L=20$ to account for high-frequent fur details. Additional visual results can be found in the supplementary material.

\begin{table}
\caption{\textbf{Evaluation on the Dynamic Furry Animal (DFA) Dataset \cite{artemis}.}
We show that \methodname{} can reconstruct radiance fields for dynamic sequences of non-human subjects, such as animals in the DFA dataset. Our method achieves state-of-the-art results even though all the baselines except NeuralVolumes uses skeleton information provided in the DFA. For the starred (\textsuperscript{*}) methods, we use the results from Table 1 in \citet{artemis} , and use the exact same evaluation configuration for our method. The \highlightBest{best} and the \highlightSecondBest{second best} results are highlighted.}

\begin{tabular}{clcccc}                                          
\toprule                                                                  
\multicolumn{1}{l}{Method} & Metric & Panda & Cat & Dog & Lion \\                                                                                                                                                     
\midrule        
\multirow{3}{*}{Ours} & $\downarrow$ LPIPS & \bestCellColor{0.030} & \bestCellColor{0.008} & \bestCellColor{0.013} & \bestCellColor{0.025} \\ & $\uparrow$ PSNR & \bestCellColor{36.00} & \bestCellColor{38.43} & \secondBestCellColor{37.79} & \bestCellColor{35.40} \\
& $\uparrow$ SSIM & \bestCellColor{0.986} & \bestCellColor{0.992} & \secondBestCellColor{0.986} & \bestCellColor{0.979}\\

\midrule        
\multirow{3}{*}{Artemis\textsuperscript{*}} & $\downarrow$ LPIPS & \secondBestCellColor{0.031} & \secondBestCellColor{0.012} & \secondBestCellColor{0.022} & \secondBestCellColor{0.035}
\\ & $\uparrow$ PSNR & \secondBestCellColor{33.63} & \secondBestCellColor{37.54} & \bestCellColor{38.95} & \secondBestCellColor{33.09} \\
& $\uparrow$ SSIM & \secondBestCellColor{0.985} & \secondBestCellColor{0.989} & \bestCellColor{0.989} & \secondBestCellColor{0.966}\\

\midrule        
\multirow{3}{*}{Animatable NeRF\textsuperscript{*}} & $\downarrow$ LPIPS & 0.112 & 0.061 & 0.074 & 0.123
\\ & $\uparrow$ PSNR & 26.51 & 31.37 & 31.19 & 27.87 \\
& $\uparrow$ SSIM & 0.957 & 0.973 & 0.975 & 0.944\\

\midrule        
\multirow{3}{*}{Neural Volumes\textsuperscript{*}} & $\downarrow$ LPIPS & 0.116 & 0.087 & 0.129 & 0.123
\\ & $\uparrow$ PSNR & 30.11 & 28.14 & 26.80 & 29.59 \\
& $\uparrow$ SSIM & 0.965 & 0.951 & 0.945 & 0.947\\

\midrule        
\multirow{3}{*}{Neural Body\textsuperscript{*}} & $\downarrow$ LPIPS & 0.110 & 0.067 & 0.075 & 0.111
\\ & $\uparrow$ PSNR & 30.38 & 30.77 & 32.27 & 30.11 \\
& $\uparrow$ SSIM & 0.970 & 0.972 & 0.978 & 0.956\\

\bottomrule
\end{tabular}

\label{tab:dfa}
\end{table}

\subsection{Limitations and Future Work}
Our method can produce high-fidelity radiance field reconstructions of humans in motion, achieving accurate results on novel view synthesis; however, important limitations remain.
To achieve such high-quality results, \methodname{} relies on our newly-introduced \datasetname{} dataset, and optimizes a separate radiance field for each sequence.
It would be interesting to explore training a model on high-end recordings which could then be used as an avatar to target monocular-only test sequences.
While our method reconstructs each frame of a motion sequence, we still do not have explicit control over articulation of the actor outside the training poses. 
One possible way to gain control could be to learn a deformation network for each segment, or to operate with a parametric model to control explicit parameters. 
At the same time, there is also significant room to speed up render times of our method.
Here, a promising direction could be the conversion of our reconstructed radiance field into a hybrid, implicit-explicit representation such as in MobileNeRF~\cite{chen2022mobilenerf}.
Finally, although our model is temporally-stable, the foreground masks are not necessarily consistent across different time frames because they are inferred from independent, per-frame mesh reconstructions, which leads to flickering effect on the silhouette edges. Here, our work would benefit from temporally-consistent background matting techniques, such as~\citet{lin2022robust}.

\section{Conclusion}
We have presented \methodname{}, a novel method to reconstruct a spatio-temporal radiance field that captures human performance at high-fidelity.
At the core of our method lies an intra-frame decomposition of a 4D representation based on a multi-resolution hash grid to capture details.
To handle arbitrarily long sequences with a practical memory budget, we introduce an adaptive splitting technique to share as many features as possible between frames and produce a memory-efficient representation.
To demonstrate the advantages of our method, we have introduced \datasetname{}, the first publicly available multi-view dataset captured with 160 cameras recording 12MP footage. 
Our results have shown high-quality free-viewpoint video, which we believe makes an important step towards production-level novel view synthesis.
Finally, we hope that the release of \datasetname{} dataset and the source code for \methodname{} will enable researchers to drive new advances in photo-realistic reconstruction of virtual humans.

\begin{acks}
We thank Lee Perry-Smith and Henry Pearce at Infinite Realities Ltd for specialist volumetric scanning services to build \datasetname{} using the AEONX Motion Scanning System. Thanks also to Carolin Hecking-Veltman, Daniel Thul, and Tymoteusz Bleja for their generous efforts to help prepare the dataset, Haimin Luo for providing the details of the evaluation for DFA, and the anonymous Siggraph reviewers for their valuable suggestions.
\end{acks}

\bibliographystyle{ACM-Reference-Format}

\begin{thebibliography}{61}


\ifx \showCODEN    \undefined \def \showCODEN     #1{\unskip}     \fi
\ifx \showDOI      \undefined \def \showDOI       #1{#1}\fi
\ifx \showISBNx    \undefined \def \showISBNx     #1{\unskip}     \fi
\ifx \showISBNxiii \undefined \def \showISBNxiii  #1{\unskip}     \fi
\ifx \showISSN     \undefined \def \showISSN      #1{\unskip}     \fi
\ifx \showLCCN     \undefined \def \showLCCN      #1{\unskip}     \fi
\ifx \shownote     \undefined \def \shownote      #1{#1}          \fi
\ifx \showarticletitle \undefined \def \showarticletitle #1{#1}   \fi
\ifx \showURL      \undefined \def \showURL       {\relax}        \fi
\providecommand\bibfield[2]{#2}
\providecommand\bibinfo[2]{#2}
\providecommand\natexlab[1]{#1}
\providecommand\showeprint[2][]{arXiv:#2}

\bibitem[Broxton et~al\mbox{.}(2020)]%
        {broxton2020immersive}
\bibfield{author}{\bibinfo{person}{Michael Broxton}, \bibinfo{person}{John
  Flynn}, \bibinfo{person}{Ryan Overbeck}, \bibinfo{person}{Daniel Erickson},
  \bibinfo{person}{Peter Hedman}, \bibinfo{person}{Matthew Duvall},
  \bibinfo{person}{Jason Dourgarian}, \bibinfo{person}{Jay Busch},
  \bibinfo{person}{Matt Whalen}, {and} \bibinfo{person}{Paul Debevec}.}
  \bibinfo{year}{2020}\natexlab{}.
\newblock \showarticletitle{Immersive light field video with a layered mesh
  representation}.
\newblock \bibinfo{journal}{\emph{ACM Transactions on Graphics (TOG)}}
  \bibinfo{volume}{39}, \bibinfo{number}{4} (\bibinfo{year}{2020}),
  \bibinfo{pages}{86--1}.
\newblock


\bibitem[Cao and Johnson(2023)]%
        {hexplanes}
\bibfield{author}{\bibinfo{person}{Ang Cao} {and} \bibinfo{person}{Justin
  Johnson}.} \bibinfo{year}{2023}\natexlab{}.
\newblock \showarticletitle{HexPlane: A Fast Representation for Dynamic
  Scenes}.
\newblock \bibinfo{journal}{\emph{CVPR}} (\bibinfo{year}{2023}).
\newblock


\bibitem[Carranza et~al\mbox{.}(2003)]%
        {Carranza2003FreeviewpointVO}
\bibfield{author}{\bibinfo{person}{Joel Carranza}, \bibinfo{person}{Christian
  Theobalt}, \bibinfo{person}{Marcus~A. Magnor}, {and}
  \bibinfo{person}{Hans-Peter Seidel}.} \bibinfo{year}{2003}\natexlab{}.
\newblock \showarticletitle{Free-viewpoint video of human actors}. In
  \bibinfo{booktitle}{\emph{ACM Transactions on Graphics (TOG)}}.
\newblock


\bibitem[Chen et~al\mbox{.}(2022b)]%
        {tensorf}
\bibfield{author}{\bibinfo{person}{Anpei Chen}, \bibinfo{person}{Zexiang Xu},
  \bibinfo{person}{Andreas Geiger}, \bibinfo{person}{Jingyi Yu}, {and}
  \bibinfo{person}{Hao Su}.} \bibinfo{year}{2022}\natexlab{b}.
\newblock \showarticletitle{TensoRF: Tensorial Radiance Fields}. In
  \bibinfo{booktitle}{\emph{European Conference on Computer Vision (ECCV)}}.
\newblock


\bibitem[Chen et~al\mbox{.}(2021)]%
        {chen2021snarf}
\bibfield{author}{\bibinfo{person}{Xu Chen}, \bibinfo{person}{Yufeng Zheng},
  \bibinfo{person}{Michael~J Black}, \bibinfo{person}{Otmar Hilliges}, {and}
  \bibinfo{person}{Andreas Geiger}.} \bibinfo{year}{2021}\natexlab{}.
\newblock \showarticletitle{SNARF: Differentiable forward skinning for
  animating non-rigid neural implicit shapes}. In
  \bibinfo{booktitle}{\emph{Proceedings of the IEEE/CVF International
  Conference on Computer Vision}}. \bibinfo{pages}{11594--11604}.
\newblock


\bibitem[Chen et~al\mbox{.}(2022a)]%
        {chen2022mobilenerf}
\bibfield{author}{\bibinfo{person}{Zhiqin Chen}, \bibinfo{person}{Thomas
  Funkhouser}, \bibinfo{person}{Peter Hedman}, {and} \bibinfo{person}{Andrea
  Tagliasacchi}.} \bibinfo{year}{2022}\natexlab{a}.
\newblock \showarticletitle{Mobilenerf: Exploiting the polygon rasterization
  pipeline for efficient neural field rendering on mobile architectures}.
\newblock \bibinfo{journal}{\emph{arXiv preprint arXiv:2208.00277}}
  (\bibinfo{year}{2022}).
\newblock


\bibitem[Collet et~al\mbox{.}(2015)]%
        {Collet2015HighqualitySF}
\bibfield{author}{\bibinfo{person}{Alvaro Collet}, \bibinfo{person}{Ming
  Chuang}, \bibinfo{person}{Pat Sweeney}, \bibinfo{person}{Don Gillett},
  \bibinfo{person}{Dennis Evseev}, \bibinfo{person}{David Calabrese},
  \bibinfo{person}{Hugues Hoppe}, \bibinfo{person}{Adam~G. Kirk}, {and}
  \bibinfo{person}{Steve Sullivan}.} \bibinfo{year}{2015}\natexlab{}.
\newblock \showarticletitle{High-quality streamable free-viewpoint video}. In
  \bibinfo{booktitle}{\emph{ACM Transactions on Graphics (TOG)}},
  Vol.~\bibinfo{volume}{34}. \bibinfo{pages}{1 -- 13}.
\newblock


\bibitem[Collins(1976)]%
        {collins1976robust}
\bibfield{author}{\bibinfo{person}{John~R Collins}.}
  \bibinfo{year}{1976}\natexlab{}.
\newblock \showarticletitle{Robust estimation of a location parameter in the
  presence of asymmetry}.
\newblock \bibinfo{journal}{\emph{The Annals of Statistics}}
  (\bibinfo{year}{1976}), \bibinfo{pages}{68--85}.
\newblock


\bibitem[Epic~Games(2022)]%
        {RealityCapture}
\bibfield{author}{\bibinfo{person}{Inc. Epic~Games}.}
  \bibinfo{year}{2022}\natexlab{}.
\newblock \bibinfo{booktitle}{\emph{{RealityCapture}}}.
\newblock
\urldef\tempurl%
\url{https://www.capturingreality.com}
\showURL{%
\tempurl}
\newblock
\shownote{Accessed: 2023-01-12}.


\bibitem[Fang et~al\mbox{.}(2022a)]%
        {fang2022fast}
\bibfield{author}{\bibinfo{person}{Jiemin Fang}, \bibinfo{person}{Taoran Yi},
  \bibinfo{person}{Xinggang Wang}, \bibinfo{person}{Lingxi Xie},
  \bibinfo{person}{Xiaopeng Zhang}, \bibinfo{person}{Wenyu Liu},
  \bibinfo{person}{Matthias Nie{\ss}ner}, {and} \bibinfo{person}{Qi Tian}.}
  \bibinfo{year}{2022}\natexlab{a}.
\newblock \showarticletitle{Fast Dynamic Radiance Fields with Time-Aware Neural
  Voxels}.
\newblock \bibinfo{journal}{\emph{arXiv preprint arXiv:2205.15285}}
  (\bibinfo{year}{2022}).
\newblock


\bibitem[Fang et~al\mbox{.}(2022b)]%
        {TiNeuVox}
\bibfield{author}{\bibinfo{person}{Jiemin Fang}, \bibinfo{person}{Taoran Yi},
  \bibinfo{person}{Xinggang Wang}, \bibinfo{person}{Lingxi Xie},
  \bibinfo{person}{Xiaopeng Zhang}, \bibinfo{person}{Wenyu Liu},
  \bibinfo{person}{Matthias Nie\ss{}ner}, {and} \bibinfo{person}{Qi Tian}.}
  \bibinfo{year}{2022}\natexlab{b}.
\newblock \showarticletitle{Fast Dynamic Radiance Fields with Time-Aware Neural
  Voxels}. In \bibinfo{booktitle}{\emph{SIGGRAPH Asia 2022 Conference Papers}}.
\newblock


\bibitem[Fridovich-Keil et~al\mbox{.}(2023)]%
        {kplanes}
\bibfield{author}{\bibinfo{person}{Sara Fridovich-Keil},
  \bibinfo{person}{Giacomo Meanti}, \bibinfo{person}{Frederik~Rahbæk Warburg},
  \bibinfo{person}{Benjamin Recht}, {and} \bibinfo{person}{Angjoo Kanazawa}.}
  \bibinfo{year}{2023}\natexlab{}.
\newblock \showarticletitle{K-Planes: Explicit Radiance Fields in Space, Time,
  and Appearance}. In \bibinfo{booktitle}{\emph{CVPR}}.
\newblock


\bibitem[Fridovich-Keil et~al\mbox{.}(2022)]%
        {fridovich2022plenoxels}
\bibfield{author}{\bibinfo{person}{Sara Fridovich-Keil}, \bibinfo{person}{Alex
  Yu}, \bibinfo{person}{Matthew Tancik}, \bibinfo{person}{Qinhong Chen},
  \bibinfo{person}{Benjamin Recht}, {and} \bibinfo{person}{Angjoo Kanazawa}.}
  \bibinfo{year}{2022}\natexlab{}.
\newblock \showarticletitle{Plenoxels: Radiance Fields Without Neural
  Networks}. In \bibinfo{booktitle}{\emph{Proceedings of the IEEE/CVF
  Conference on Computer Vision and Pattern Recognition}}.
  \bibinfo{pages}{5501--5510}.
\newblock


\bibitem[Guo et~al\mbox{.}(2022a)]%
        {guo2022neural}
\bibfield{author}{\bibinfo{person}{Xiang Guo}, \bibinfo{person}{Guanying Chen},
  \bibinfo{person}{Yuchao Dai}, \bibinfo{person}{Xiaoqing Ye},
  \bibinfo{person}{Jiadai Sun}, \bibinfo{person}{Xiao Tan}, {and}
  \bibinfo{person}{Errui Ding}.} \bibinfo{year}{2022}\natexlab{a}.
\newblock \showarticletitle{Neural Deformable Voxel Grid for Fast Optimization
  of Dynamic View Synthesis}. In \bibinfo{booktitle}{\emph{Proceedings of the
  Asian Conference on Computer Vision}}. \bibinfo{pages}{3757--3775}.
\newblock


\bibitem[Guo et~al\mbox{.}(2022b)]%
        {ndvg}
\bibfield{author}{\bibinfo{person}{Xiang Guo}, \bibinfo{person}{Guanying Chen},
  \bibinfo{person}{Yuchao Dai}, \bibinfo{person}{Xiaoqing Ye},
  \bibinfo{person}{Jiadai Sun}, \bibinfo{person}{Xiao Tan}, {and}
  \bibinfo{person}{Errui Ding}.} \bibinfo{year}{2022}\natexlab{b}.
\newblock \showarticletitle{Neural Deformable Voxel Grid for Fast Optimization
  of Dynamic View Synthesis}. In \bibinfo{booktitle}{\emph{Proceedings of the
  Asian Conference on Computer Vision (ACCV)}}.
\newblock


\bibitem[Habermann et~al\mbox{.}(2021)]%
        {habermann2021real}
\bibfield{author}{\bibinfo{person}{Marc Habermann}, \bibinfo{person}{Lingjie
  Liu}, \bibinfo{person}{Weipeng Xu}, \bibinfo{person}{Michael Zollhoefer},
  \bibinfo{person}{Gerard Pons-Moll}, {and} \bibinfo{person}{Christian
  Theobalt}.} \bibinfo{year}{2021}\natexlab{}.
\newblock \showarticletitle{Real-time deep dynamic characters}.
\newblock \bibinfo{journal}{\emph{ACM Transactions on Graphics (TOG)}}
  \bibinfo{volume}{40}, \bibinfo{number}{4} (\bibinfo{year}{2021}),
  \bibinfo{pages}{1--16}.
\newblock


\bibitem[Ionescu et~al\mbox{.}(2013)]%
        {ionescu2013human3}
\bibfield{author}{\bibinfo{person}{Catalin Ionescu}, \bibinfo{person}{Dragos
  Papava}, \bibinfo{person}{Vlad Olaru}, {and} \bibinfo{person}{Cristian
  Sminchisescu}.} \bibinfo{year}{2013}\natexlab{}.
\newblock \showarticletitle{Human3. 6m: Large scale datasets and predictive
  methods for 3d human sensing in natural environments}.
\newblock \bibinfo{journal}{\emph{IEEE transactions on pattern analysis and
  machine intelligence}} \bibinfo{volume}{36}, \bibinfo{number}{7}
  (\bibinfo{year}{2013}), \bibinfo{pages}{1325--1339}.
\newblock


\bibitem[Kanade et~al\mbox{.}(1997)]%
        {kanade1997virtualized}
\bibfield{author}{\bibinfo{person}{Takeo Kanade}, \bibinfo{person}{Peter
  Rander}, {and} \bibinfo{person}{PJ Narayanan}.}
  \bibinfo{year}{1997}\natexlab{}.
\newblock \showarticletitle{Virtualized reality: Constructing virtual worlds
  from real scenes}.
\newblock \bibinfo{journal}{\emph{IEEE multimedia}} \bibinfo{volume}{4},
  \bibinfo{number}{1} (\bibinfo{year}{1997}), \bibinfo{pages}{34--47}.
\newblock


\bibitem[Kutulakos and Seitz(2000)]%
        {space_carving}
\bibfield{author}{\bibinfo{person}{Kiriakos~N Kutulakos} {and}
  \bibinfo{person}{Steven~M Seitz}.} \bibinfo{year}{2000}\natexlab{}.
\newblock \showarticletitle{A theory of shape by space carving}.
\newblock \bibinfo{journal}{\emph{International journal of computer vision}}
  \bibinfo{volume}{38}, \bibinfo{number}{3} (\bibinfo{year}{2000}),
  \bibinfo{pages}{199--218}.
\newblock


\bibitem[Li et~al\mbox{.}(2022b)]%
        {li2022tava}
\bibfield{author}{\bibinfo{person}{Ruilong Li}, \bibinfo{person}{Julian Tanke},
  \bibinfo{person}{Minh Vo}, \bibinfo{person}{Michael Zollh{\"o}fer},
  \bibinfo{person}{J{\"u}rgen Gall}, \bibinfo{person}{Angjoo Kanazawa}, {and}
  \bibinfo{person}{Christoph Lassner}.} \bibinfo{year}{2022}\natexlab{b}.
\newblock \showarticletitle{Tava: Template-free animatable volumetric actors}.
  In \bibinfo{booktitle}{\emph{Computer Vision--ECCV 2022: 17th European
  Conference, Tel Aviv, Israel, October 23--27, 2022, Proceedings, Part
  XXXII}}. Springer, \bibinfo{pages}{419--436}.
\newblock


\bibitem[Li et~al\mbox{.}(2022a)]%
        {li2022neural}
\bibfield{author}{\bibinfo{person}{Tianye Li}, \bibinfo{person}{Mira
  Slavcheva}, \bibinfo{person}{Michael Zollhoefer}, \bibinfo{person}{Simon
  Green}, \bibinfo{person}{Christoph Lassner}, \bibinfo{person}{Changil Kim},
  \bibinfo{person}{Tanner Schmidt}, \bibinfo{person}{Steven Lovegrove},
  \bibinfo{person}{Michael Goesele}, \bibinfo{person}{Richard Newcombe},
  {et~al\mbox{.}}} \bibinfo{year}{2022}\natexlab{a}.
\newblock \showarticletitle{Neural 3D Video Synthesis From Multi-View Video}.
  In \bibinfo{booktitle}{\emph{Proceedings of the IEEE/CVF Conference on
  Computer Vision and Pattern Recognition}}. \bibinfo{pages}{5521--5531}.
\newblock


\bibitem[Li et~al\mbox{.}(2016)]%
        {li2016toward}
\bibfield{author}{\bibinfo{person}{Zhi Li}, \bibinfo{person}{Anne Aaron},
  \bibinfo{person}{Ioannis Katsavounidis}, \bibinfo{person}{Anush Moorthy},
  {and} \bibinfo{person}{Megha Manohara}.} \bibinfo{year}{2016}\natexlab{}.
\newblock \showarticletitle{Toward a practical perceptual video quality
  metric}.
\newblock \bibinfo{journal}{\emph{The Netflix Tech Blog}} \bibinfo{volume}{6},
  \bibinfo{number}{2} (\bibinfo{year}{2016}).
\newblock


\bibitem[Li et~al\mbox{.}(2021)]%
        {li2021neural}
\bibfield{author}{\bibinfo{person}{Zhengqi Li}, \bibinfo{person}{Simon
  Niklaus}, \bibinfo{person}{Noah Snavely}, {and} \bibinfo{person}{Oliver
  Wang}.} \bibinfo{year}{2021}\natexlab{}.
\newblock \showarticletitle{Neural scene flow fields for space-time view
  synthesis of dynamic scenes}. In \bibinfo{booktitle}{\emph{Proceedings of the
  IEEE/CVF Conference on Computer Vision and Pattern Recognition}}.
  \bibinfo{pages}{6498--6508}.
\newblock


\bibitem[Li et~al\mbox{.}(2022c)]%
        {DynIBaR}
\bibfield{author}{\bibinfo{person}{Zhengqi Li}, \bibinfo{person}{Qianqian
  Wang}, \bibinfo{person}{Forrester Cole}, \bibinfo{person}{Richard Tucker},
  {and} \bibinfo{person}{Noah Snavely}.} \bibinfo{year}{2022}\natexlab{c}.
\newblock \showarticletitle{DynIBaR: Neural Dynamic Image-Based Rendering}.
\newblock \bibinfo{journal}{\emph{arXiv preprint arXiv:2211.11082}}
  (\bibinfo{year}{2022}).
\newblock


\bibitem[Lin et~al\mbox{.}(2022)]%
        {lin2022robust}
\bibfield{author}{\bibinfo{person}{Shanchuan Lin}, \bibinfo{person}{Linjie
  Yang}, \bibinfo{person}{Imran Saleemi}, {and} \bibinfo{person}{Soumyadip
  Sengupta}.} \bibinfo{year}{2022}\natexlab{}.
\newblock \showarticletitle{Robust high-resolution video matting with temporal
  guidance}. In \bibinfo{booktitle}{\emph{Proceedings of the IEEE/CVF Winter
  Conference on Applications of Computer Vision}}. \bibinfo{pages}{238--247}.
\newblock


\bibitem[Liu et~al\mbox{.}(2022)]%
        {liu2022devrf}
\bibfield{author}{\bibinfo{person}{Jia-Wei Liu}, \bibinfo{person}{Yan-Pei Cao},
  \bibinfo{person}{Weijia Mao}, \bibinfo{person}{Wenqiao Zhang},
  \bibinfo{person}{David~Junhao Zhang}, \bibinfo{person}{Jussi Keppo},
  \bibinfo{person}{Ying Shan}, \bibinfo{person}{Xiaohu Qie}, {and}
  \bibinfo{person}{Mike~Zheng Shou}.} \bibinfo{year}{2022}\natexlab{}.
\newblock \showarticletitle{DeVRF: Fast Deformable Voxel Radiance Fields for
  Dynamic Scenes}.
\newblock \bibinfo{journal}{\emph{Advances in Neural Information Processing
  Systems}}.
\newblock


\bibitem[Liu et~al\mbox{.}(2021)]%
        {liu2021neural}
\bibfield{author}{\bibinfo{person}{Lingjie Liu}, \bibinfo{person}{Marc
  Habermann}, \bibinfo{person}{Viktor Rudnev}, \bibinfo{person}{Kripasindhu
  Sarkar}, \bibinfo{person}{Jiatao Gu}, {and} \bibinfo{person}{Christian
  Theobalt}.} \bibinfo{year}{2021}\natexlab{}.
\newblock \showarticletitle{Neural actor: Neural free-view synthesis of human
  actors with pose control}.
\newblock \bibinfo{journal}{\emph{ACM Transactions on Graphics (TOG)}}
  \bibinfo{volume}{40}, \bibinfo{number}{6} (\bibinfo{year}{2021}),
  \bibinfo{pages}{1--16}.
\newblock


\bibitem[Lombardi et~al\mbox{.}(2019)]%
        {neural_volumes}
\bibfield{author}{\bibinfo{person}{Stephen Lombardi}, \bibinfo{person}{Tomas
  Simon}, \bibinfo{person}{Jason Saragih}, \bibinfo{person}{Gabriel Schwartz},
  \bibinfo{person}{Andreas Lehrmann}, {and} \bibinfo{person}{Yaser Sheikh}.}
  \bibinfo{year}{2019}\natexlab{}.
\newblock \showarticletitle{Neural Volumes: Learning Dynamic Renderable Volumes
  from Images}.
\newblock \bibinfo{journal}{\emph{ACM Trans. Graph.}} \bibinfo{volume}{38},
  \bibinfo{number}{4}, Article \bibinfo{articleno}{65} (\bibinfo{date}{July}
  \bibinfo{year}{2019}), \bibinfo{numpages}{14}~pages.
\newblock


\bibitem[Lombardi et~al\mbox{.}(2021)]%
        {lombardi2021mixture}
\bibfield{author}{\bibinfo{person}{Stephen Lombardi}, \bibinfo{person}{Tomas
  Simon}, \bibinfo{person}{Gabriel Schwartz}, \bibinfo{person}{Michael
  Zollhoefer}, \bibinfo{person}{Yaser Sheikh}, {and} \bibinfo{person}{Jason
  Saragih}.} \bibinfo{year}{2021}\natexlab{}.
\newblock \showarticletitle{Mixture of volumetric primitives for efficient
  neural rendering}.
\newblock \bibinfo{journal}{\emph{ACM Transactions on Graphics (TOG)}}
  \bibinfo{volume}{40}, \bibinfo{number}{4} (\bibinfo{year}{2021}),
  \bibinfo{pages}{1--13}.
\newblock


\bibitem[Loper et~al\mbox{.}(2015)]%
        {Loper2015SMPLAS}
\bibfield{author}{\bibinfo{person}{Matthew Loper}, \bibinfo{person}{Naureen
  Mahmood}, \bibinfo{person}{Javier Romero}, \bibinfo{person}{Gerard
  Pons-Moll}, {and} \bibinfo{person}{Michael~J. Black}.}
  \bibinfo{year}{2015}\natexlab{}.
\newblock \showarticletitle{SMPL: a skinned multi-person linear model}.
\newblock \bibinfo{journal}{\emph{ACM Trans. Graph.}}  \bibinfo{volume}{34}
  (\bibinfo{year}{2015}), \bibinfo{pages}{248:1--248:16}.
\newblock


\bibitem[Luo et~al\mbox{.}(2022)]%
        {artemis}
\bibfield{author}{\bibinfo{person}{Haimin Luo}, \bibinfo{person}{Teng Xu},
  \bibinfo{person}{Yuheng Jiang}, \bibinfo{person}{Chenglin Zhou},
  \bibinfo{person}{Qiwei Qiu}, \bibinfo{person}{Yingliang Zhang},
  \bibinfo{person}{Wei Yang}, \bibinfo{person}{Lan Xu}, {and}
  \bibinfo{person}{Jingyi Yu}.} \bibinfo{year}{2022}\natexlab{}.
\newblock \showarticletitle{Artemis: Articulated Neural Pets with Appearance
  and Motion Synthesis}.
\newblock \bibinfo{journal}{\emph{ACM Trans. Graph.}} \bibinfo{volume}{41},
  \bibinfo{number}{4}, Article \bibinfo{articleno}{164} (\bibinfo{date}{jul}
  \bibinfo{year}{2022}), \bibinfo{numpages}{19}~pages.
\newblock
\showISSN{0730-0301}
\urldef\tempurl%
\url{https://doi.org/10.1145/3528223.3530086}
\showDOI{\tempurl}


\bibitem[Max(1995)]%
        {max1995optical}
\bibfield{author}{\bibinfo{person}{Nelson Max}.}
  \bibinfo{year}{1995}\natexlab{}.
\newblock \showarticletitle{Optical models for direct volume rendering}.
\newblock \bibinfo{journal}{\emph{IEEE Transactions on Visualization and
  Computer Graphics}} \bibinfo{volume}{1}, \bibinfo{number}{2}
  (\bibinfo{year}{1995}), \bibinfo{pages}{99--108}.
\newblock


\bibitem[Mehta et~al\mbox{.}(2017)]%
        {mehta2017monocular}
\bibfield{author}{\bibinfo{person}{Dushyant Mehta}, \bibinfo{person}{Helge
  Rhodin}, \bibinfo{person}{Dan Casas}, \bibinfo{person}{Pascal Fua},
  \bibinfo{person}{Oleksandr Sotnychenko}, \bibinfo{person}{Weipeng Xu}, {and}
  \bibinfo{person}{Christian Theobalt}.} \bibinfo{year}{2017}\natexlab{}.
\newblock \showarticletitle{Monocular 3d human pose estimation in the wild
  using improved cnn supervision}. In \bibinfo{booktitle}{\emph{2017
  international conference on 3D vision (3DV)}}. IEEE,
  \bibinfo{pages}{506--516}.
\newblock


\bibitem[Mescheder et~al\mbox{.}(2019)]%
        {mescheder2019occupancy}
\bibfield{author}{\bibinfo{person}{Lars Mescheder}, \bibinfo{person}{Michael
  Oechsle}, \bibinfo{person}{Michael Niemeyer}, \bibinfo{person}{Sebastian
  Nowozin}, {and} \bibinfo{person}{Andreas Geiger}.}
  \bibinfo{year}{2019}\natexlab{}.
\newblock \showarticletitle{Occupancy networks: Learning 3d reconstruction in
  function space}. In \bibinfo{booktitle}{\emph{Proceedings of the IEEE/CVF
  conference on computer vision and pattern recognition}}.
  \bibinfo{pages}{4460--4470}.
\newblock


\bibitem[Mildenhall et~al\mbox{.}(2020)]%
        {mildenhall2020nerf}
\bibfield{author}{\bibinfo{person}{Ben Mildenhall}, \bibinfo{person}{Pratul~P.
  Srinivasan}, \bibinfo{person}{Matthew Tancik}, \bibinfo{person}{Jonathan~T.
  Barron}, \bibinfo{person}{Ravi Ramamoorthi}, {and} \bibinfo{person}{Ren Ng}.}
  \bibinfo{year}{2020}\natexlab{}.
\newblock \showarticletitle{NeRF: Representing Scenes as Neural Radiance Fields
  for View Synthesis}. In \bibinfo{booktitle}{\emph{ECCV}}.
\newblock


\bibitem[M\"uller(2021)]%
        {tiny-cuda-nn}
\bibfield{author}{\bibinfo{person}{Thomas M\"uller}.}
  \bibinfo{year}{2021}\natexlab{}.
\newblock \bibinfo{booktitle}{\emph{{tiny-cuda-nn}}}.
\newblock
\urldef\tempurl%
\url{https://github.com/NVlabs/tiny-cuda-nn}
\showURL{%
\tempurl}
\newblock
\shownote{Accessed: 2022-10-21}.


\bibitem[M\"uller et~al\mbox{.}(2022)]%
        {mueller2022instant}
\bibfield{author}{\bibinfo{person}{Thomas M\"uller}, \bibinfo{person}{Alex
  Evans}, \bibinfo{person}{Christoph Schied}, {and} \bibinfo{person}{Alexander
  Keller}.} \bibinfo{year}{2022}\natexlab{}.
\newblock \showarticletitle{Instant Neural Graphics Primitives with a
  Multiresolution Hash Encoding}.
\newblock \bibinfo{journal}{\emph{ACM Trans. Graph.}} \bibinfo{volume}{41},
  \bibinfo{number}{4}, Article \bibinfo{articleno}{102} (\bibinfo{date}{July}
  \bibinfo{year}{2022}), \bibinfo{numpages}{15}~pages.
\newblock
\urldef\tempurl%
\url{https://doi.org/10.1145/3528223.3530127}
\showDOI{\tempurl}


\bibitem[Noguchi et~al\mbox{.}(2021)]%
        {noguchi2021neural}
\bibfield{author}{\bibinfo{person}{Atsuhiro Noguchi}, \bibinfo{person}{Xiao
  Sun}, \bibinfo{person}{Stephen Lin}, {and} \bibinfo{person}{Tatsuya Harada}.}
  \bibinfo{year}{2021}\natexlab{}.
\newblock \showarticletitle{Neural articulated radiance field}. In
  \bibinfo{booktitle}{\emph{Proceedings of the IEEE/CVF International
  Conference on Computer Vision}}. \bibinfo{pages}{5762--5772}.
\newblock


\bibitem[Park et~al\mbox{.}(2019)]%
        {park2019deepsdf}
\bibfield{author}{\bibinfo{person}{Jeong~Joon Park}, \bibinfo{person}{Peter
  Florence}, \bibinfo{person}{Julian Straub}, \bibinfo{person}{Richard
  Newcombe}, {and} \bibinfo{person}{Steven Lovegrove}.}
  \bibinfo{year}{2019}\natexlab{}.
\newblock \showarticletitle{Deepsdf: Learning continuous signed distance
  functions for shape representation}. In \bibinfo{booktitle}{\emph{Proceedings
  of the IEEE/CVF conference on computer vision and pattern recognition}}.
  \bibinfo{pages}{165--174}.
\newblock


\bibitem[Park et~al\mbox{.}(2021a)]%
        {park2021nerfies}
\bibfield{author}{\bibinfo{person}{Keunhong Park}, \bibinfo{person}{Utkarsh
  Sinha}, \bibinfo{person}{Jonathan~T Barron}, \bibinfo{person}{Sofien
  Bouaziz}, \bibinfo{person}{Dan~B Goldman}, \bibinfo{person}{Steven~M Seitz},
  {and} \bibinfo{person}{Ricardo Martin-Brualla}.}
  \bibinfo{year}{2021}\natexlab{a}.
\newblock \showarticletitle{Nerfies: Deformable neural radiance fields}. In
  \bibinfo{booktitle}{\emph{Proceedings of the IEEE/CVF International
  Conference on Computer Vision}}. \bibinfo{pages}{5865--5874}.
\newblock


\bibitem[Park et~al\mbox{.}(2021b)]%
        {park2021hypernerf}
\bibfield{author}{\bibinfo{person}{Keunhong Park}, \bibinfo{person}{Utkarsh
  Sinha}, \bibinfo{person}{Peter Hedman}, \bibinfo{person}{Jonathan~T Barron},
  \bibinfo{person}{Sofien Bouaziz}, \bibinfo{person}{Dan~B Goldman},
  \bibinfo{person}{Ricardo Martin-Brualla}, {and} \bibinfo{person}{Steven~M
  Seitz}.} \bibinfo{year}{2021}\natexlab{b}.
\newblock \showarticletitle{Hypernerf: A higher-dimensional representation for
  topologically varying neural radiance fields}.
\newblock \bibinfo{journal}{\emph{ACM Trans. Graph.}} \bibinfo{volume}{40},
  \bibinfo{number}{6} (\bibinfo{date}{dec} \bibinfo{year}{2021}).
\newblock


\bibitem[Peng et~al\mbox{.}(2021)]%
        {peng2021neural}
\bibfield{author}{\bibinfo{person}{Sida Peng}, \bibinfo{person}{Yuanqing
  Zhang}, \bibinfo{person}{Yinghao Xu}, \bibinfo{person}{Qianqian Wang},
  \bibinfo{person}{Qing Shuai}, \bibinfo{person}{Hujun Bao}, {and}
  \bibinfo{person}{Xiaowei Zhou}.} \bibinfo{year}{2021}\natexlab{}.
\newblock \showarticletitle{Neural body: Implicit neural representations with
  structured latent codes for novel view synthesis of dynamic humans}. In
  \bibinfo{booktitle}{\emph{Proceedings of the IEEE/CVF Conference on Computer
  Vision and Pattern Recognition}}. \bibinfo{pages}{9054--9063}.
\newblock


\bibitem[Pumarola et~al\mbox{.}(2021)]%
        {pumarola2021d}
\bibfield{author}{\bibinfo{person}{Albert Pumarola}, \bibinfo{person}{Enric
  Corona}, \bibinfo{person}{Gerard Pons-Moll}, {and} \bibinfo{person}{Francesc
  Moreno-Noguer}.} \bibinfo{year}{2021}\natexlab{}.
\newblock \showarticletitle{D-nerf: Neural radiance fields for dynamic scenes}.
  In \bibinfo{booktitle}{\emph{Proceedings of the IEEE/CVF Conference on
  Computer Vision and Pattern Recognition}}. \bibinfo{pages}{10318--10327}.
\newblock


\bibitem[Shao et~al\mbox{.}(2022)]%
        {shao2022tensor4d}
\bibfield{author}{\bibinfo{person}{Ruizhi Shao}, \bibinfo{person}{Zerong
  Zheng}, \bibinfo{person}{Hanzhang Tu}, \bibinfo{person}{Boning Liu},
  \bibinfo{person}{Hongwen Zhang}, {and} \bibinfo{person}{Yebin Liu}.}
  \bibinfo{year}{2022}\natexlab{}.
\newblock \showarticletitle{Tensor4D: Efficient Neural 4D Decomposition for
  High-fidelity Dynamic Reconstruction and Rendering}.
\newblock \bibinfo{journal}{\emph{arXiv preprint arXiv:2211.11610}}
  (\bibinfo{year}{2022}).
\newblock


\bibitem[Song et~al\mbox{.}(2022)]%
        {song2022nerfplayer}
\bibfield{author}{\bibinfo{person}{Liangchen Song}, \bibinfo{person}{Anpei
  Chen}, \bibinfo{person}{Zhong Li}, \bibinfo{person}{Zhang Chen},
  \bibinfo{person}{Lele Chen}, \bibinfo{person}{Junsong Yuan},
  \bibinfo{person}{Yi Xu}, {and} \bibinfo{person}{Andreas Geiger}.}
  \bibinfo{year}{2022}\natexlab{}.
\newblock \showarticletitle{NeRFPlayer: A Streamable Dynamic Scene
  Representation with Decomposed Neural Radiance Fields}.
\newblock \bibinfo{journal}{\emph{arXiv preprint arXiv:2210.15947}}
  (\bibinfo{year}{2022}).
\newblock


\bibitem[Starck and Hilton(2007)]%
        {starck2007surface}
\bibfield{author}{\bibinfo{person}{Jonathan Starck} {and}
  \bibinfo{person}{Adrian Hilton}.} \bibinfo{year}{2007}\natexlab{}.
\newblock \showarticletitle{Surface capture for performance-based animation}.
\newblock \bibinfo{journal}{\emph{IEEE computer graphics and applications}}
  \bibinfo{volume}{27}, \bibinfo{number}{3} (\bibinfo{year}{2007}),
  \bibinfo{pages}{21--31}.
\newblock


\bibitem[Su et~al\mbox{.}(2021)]%
        {su2021nerf}
\bibfield{author}{\bibinfo{person}{Shih-Yang Su}, \bibinfo{person}{Frank Yu},
  \bibinfo{person}{Michael Zollh{\"o}fer}, {and} \bibinfo{person}{Helge
  Rhodin}.} \bibinfo{year}{2021}\natexlab{}.
\newblock \showarticletitle{A-nerf: Articulated neural radiance fields for
  learning human shape, appearance, and pose}.
\newblock \bibinfo{journal}{\emph{Advances in Neural Information Processing
  Systems}}  \bibinfo{volume}{34} (\bibinfo{year}{2021}),
  \bibinfo{pages}{12278--12291}.
\newblock


\bibitem[Sun et~al\mbox{.}(2022)]%
        {SunSC22}
\bibfield{author}{\bibinfo{person}{Cheng Sun}, \bibinfo{person}{Min Sun}, {and}
  \bibinfo{person}{Hwann{-}Tzong Chen}.} \bibinfo{year}{2022}\natexlab{}.
\newblock \showarticletitle{Direct Voxel Grid Optimization: Super-fast
  Convergence for Radiance Fields Reconstruction}. In
  \bibinfo{booktitle}{\emph{CVPR}}.
\newblock


\bibitem[Tang et~al\mbox{.}(2022)]%
        {ccnerf}
\bibfield{author}{\bibinfo{person}{Jiaxiang Tang}, \bibinfo{person}{Xiaokang
  Chen}, \bibinfo{person}{Jingbo Wang}, {and} \bibinfo{person}{Gang Zeng}.}
  \bibinfo{year}{2022}\natexlab{}.
\newblock \showarticletitle{Compressible-composable NeRF via Rank-residual
  Decomposition}.
\newblock \bibinfo{journal}{\emph{arXiv preprint arXiv:2205.14870}}
  (\bibinfo{year}{2022}).
\newblock


\bibitem[Wang et~al\mbox{.}(2022b)]%
        {wang2022fourier}
\bibfield{author}{\bibinfo{person}{Liao Wang}, \bibinfo{person}{Jiakai Zhang},
  \bibinfo{person}{Xinhang Liu}, \bibinfo{person}{Fuqiang Zhao},
  \bibinfo{person}{Yanshun Zhang}, \bibinfo{person}{Yingliang Zhang},
  \bibinfo{person}{Minye Wu}, \bibinfo{person}{Jingyi Yu}, {and}
  \bibinfo{person}{Lan Xu}.} \bibinfo{year}{2022}\natexlab{b}.
\newblock \showarticletitle{Fourier PlenOctrees for Dynamic Radiance Field
  Rendering in Real-time}. In \bibinfo{booktitle}{\emph{Proceedings of the
  IEEE/CVF Conference on Computer Vision and Pattern Recognition}}.
  \bibinfo{pages}{13524--13534}.
\newblock


\bibitem[Wang et~al\mbox{.}(2022a)]%
        {wang2022arah}
\bibfield{author}{\bibinfo{person}{Shaofei Wang}, \bibinfo{person}{Katja
  Schwarz}, \bibinfo{person}{Andreas Geiger}, {and} \bibinfo{person}{Siyu
  Tang}.} \bibinfo{year}{2022}\natexlab{a}.
\newblock \showarticletitle{Arah: Animatable volume rendering of articulated
  human sdfs}. In \bibinfo{booktitle}{\emph{Computer Vision--ECCV 2022: 17th
  European Conference, Tel Aviv, Israel, October 23--27, 2022, Proceedings,
  Part XXXII}}. Springer, \bibinfo{pages}{1--19}.
\newblock


\bibitem[Wang et~al\mbox{.}(2021)]%
        {hybrid_nerf}
\bibfield{author}{\bibinfo{person}{Ziyan Wang}, \bibinfo{person}{Timur
  Bagautdinov}, \bibinfo{person}{Stephen Lombardi}, \bibinfo{person}{Tomas
  Simon}, \bibinfo{person}{Jason Saragih}, \bibinfo{person}{Jessica Hodgins},
  {and} \bibinfo{person}{Michael Zollhofer}.} \bibinfo{year}{2021}\natexlab{}.
\newblock \showarticletitle{Learning Compositional Radiance Fields of Dynamic
  Human Heads}. In \bibinfo{booktitle}{\emph{Proceedings of the IEEE/CVF
  Conference on Computer Vision and Pattern Recognition (CVPR)}}.
  \bibinfo{pages}{5704--5713}.
\newblock


\bibitem[Wang et~al\mbox{.}(2004)]%
        {wang2004image}
\bibfield{author}{\bibinfo{person}{Zhou Wang}, \bibinfo{person}{Alan~C Bovik},
  \bibinfo{person}{Hamid~R Sheikh}, {and} \bibinfo{person}{Eero~P Simoncelli}.}
  \bibinfo{year}{2004}\natexlab{}.
\newblock \showarticletitle{Image quality assessment: from error visibility to
  structural similarity}.
\newblock \bibinfo{journal}{\emph{IEEE transactions on image processing}}
  \bibinfo{volume}{13}, \bibinfo{number}{4} (\bibinfo{year}{2004}),
  \bibinfo{pages}{600--612}.
\newblock


\bibitem[Xu et~al\mbox{.}(2022)]%
        {xu2022surface}
\bibfield{author}{\bibinfo{person}{Tianhan Xu}, \bibinfo{person}{Yasuhiro
  Fujita}, {and} \bibinfo{person}{Eiichi Matsumoto}.}
  \bibinfo{year}{2022}\natexlab{}.
\newblock \showarticletitle{Surface-Aligned Neural Radiance Fields for
  Controllable 3D Human Synthesis}. In \bibinfo{booktitle}{\emph{Proceedings of
  the IEEE/CVF Conference on Computer Vision and Pattern Recognition}}.
  \bibinfo{pages}{15883--15892}.
\newblock


\bibitem[Yariv et~al\mbox{.}(2020)]%
        {yariv2020multiview}
\bibfield{author}{\bibinfo{person}{Lior Yariv}, \bibinfo{person}{Yoni Kasten},
  \bibinfo{person}{Dror Moran}, \bibinfo{person}{Meirav Galun},
  \bibinfo{person}{Matan Atzmon}, \bibinfo{person}{Basri Ronen}, {and}
  \bibinfo{person}{Yaron Lipman}.} \bibinfo{year}{2020}\natexlab{}.
\newblock \showarticletitle{Multiview Neural Surface Reconstruction by
  Disentangling Geometry and Appearance}.
\newblock \bibinfo{journal}{\emph{Advances in Neural Information Processing
  Systems}}  \bibinfo{volume}{33} (\bibinfo{year}{2020}).
\newblock


\bibitem[Yu et~al\mbox{.}(2021)]%
        {yu2021plenoctrees}
\bibfield{author}{\bibinfo{person}{Alex Yu}, \bibinfo{person}{Ruilong Li},
  \bibinfo{person}{Matthew Tancik}, \bibinfo{person}{Hao Li},
  \bibinfo{person}{Ren Ng}, {and} \bibinfo{person}{Angjoo Kanazawa}.}
  \bibinfo{year}{2021}\natexlab{}.
\newblock \showarticletitle{Plenoctrees for real-time rendering of neural
  radiance fields}. In \bibinfo{booktitle}{\emph{Proceedings of the IEEE/CVF
  International Conference on Computer Vision}}. \bibinfo{pages}{5752--5761}.
\newblock


\bibitem[Zhang et~al\mbox{.}(2022)]%
        {zhang2022neuvv}
\bibfield{author}{\bibinfo{person}{Jiakai Zhang}, \bibinfo{person}{Liao Wang},
  \bibinfo{person}{Xinhang Liu}, \bibinfo{person}{Fuqiang Zhao},
  \bibinfo{person}{Minzhang Li}, \bibinfo{person}{Haizhao Dai},
  \bibinfo{person}{Boyuan Zhang}, \bibinfo{person}{Wei Yang},
  \bibinfo{person}{Lan Xu}, {and} \bibinfo{person}{Jingyi Yu}.}
  \bibinfo{year}{2022}\natexlab{}.
\newblock \showarticletitle{NeuVV: Neural Volumetric Videos with Immersive
  Rendering and Editing}.
\newblock \bibinfo{journal}{\emph{arXiv preprint arXiv:2202.06088}}
  (\bibinfo{year}{2022}).
\newblock


\bibitem[Zhang et~al\mbox{.}(2018)]%
        {zhang2018perceptual}
\bibfield{author}{\bibinfo{person}{Richard Zhang}, \bibinfo{person}{Phillip
  Isola}, \bibinfo{person}{Alexei~A Efros}, \bibinfo{person}{Eli Shechtman},
  {and} \bibinfo{person}{Oliver Wang}.} \bibinfo{year}{2018}\natexlab{}.
\newblock \showarticletitle{The Unreasonable Effectiveness of Deep Features as
  a Perceptual Metric}. In \bibinfo{booktitle}{\emph{CVPR}}.
\newblock


\bibitem[Zhao et~al\mbox{.}(2022)]%
        {Zhao2022HumanPM}
\bibfield{author}{\bibinfo{person}{Fuqiang Zhao}, \bibinfo{person}{Yuheng
  Jiang}, \bibinfo{person}{Kaixin Yao}, \bibinfo{person}{Jiakai Zhang},
  \bibinfo{person}{Liao Wang}, \bibinfo{person}{Haizhao Dai},
  \bibinfo{person}{Yuhui Zhong}, \bibinfo{person}{Yingliang Zhang},
  \bibinfo{person}{Minye Wu}, \bibinfo{person}{Lan Xu}, {and}
  \bibinfo{person}{Jingyi Yu}.} \bibinfo{year}{2022}\natexlab{}.
\newblock \showarticletitle{Human Performance Modeling and Rendering via Neural
  Animated Mesh}. In \bibinfo{booktitle}{\emph{ACM Transactions on Graphics
  (TOG)}}, Vol.~\bibinfo{volume}{41}. \bibinfo{pages}{1 -- 17}.
\newblock


\bibitem[Zheng et~al\mbox{.}(2022)]%
        {zheng2022structured}
\bibfield{author}{\bibinfo{person}{Zerong Zheng}, \bibinfo{person}{Han Huang},
  \bibinfo{person}{Tao Yu}, \bibinfo{person}{Hongwen Zhang},
  \bibinfo{person}{Yandong Guo}, {and} \bibinfo{person}{Yebin Liu}.}
  \bibinfo{year}{2022}\natexlab{}.
\newblock \showarticletitle{Structured local radiance fields for human avatar
  modeling}. In \bibinfo{booktitle}{\emph{Proceedings of the IEEE/CVF
  Conference on Computer Vision and Pattern Recognition}}.
  \bibinfo{pages}{15893--15903}.
\newblock


\bibitem[Zheng et~al\mbox{.}(2019)]%
        {zheng2019deephuman}
\bibfield{author}{\bibinfo{person}{Zerong Zheng}, \bibinfo{person}{Tao Yu},
  \bibinfo{person}{Yixuan Wei}, \bibinfo{person}{Qionghai Dai}, {and}
  \bibinfo{person}{Yebin Liu}.} \bibinfo{year}{2019}\natexlab{}.
\newblock \showarticletitle{Deephuman: 3d human reconstruction from a single
  image}. In \bibinfo{booktitle}{\emph{Proceedings of the IEEE/CVF
  International Conference on Computer Vision}}. \bibinfo{pages}{7739--7749}.
\newblock


\end{thebibliography}

\end{document}


\title{HumanRF: Supplementary Material}

\author{Mustafa Işık}
\email{mustafa.isik@synthesia.io}
\affiliation{
  \institution{Synthesia}
  \city{Munich}
  \country{Germany}
}
\author{Martin Rünz}
\email{martin@synthesia.io}
\affiliation{
  \institution{Synthesia}
  \city{Munich}
  \country{Germany}
}
\author{Markos Georgopoulos}
\email{markos@synthesia.io}
\affiliation{
  \institution{Synthesia}
  \city{London}
  \country{United Kingdom}
}
\author{Taras Khakhulin}
\email{taras.khakhulin@synthesia.io}
\affiliation{
  \institution{Synthesia}
  \city{London}
  \country{United Kingdom}
}
\author{Jonathan Starck}
\email{jon@synthesia.io}
\affiliation{
  \institution{Synthesia}
  \city{London}
  \country{United Kingdom}
}
\author{Lourdes Agapito}
\email{l.agapito@cs.ucl.ac.uk}
\affiliation{
  \institution{University College London}
  \city{London}
  \country{United Kingdom}
}
\author{Matthias Nießner}
\email{niessner@tum.de}
\affiliation{
  \institution{Technical University of Munich}
  \city{Munich}
  \country{Germany}
}

\renewcommand{\shortauthors}{Işık et al.}

\maketitle

\section{Implementation Details}
Here, we discuss our implementation tricks that enable neural rendering of terabytes of multi-view data.

\subsection{\methodname{}}
Our method is implemented in PyTorch \cite{paszke2019pytorch} and in CUDA for some of the parts that require performance. We use \emph{Tiny CUDA neural networks} framework \cite{tiny-cuda-nn} to create four 3D hash grid representations. To reduce the amount of intermediate memory usage during training and improve performance, we write a CUDA kernel that samples from four 1D dense grids and compose the results with the sampled features from the hash grid. Furthermore, we utilize some of the functionalities from \emph{torch-ngp} \cite{torch-ngp} and \emph{NerfAcc} \cite{li2022nerfacc}.

\begin{figure}
    \includegraphics[width=\linewidth]{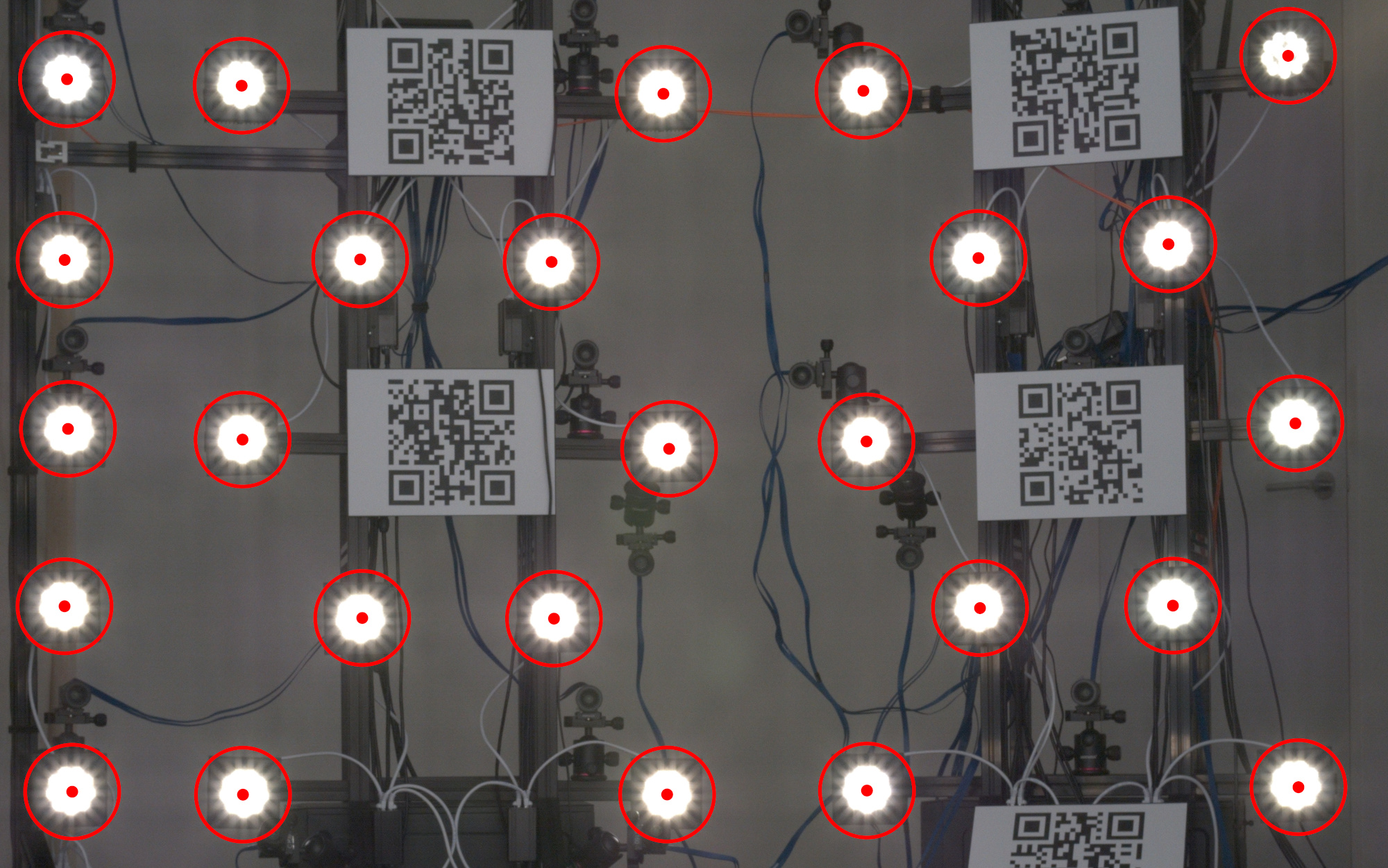}
    \caption{\textbf{Light annotations.} Light bloom can have a significant impact on photometric consistency. Light annotations are used to avoid using these regions during training.}
    \label{fig:light-annotations}
\end{figure}

In our high resolution video results on the supplemental video, we make use of per-camera embeddings \cite{martinbrualla2020nerfw} which are concatenated to the input of the radiance MLP. This helps removing the brightness and lighting inconsistencies that arise for some cameras. In addition, we filter the light bloom effect based on the light source annotations shown in Fig.~\ref{fig:light-annotations}. That is, we do not sample rays from the annotated circular regions to prevent using pixels that have light diffused into it. We note that this modified version is not used in any of the comparisons made in the main paper or supplementary material for fairness. It is simply used for the stand-alone 12MP video results.

\subsection{\datasetname{} Data Loader}
Considering the terabytes of data we need to deal with, storing training data in memory, or preparing batches in a pre-computation stage and saving it to the file system is impractical in terms of memory or hard disk requirements. The idea behind our data loader is to bypass the reading and writing large chunks of data by sampling batch of rays on the fly from as many images as possible across different cameras and time frames. To do this, we define a pool of images, and randomly sample from this pool continuously in the main thread while another thread is working in the background to replace the images in the pool. By replacing and sampling concurrently, we use only a modest amount of GPU and CPU memory to accommodate the pool. Also, we implement custom CUDA kernels to make use of the occupancy grids (that are initialized from masks) to skip empty space during ray sampling. This speeds up the rendering significantly, and increases the effective capacity of the model because the empty space does not have to be modeled.

\subsection{Training}
We use ADAM optimizer \cite{kingma2014adam} with the initial learning rate of $10^{-2}$. We decay this learning rate to $5\cdot10^{-3}$ until the end of each training. We utilize FP16 operations for fast training and inference. For experiments with $4\times$ downscaled input, we train for $N\times1000$ iterations where $N$ depicts the number of frames in the training sequence. On average, our implementation performs 8 to 12 training iterations per second on a single NVIDIA GeForce RTX 3090 with 24GB memory. This corresponds to roughly a day of compute for 1000-frame sequences. On the other hand, we train for $N\times2500$ iterations for full-resolution trainings.

We define the training batch size in terms of maximum number of samples over all the training rays in a batch. We start by sampling 8192 rays per batch, and dynamically adjust number of rays such that the maximum number of samples is reached every iteration. This lets us achieve high GPU utilization during training. We set the maximum number of samples to $640\text{K}$, $576\text{K}$ and $512\text{K}$ for $4\times$ downscaled input, $2\times$ downscaled input and full-resolution input, respectively.

As our method partitions a given sequence into segments, it is possible to scale the training to thousands of frames. This is because we sample rays across fixed number of time frames, which we set to $8$ for all our experiments. Therefore, in the worst case scenario, only 8 different segments need to live in the GPU memory. On average, our segments have a size of 12 which would translate to around $64$ million parameters (256MB) that need to reside in the GPU memory at a time instance on average.

\section{Evaluation}
In this section, we clarify the details of camera and frame configurations used during training, validation and testing. In addition, we explain how the metrics are calculated to generate numerical results.

\subsection{Evaluation Protocol}

In the following, we describe the evaluation protocol used for our baseline comparison experiments. The dataset was split into 4 disjunct sets of cameras that are listed by their 1-based index:

\begin{itemize}
\item 124 training cameras: 2,   3,   4,   5,   6,   7,   8,   9,  10,  12,  13,  15,  16,
        17,  18,  19,  21,  22,  23,  24,  26,  27,  28,  29,  30,  32,
        33,  35,  36,  37,  38,  39,  40,  41,  42,  43,  46,  47,  48,
        49,  50,  52,  53,  54,  55,  56,  57,  59,  60,  61,  62,  63,
        66,  67,  68,  69,  70,  72,  73,  75,  76,  77,  78,  79,  80,
        81,  82,  83,  86,  87,  88,  89,  90,  92,  93,  94,  95,  96,
        97,  99, 100, 101, 102, 103, 106, 107, 108, 109, 110, 111, 112,
       113, 114, 116, 117, 119, 120, 121, 122, 123, 124, 125, 126, 128,
       131, 132, 133, 134, 135, 136, 139, 140, 141, 142, 143, 144, 149,
       150, 151, 152, 157, 158, 159, 160
\item 10 validation cameras: 11, 20, 34, 45, 51, 74, 84, 91, 105, 118
\item 13 per-frame test cameras: 1,  14,  25,  31,  44,  58,  64,  65,  71,  85,  98, 104, 115
\item 1 VMAF test camera: 127
\end{itemize}
When computing per-frame scores, we alternate the test cameras in the following order: $1,  64,  98,  31,  14,  71, 115,  25,  85,  44,  65, 104,  58$ and temporally subsample every fifth frame leading to frame-camera pairs such as $\{(1, 1), (6, 64), (11, 98), ...\}$. To reduce the computational burden to execute this comparison, we use one of the sequences per actor alternatingly, i.e. Actor1 Sequence1, Actor2 Sequence2, Actor3 Sequence1, Actor4 Sequence2, Actor5 Sequence1, Actor6 Sequence2, Actor7 Sequence1 and Actor8 Sequence2. Sequence1s contain moderate movements while Sequence2s contain stronger motion. PSNR scores are computed only on the foreground depicted by the ground truth masks while SSIM and LPIPS are computed by tightly cropping images to fit ground truth foreground masks. Finally, VMAF is computed on the video that is compiled by rendering every third frame from the hero camera (camera 127).

\subsection{Numerical Results of Input Resolution Experiment}

In Table~\ref{tab:resolution-vs-quality}, we provide additional results concerning the input resolution experiment we present in the main paper.

\subsection{Additional Numerical Results on Baseline Comparison}

\begin{figure*}
\captionsetup[subfigure]{labelformat=empty}
 \begin{subfigure}{0.47\linewidth}
    \centering
    \includegraphics[width=1\textwidth]{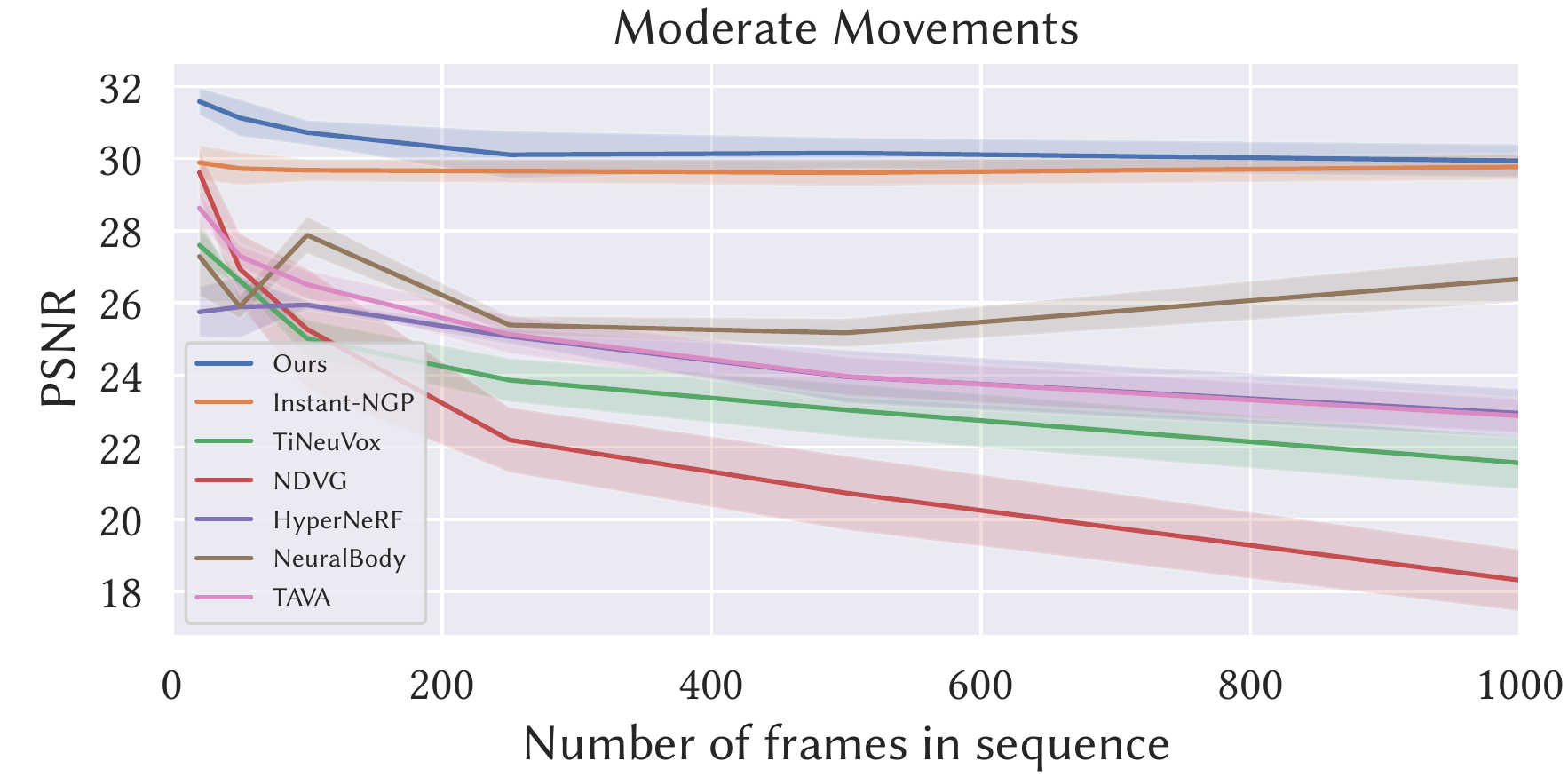}
\end{subfigure}
\hspace{0.1cm}
\begin{subfigure}{0.47\linewidth} 
    \centering
    \includegraphics[width=1\textwidth]{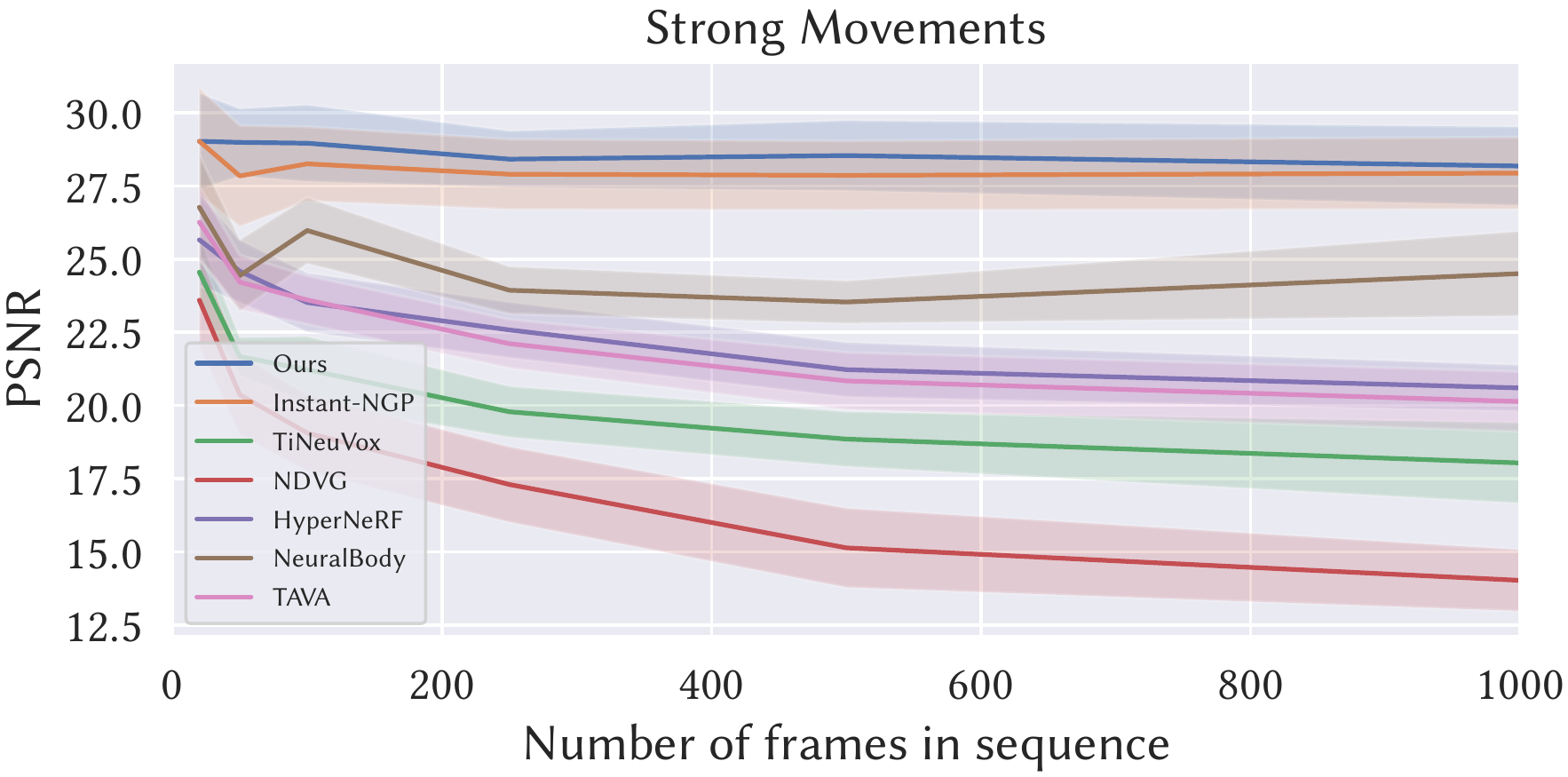}
\end{subfigure}
\caption{\textbf{Influence of increasing the sequence length.} Thanks to using adaptively-placed 4D segments, our method consistently outperforms the deformation-based baselines as they struggle to capture complex motion over long sequences. Although NeuralBody does not lose its representation power for long sequences, its overall quality is inferior to \methodname{}.}
\label{fig:nframes_seq}
\end{figure*}

We provide additional results over different motion complexities in Table~\ref{tab:frames-vs-quality-s1} and Table~\ref{tab:frames-vs-quality-s2}, and per sequence results. Moreover, we provide a plot in Fig.~\ref{fig:nframes_seq} to better illustrate the effect of increasing the sequence length.

\subsection{Visual Results on DFA Dataset}
\begin{figure*}
    \includegraphics[height=\textheight]{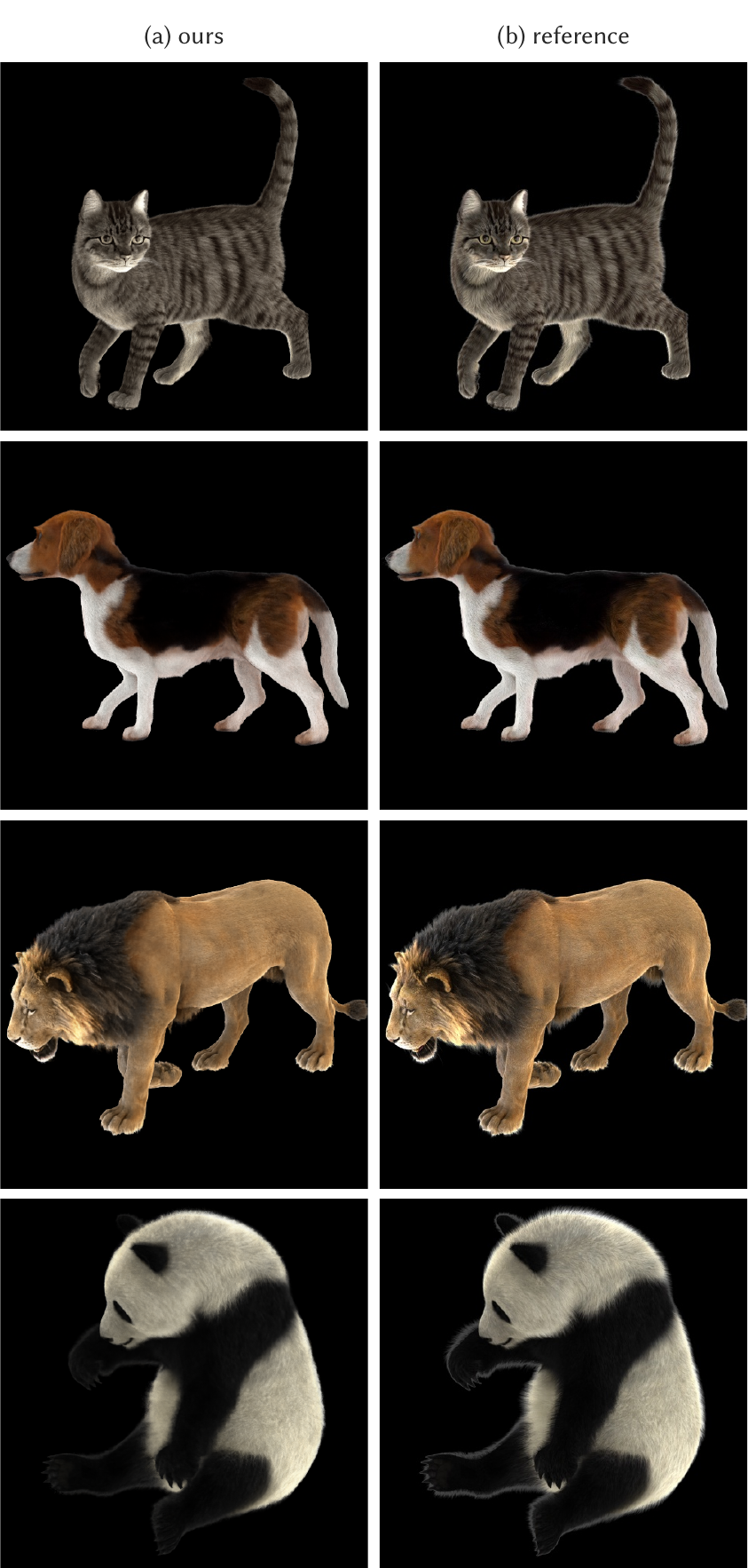}
    \caption{\textbf{Visual results from DFA dataset.}
    }
    \label{fig:dfa_figures}
\end{figure*}

In Fig.~\ref{fig:dfa_figures}, we present results of our method for four scenes we choose from DFA. We infer that \methodname{} can produce high-fidelity results for non-human subjects as well.

\section{Hex4D and TNGP formulations}
Following the notations we have used to define Equation 1 in the main paper, we define Hex4D formulation as follows:
\begin{equation}
\label{eq:hex4d}
\begin{split}
\featureTensor_{xyzt}(\reprPoint_{xyzt}) = \featureTensor_{xy}(\reprPoint_{xy}) & \odot \featureTensor_{zt}(\reprPoint_{zt})
\\
+ \featureTensor_{yz}(\reprPoint_{yz}) & \odot \featureTensor_{xt}(\reprPoint_{xt})
\\
+ \featureTensor_{xz}(\reprPoint_{xz}) & \odot \featureTensor_{yt}(\reprPoint_{yt})
\\
\end{split} \ ,
\end{equation}
where we represent six 2D planes ($\featureTensor_{xy}, \featureTensor_{yz}, \featureTensor_{xz}, \featureTensor_{zt},
\featureTensor_{xy},
\featureTensor_{yt}: \R^2 \mapsto \R^{\featureDim}$) using multi-resolution dense grids. Notice that HexPlanes~\cite{hexplanes} uses concatenation operation instead of addition as opposed to our formulation. However, for our experiments, we did not observe an improvement when using concatenation over addition.

On the other hand, tNGP simply uses a 4D hash grid to represent $\featureTensor_{xyzt}: \R^4 \mapsto \R^{\featureDim}$ without utilizing any kind of decomposition techniques.

\begin{table}
\caption{\textbf{Frame resolution vs representation quality.} Training \methodname{} at different input resolutions while rendering results at full resolution shows that additional details can be represented, and our method can make use of the extra information provided with the higher resolutions.}
\begin{tabular*}{\linewidth}{l@{\extracolsep{\fill}}cccc}
\toprule
Resolution & PSNR $\uparrow$ & LPIPS $\downarrow$ & SSIM $\uparrow$ & VMAF $\uparrow$ \\
\midrule
12MP    & 28.07    & 0.348     & 0.812    & 68.76    \\
12MP/$(2\times2)$  & 27.69    & 0.360     & 0.809    & 65.37    \\
12MP/$(4\times4)$ & 27.29    & 0.375     & 0.799    & 59.31    \\
\bottomrule
\end{tabular*}
\label{tab:resolution-vs-quality}
\end{table}

\clearpage
\newpage

\begin{table}
 \centering
 \resizebox{\linewidth}{!}{
\begin{tabular}{clcccccc}
\multicolumn{1}{l}{Method} & Metric & 20 & 50 & 100 & 250 & 500 & 1000 \\
\midrule
\multirow{4}{*}{Ours}& $\downarrow$ LPIPS & \bestCellColor{0.081} & \bestCellColor{0.090} & \bestCellColor{0.088} & \bestCellColor{0.095} & \bestCellColor{0.100} & \secondBestCellColor{0.103} \\
& $\uparrow$ PSNR & \bestCellColor{31.58} & \bestCellColor{31.12} & \bestCellColor{30.72} & \bestCellColor{30.10} & \bestCellColor{30.15} & \bestCellColor{29.93} \\
& $\uparrow$ SSIM & \bestCellColor{0.933} & \bestCellColor{0.930} & \bestCellColor{0.932} & \bestCellColor{0.927} & \bestCellColor{0.924} & \bestCellColor{0.920} \\
& $\uparrow$ VMAF & \bestCellColor{79.13} & \bestCellColor{79.49} & \bestCellColor{81.66} & \bestCellColor{81.93} & \bestCellColor{82.05} & \bestCellColor{83.15} \\
\midrule
\multirow{4}{*}{Instant-NGP}& $\downarrow$ LPIPS & \secondBestCellColor{0.100} & \secondBestCellColor{0.096} & \secondBestCellColor{0.099} & \secondBestCellColor{0.100} & \secondBestCellColor{0.100} & \bestCellColor{0.099} \\
& $\uparrow$ PSNR & \secondBestCellColor{29.88} & \secondBestCellColor{29.72} & \secondBestCellColor{29.67} & \secondBestCellColor{29.65} & \secondBestCellColor{29.60} & \secondBestCellColor{29.76} \\
& $\uparrow$ SSIM & \secondBestCellColor{0.884} & \secondBestCellColor{0.910} & \secondBestCellColor{0.910} & \secondBestCellColor{0.910} & \secondBestCellColor{0.908} & \secondBestCellColor{0.911} \\
& $\uparrow$ VMAF & 68.91 & 65.83 & \secondBestCellColor{70.84} & \secondBestCellColor{71.45} & \secondBestCellColor{72.11} & \secondBestCellColor{73.00} \\
\midrule
\multirow{4}{*}{TiNeuVox}& $\downarrow$ LPIPS & 0.287 & 0.252 & 0.308 & 0.334 & 0.337 & 0.361 \\
& $\uparrow$ PSNR & 27.60 & 26.61 & 25.01 & 23.86 & 23.03 & 21.56 \\
& $\uparrow$ SSIM & 0.819 & 0.826 & 0.809 & 0.804 & 0.797 & 0.779 \\
& $\uparrow$ VMAF & 59.43 & 59.58 & 47.69 & 38.91 & 32.23 & 22.75 \\
\midrule
\multirow{4}{*}{NDVG}& $\downarrow$ LPIPS & 0.195 & 0.212 & 0.247 & 0.312 & 0.337 & 0.376 \\
& $\uparrow$ PSNR & 29.61 & 26.94 & 25.26 & 22.20 & 20.73 & 18.31 \\
& $\uparrow$ SSIM & 0.875 & 0.860 & 0.834 & 0.794 & 0.778 & 0.739 \\
& $\uparrow$ VMAF & 73.04 & 64.10 & 51.34 & 31.65 & 21.29 & 9.317 \\
\midrule
\multirow{4}{*}{HyperNeRF}& $\downarrow$ LPIPS & 0.228 & 0.222 & 0.236 & 0.256 & 0.294 & 0.321 \\
& $\uparrow$ PSNR & 25.74 & 25.88 & 25.94 & 25.07 & 23.95 & 22.94 \\
& $\uparrow$ SSIM & 0.841 & 0.842 & 0.835 & 0.823 & 0.805 & 0.794 \\
& $\uparrow$ VMAF & \secondBestCellColor{74.35} & \secondBestCellColor{74.72} & 69.83 & 62.03 & 48.92 & 39.10 \\
\midrule
\multirow{4}{*}{NeuralBody}& $\downarrow$ LPIPS & 0.272 & 0.289 & 0.289 & 0.303 & 0.327 & 0.356 \\
& $\uparrow$ PSNR & 27.28 & 25.89 & 27.87 & 25.38 & 25.16 & 26.66 \\
& $\uparrow$ SSIM & 0.822 & 0.814 & 0.810 & 0.809 & 0.798 & 0.763 \\
& $\uparrow$ VMAF & 42.95 & 42.22 & 41.02 & 37.28 & 32.35 & 29.29 \\
\midrule
\multirow{4}{*}{TAVA}& $\downarrow$ LPIPS & 0.218 & 0.236 & 0.260 & 0.313 & 0.362 & 0.411 \\
& $\uparrow$ PSNR & 28.62 & 27.29 & 26.50 & 25.12 & 23.96 & 22.86 \\
& $\uparrow$ SSIM & 0.848 & 0.841 & 0.830 & 0.806 & 0.782 & 0.757 \\
& $\uparrow$ VMAF & 66.45 & 62.64 & 57.15 & 43.75 & 30.22 & 18.56 \\
\bottomrule
\end{tabular}
}
\caption{moderate movements}
\label{tab:frames-vs-quality-s1}
\end{table}

\begin{table}
 \centering
 \resizebox{\linewidth}{!}{

\begin{tabular}{clcccccc}
\multicolumn{1}{l}{Method} & Metric & 20 & 50 & 100 & 250 & 500 & 1000 \\
\midrule
\multirow{4}{*}{Ours}& $\downarrow$ LPIPS & \secondBestCellColor{0.108} & \secondBestCellColor{0.111} & \secondBestCellColor{0.106} & \secondBestCellColor{0.105} & \secondBestCellColor{0.104} & \secondBestCellColor{0.112} \\
& $\uparrow$ PSNR & \secondBestCellColor{29.02} & \bestCellColor{28.98} & \bestCellColor{28.95} & \bestCellColor{28.41} & \bestCellColor{28.53} & \bestCellColor{28.17} \\
& $\uparrow$ SSIM & \bestCellColor{0.903} & \bestCellColor{0.906} & \bestCellColor{0.910} & \bestCellColor{0.912} & \bestCellColor{0.913} & \bestCellColor{0.906} \\
& $\uparrow$ VMAF & \bestCellColor{88.20} & \bestCellColor{89.37} & \bestCellColor{89.59} & \bestCellColor{88.63} & \bestCellColor{88.61} & \bestCellColor{88.33} \\
\midrule
\multirow{4}{*}{Instant-NGP}& $\downarrow$ LPIPS & \bestCellColor{0.090} & \bestCellColor{0.087} & \bestCellColor{0.088} & \bestCellColor{0.086} & \bestCellColor{0.086} & \bestCellColor{0.087} \\
& $\uparrow$ PSNR & \bestCellColor{29.02} & \secondBestCellColor{27.84} & \secondBestCellColor{28.25} & \secondBestCellColor{27.89} & \secondBestCellColor{27.86} & \secondBestCellColor{27.93} \\
& $\uparrow$ SSIM & \secondBestCellColor{0.877} & \secondBestCellColor{0.886} & \secondBestCellColor{0.894} & \secondBestCellColor{0.899} & \secondBestCellColor{0.900} & \secondBestCellColor{0.899} \\
& $\uparrow$ VMAF & \secondBestCellColor{79.38} & \secondBestCellColor{80.64} & \secondBestCellColor{82.55} & \secondBestCellColor{82.10} & \secondBestCellColor{82.44} & \secondBestCellColor{82.20} \\
\midrule
\multirow{4}{*}{TiNeuVox}& $\downarrow$ LPIPS & 0.337 & 0.357 & 0.346 & 0.359 & 0.359 & 0.381 \\
& $\uparrow$ PSNR & 24.55 & 21.66 & 21.22 & 19.78 & 18.85 & 18.03 \\
& $\uparrow$ SSIM & 0.764 & 0.774 & 0.779 & 0.780 & 0.775 & 0.765 \\
& $\uparrow$ VMAF & 53.09 & 35.03 & 32.86 & 23.32 & 17.24 & 13.48 \\
\midrule
\multirow{4}{*}{NDVG}& $\downarrow$ LPIPS & 0.342 & 0.339 & 0.353 & 0.364 & 0.396 & 0.405 \\
& $\uparrow$ PSNR & 23.60 & 20.36 & 19.06 & 17.30 & 15.13 & 14.02 \\
& $\uparrow$ SSIM & 0.771 & 0.763 & 0.753 & 0.736 & 0.704 & 0.693 \\
& $\uparrow$ VMAF & 53.48 & 36.96 & 24.92 & 12.11 & 4.236 & 3.048 \\
\midrule
\multirow{4}{*}{HyperNeRF}& $\downarrow$ LPIPS & 0.272 & 0.248 & 0.266 & 0.283 & 0.310 & 0.328 \\
& $\uparrow$ PSNR & 25.65 & 24.58 & 23.51 & 22.58 & 21.22 & 20.60 \\
& $\uparrow$ SSIM & 0.800 & 0.822 & 0.817 & 0.810 & 0.797 & 0.787 \\
& $\uparrow$ VMAF & 71.81 & 71.38 & 64.51 & 52.98 & 40.77 & 34.92 \\
\midrule
\multirow{4}{*}{NeuralBody}& $\downarrow$ LPIPS & 0.339 & 0.326 & 0.332 & 0.333 & 0.353 & 0.377 \\
& $\uparrow$ PSNR & 26.77 & 24.43 & 25.97 & 23.93 & 23.53 & 24.50 \\
& $\uparrow$ SSIM & 0.791 & 0.800 & 0.800 & 0.801 & 0.789 & 0.770 \\
& $\uparrow$ VMAF & 50.49 & 48.32 & 42.64 & 40.15 & 31.90 & 24.61 \\
\midrule
\multirow{4}{*}{TAVA}& $\downarrow$ LPIPS & 0.322 & 0.318 & 0.330 & 0.376 & 0.415 & 0.447 \\
& $\uparrow$ PSNR & 26.26 & 24.21 & 23.60 & 22.11 & 20.84 & 20.13 \\
& $\uparrow$ SSIM & 0.792 & 0.801 & 0.802 & 0.778 & 0.749 & 0.722 \\
& $\uparrow$ VMAF & 67.40 & 54.67 & 51.41 & 33.06 & 16.68 & 8.129 \\
\bottomrule
\end{tabular}

}
\caption{strong movements}
\label{tab:frames-vs-quality-s2}
\end{table}

\begin{figure*}
    \setlength{\tabcolsep}{1pt}
    \setlength{\mrgone}{0.15\textwidth}
    \centering
   \resizebox{0.8\linewidth}{!}{

    \begin{tabular}{ccccccc}

        NDVG 
         & \includegraphics[width=\mrgone]{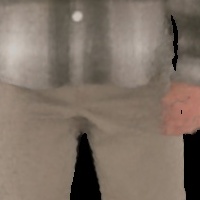}&
        \includegraphics[width=\mrgone]{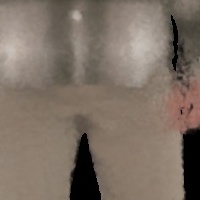}&
        \includegraphics[width=\mrgone]{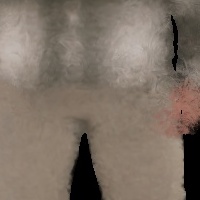}&
        \includegraphics[width=\mrgone]{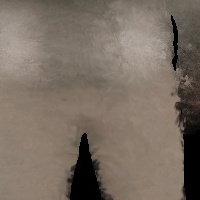}&
        \includegraphics[width=\mrgone]{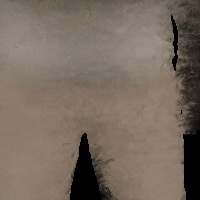}
        \\
                HyperNeRF
         & \includegraphics[width=\mrgone]{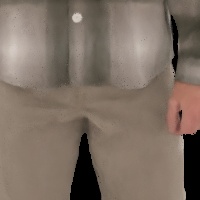}&
        \includegraphics[width=\mrgone]{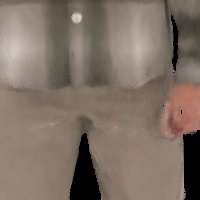}&
        \includegraphics[width=\mrgone]{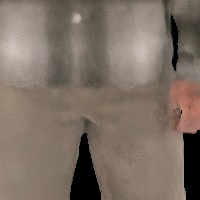}&
        \includegraphics[width=\mrgone]{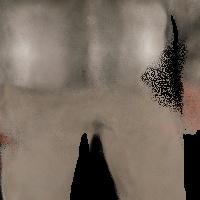}&
        \includegraphics[width=\mrgone]{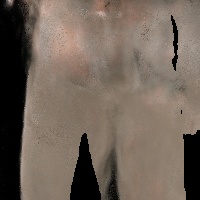}
        \\

        TiNeuVoX
         & \includegraphics[width=\mrgone]{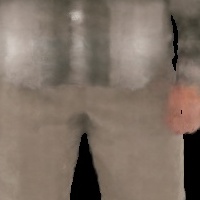}&
        \includegraphics[width=\mrgone]{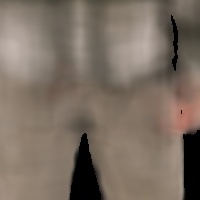}&
        \includegraphics[width=\mrgone]{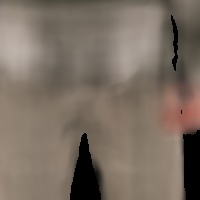}&
        \includegraphics[width=\mrgone]{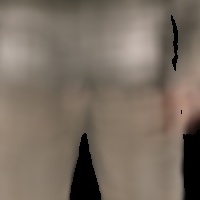}&
        \includegraphics[width=\mrgone]{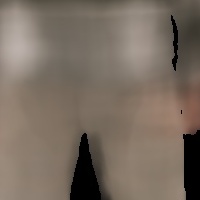}
        \\
        Ours
         & \includegraphics[width=\mrgone]{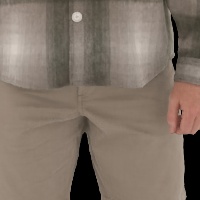}&
        \includegraphics[width=\mrgone]{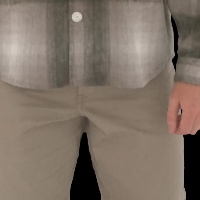}&
        \includegraphics[width=\mrgone]{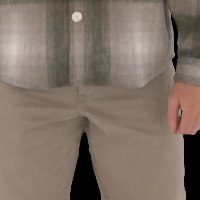}&
        \includegraphics[width=\mrgone]{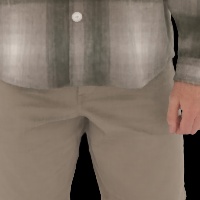}&
        \includegraphics[width=\mrgone]{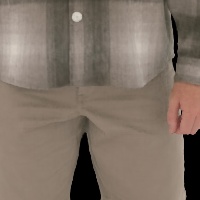}
        \\
        \hline
         \multirow{2}{*}{NDVG} 
         & \includegraphics[width=\mrgone]{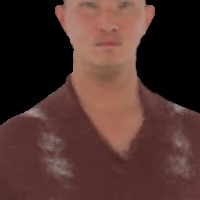}&
        \includegraphics[width=\mrgone]{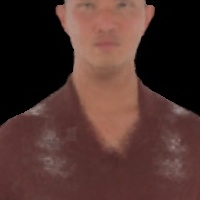}&
        \includegraphics[width=\mrgone]{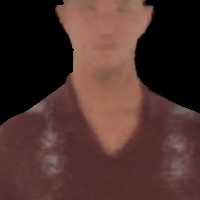}&
        \includegraphics[width=\mrgone]{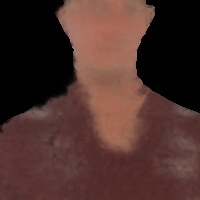}&
        \includegraphics[width=\mrgone]{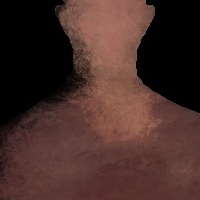}
        \\
                HyperNeRF
         & \includegraphics[width=\mrgone]{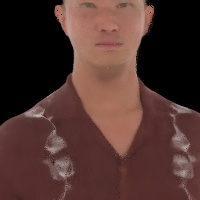}&
        \includegraphics[width=\mrgone]{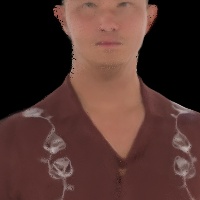}&
        \includegraphics[width=\mrgone]{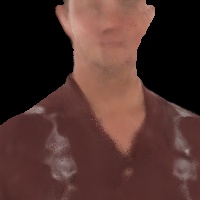}&
        \includegraphics[width=\mrgone]{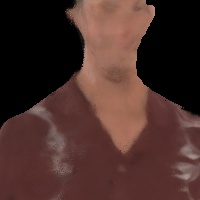}&
        \includegraphics[width=\mrgone]{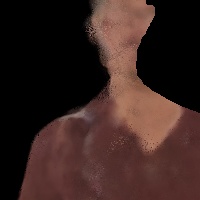}
        \\

        TiNeuVoX
         & \includegraphics[width=\mrgone]{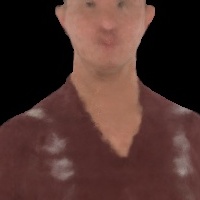}&
        \includegraphics[width=\mrgone]{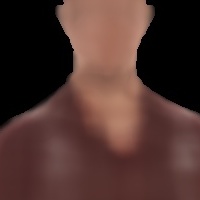}&
        \includegraphics[width=\mrgone]{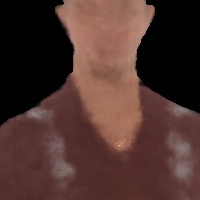}&
        \includegraphics[width=\mrgone]{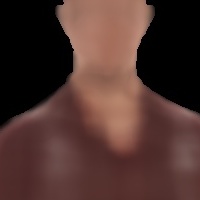}&
        \includegraphics[width=\mrgone]{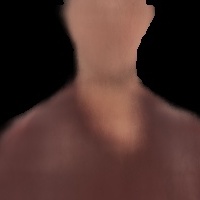}
        \\
        Ours
         & \includegraphics[width=\mrgone]{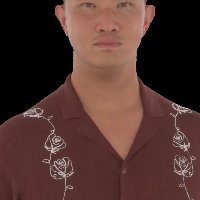}&
        \includegraphics[width=\mrgone]{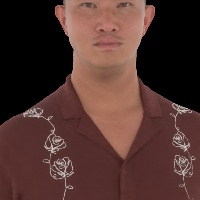}&
        \includegraphics[width=\mrgone]{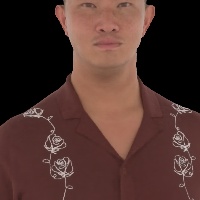}&
        \includegraphics[width=\mrgone]{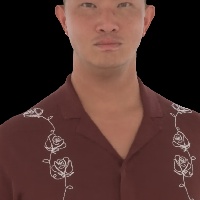}&
        \includegraphics[width=\mrgone]{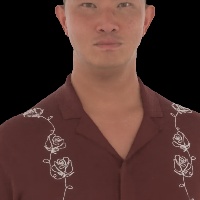}
        \\
        \hline
         Method  & 20 & 50 & 100 & 250 & 500 \\
         
    \end{tabular}
    }
    \caption{\textbf{Impact of increasing sequence length}. For deformation-based baselines synthesis quality drops when rendering the same pose and frame while increasing the sequence length. Our results on the other hand have constant quality independent of the sequence length.
    }
    \label{fig:qualitative-comparison}
\end{figure*}

\clearpage
\newpage

\begin{table}
 \centering
 \resizebox{\linewidth}{!}{
\begin{tabular}{clcccccc}
\multicolumn{1}{l}{Method} & Metric & 20 & 50 & 100 & 250 & 500 & 1000 \\
\midrule
\multirow{4}{*}{Ours}& $\downarrow$ LPIPS & \bestCellColor{0.046} & \bestCellColor{0.052} & \bestCellColor{0.055} & \bestCellColor{0.059} & \bestCellColor{0.060} & \bestCellColor{0.064} \\
& $\uparrow$ PSNR & \bestCellColor{31.51} & \bestCellColor{31.45} & \bestCellColor{30.53} & \bestCellColor{30.42} & \bestCellColor{30.23} & \bestCellColor{30.31} \\
& $\uparrow$ SSIM & \bestCellColor{0.951} & \bestCellColor{0.947} & \bestCellColor{0.945} & \bestCellColor{0.941} & \bestCellColor{0.939} & \bestCellColor{0.934} \\
& $\uparrow$ VMAF & \bestCellColor{83.10} & \bestCellColor{84.43} & \bestCellColor{86.90} & \bestCellColor{85.27} & \bestCellColor{85.47} & \bestCellColor{87.41} \\
\midrule
\multirow{4}{*}{Instant-NGP}& $\downarrow$ LPIPS & \secondBestCellColor{0.062} & \secondBestCellColor{0.062} & \secondBestCellColor{0.069} & \secondBestCellColor{0.071} & \secondBestCellColor{0.072} & \secondBestCellColor{0.073} \\
& $\uparrow$ PSNR & \secondBestCellColor{30.60} & \secondBestCellColor{29.87} & \secondBestCellColor{29.71} & \secondBestCellColor{29.65} & \secondBestCellColor{29.56} & \secondBestCellColor{29.64} \\
& $\uparrow$ SSIM & \secondBestCellColor{0.910} & \secondBestCellColor{0.927} & \secondBestCellColor{0.924} & \secondBestCellColor{0.921} & \secondBestCellColor{0.920} & \secondBestCellColor{0.920} \\
& $\uparrow$ VMAF & 73.27 & 69.36 & 73.84 & \secondBestCellColor{73.95} & \secondBestCellColor{75.41} & \secondBestCellColor{76.38} \\
\midrule
\multirow{4}{*}{TiNeuVox}& $\downarrow$ LPIPS & 0.241 & 0.191 & 0.226 & 0.290 & 0.310 & 0.338 \\
& $\uparrow$ PSNR & 27.78 & 26.55 & 25.36 & 23.61 & 22.36 & 20.50 \\
& $\uparrow$ SSIM & 0.843 & 0.847 & 0.829 & 0.820 & 0.813 & 0.778 \\
& $\uparrow$ VMAF & 65.61 & 64.23 & 59.43 & 43.08 & 31.88 & 21.92 \\
\midrule
\multirow{4}{*}{NDVG}& $\downarrow$ LPIPS & 0.121 & 0.168 & 0.213 & 0.276 & 0.313 & 0.361 \\
& $\uparrow$ PSNR & 30.51 & 26.11 & 24.22 & 21.59 & 19.45 & 17.15 \\
& $\uparrow$ SSIM & 0.904 & 0.867 & 0.841 & 0.803 & 0.786 & 0.736 \\
& $\uparrow$ VMAF & \secondBestCellColor{79.21} & 61.79 & 52.65 & 34.74 & 19.83 & 6.256 \\
\midrule
\multirow{4}{*}{HyperNeRF}& $\downarrow$ LPIPS & 0.186 & 0.178 & 0.206 & 0.222 & 0.269 & 0.301 \\
& $\uparrow$ PSNR & 25.05 & 24.94 & 26.05 & 25.08 & 23.17 & 22.56 \\
& $\uparrow$ SSIM & 0.868 & 0.864 & 0.854 & 0.844 & 0.820 & 0.806 \\
& $\uparrow$ VMAF & 79.00 & \secondBestCellColor{78.89} & \secondBestCellColor{75.35} & 68.98 & 49.06 & 41.72 \\
\midrule
\multirow{4}{*}{NeuralBody}& $\downarrow$ LPIPS & 0.184 & 0.228 & 0.240 & 0.243 & 0.270 & 0.305 \\
& $\uparrow$ PSNR & 28.87 & 26.23 & 27.63 & 25.79 & 25.38 & 26.74 \\
& $\uparrow$ SSIM & 0.864 & 0.844 & 0.837 & 0.837 & 0.827 & 0.786 \\
& $\uparrow$ VMAF & 48.10 & 46.36 & 41.70 & 46.55 & 43.24 & 35.05 \\
\midrule
\multirow{4}{*}{TAVA}& $\downarrow$ LPIPS & 0.149 & 0.178 & 0.210 & 0.259 & 0.308 & 0.373 \\
& $\uparrow$ PSNR & 29.30 & 27.62 & 26.80 & 25.22 & 23.81 & 22.63 \\
& $\uparrow$ SSIM & 0.879 & 0.868 & 0.852 & 0.827 & 0.803 & 0.770 \\
& $\uparrow$ VMAF & 72.10 & 67.40 & 63.20 & 50.00 & 32.49 & 20.46 \\
\bottomrule
\end{tabular}
}
\caption{Actor 1, Sequence 1}
\label{tab:frames-vs-quality-a1}
\end{table}

\begin{table}
 \centering
 \resizebox{\linewidth}{!}{
\begin{tabular}{clcccccc}
\multicolumn{1}{l}{Method} & Metric & 20 & 50 & 100 & 250 & 500 & 1000 \\
\midrule
\multirow{4}{*}{Ours}& $\downarrow$ LPIPS & \bestCellColor{0.081} & \secondBestCellColor{0.086} & \secondBestCellColor{0.080} & \bestCellColor{0.079} & \bestCellColor{0.077} & \bestCellColor{0.079} \\
& $\uparrow$ PSNR & \bestCellColor{30.67} & \bestCellColor{30.21} & \bestCellColor{30.54} & \bestCellColor{29.63} & \bestCellColor{29.81} & \bestCellColor{29.54} \\
& $\uparrow$ SSIM & \bestCellColor{0.946} & \bestCellColor{0.943} & \bestCellColor{0.948} & \bestCellColor{0.946} & \bestCellColor{0.947} & \bestCellColor{0.946} \\
& $\uparrow$ VMAF & \bestCellColor{85.23} & \bestCellColor{84.66} & \bestCellColor{87.98} & \bestCellColor{85.73} & \bestCellColor{85.58} & \bestCellColor{87.73} \\
\midrule
\multirow{4}{*}{Instant-NGP}& $\downarrow$ LPIPS & \secondBestCellColor{0.084} & \bestCellColor{0.083} & \bestCellColor{0.076} & \secondBestCellColor{0.081} & \secondBestCellColor{0.080} & \secondBestCellColor{0.081} \\
& $\uparrow$ PSNR & \secondBestCellColor{29.60} & \secondBestCellColor{26.99} & \secondBestCellColor{27.99} & \secondBestCellColor{27.75} & \secondBestCellColor{27.76} & \secondBestCellColor{27.77} \\
& $\uparrow$ SSIM & \secondBestCellColor{0.914} & \secondBestCellColor{0.897} & \secondBestCellColor{0.914} & \secondBestCellColor{0.912} & \secondBestCellColor{0.915} & \secondBestCellColor{0.914} \\
& $\uparrow$ VMAF & \secondBestCellColor{76.35} & \secondBestCellColor{81.46} & \secondBestCellColor{81.37} & \secondBestCellColor{81.31} & \secondBestCellColor{81.65} & \secondBestCellColor{81.57} \\
\midrule
\multirow{4}{*}{TiNeuVox}& $\downarrow$ LPIPS & 0.306 & 0.337 & 0.277 & 0.321 & 0.314 & 0.337 \\
& $\uparrow$ PSNR & 25.09 & 21.33 & 22.48 & 21.05 & 20.05 & 19.57 \\
& $\uparrow$ SSIM & 0.834 & 0.821 & 0.832 & 0.826 & 0.814 & 0.817 \\
& $\uparrow$ VMAF & 46.73 & 21.32 & 33.18 & 20.47 & 12.83 & 11.13 \\
\midrule
\multirow{4}{*}{NDVG}& $\downarrow$ LPIPS & 0.299 & 0.286 & 0.300 & 0.333 & 0.363 & 0.372 \\
& $\uparrow$ PSNR & 24.66 & 21.80 & 20.58 & 18.82 & 16.50 & 15.41 \\
& $\uparrow$ SSIM & 0.841 & 0.827 & 0.823 & 0.791 & 0.762 & 0.753 \\
& $\uparrow$ VMAF & 43.17 & 34.33 & 23.83 & 7.743 & 0.482 & 0.106 \\
\midrule
\multirow{4}{*}{HyperNeRF}& $\downarrow$ LPIPS & 0.223 & 0.194 & 0.223 & 0.244 & 0.273 & 0.291 \\
& $\uparrow$ PSNR & 25.79 & 25.55 & 24.28 & 23.62 & 22.50 & 21.74 \\
& $\uparrow$ SSIM & 0.849 & 0.876 & 0.865 & 0.851 & 0.842 & 0.833 \\
& $\uparrow$ VMAF & 54.65 & 68.94 & 56.97 & 49.56 & 40.25 & 31.78 \\
\midrule
\multirow{4}{*}{NeuralBody}& $\downarrow$ LPIPS & 0.260 & 0.257 & 0.255 & 0.268 & 0.283 & 0.295 \\
& $\uparrow$ PSNR & 27.55 & 25.48 & 26.89 & 24.94 & 24.58 & 25.76 \\
& $\uparrow$ SSIM & 0.862 & 0.858 & 0.859 & 0.853 & 0.846 & 0.838 \\
& $\uparrow$ VMAF & 46.52 & 46.26 & 39.96 & 43.91 & 37.46 & 31.37 \\
\midrule
\multirow{4}{*}{TAVA}& $\downarrow$ LPIPS & 0.278 & 0.272 & 0.283 & 0.331 & 0.362 & 0.391 \\
& $\uparrow$ PSNR & 26.79 & 25.03 & 24.40 & 23.34 & 22.37 & 21.72 \\
& $\uparrow$ SSIM & 0.860 & 0.857 & 0.855 & 0.830 & 0.814 & 0.794 \\
& $\uparrow$ VMAF & 61.51 & 53.90 & 48.98 & 35.23 & 23.89 & 14.16 \\
\bottomrule
\end{tabular}
}
\caption{Actor 2, Sequence 2}
\label{tab:frames-vs-quality-a2}
\end{table}

\begin{table}
 \centering
 \resizebox{\linewidth}{!}{
\begin{tabular}{clcccccc}
\multicolumn{1}{l}{Method} & Metric & 20 & 50 & 100 & 250 & 500 & 1000 \\
\midrule
\multirow{4}{*}{Ours}& $\downarrow$ LPIPS & \secondBestCellColor{0.120} & \secondBestCellColor{0.138} & \secondBestCellColor{0.135} & \secondBestCellColor{0.151} & \secondBestCellColor{0.155} & \secondBestCellColor{0.160} \\
& $\uparrow$ PSNR & \bestCellColor{31.02} & \bestCellColor{30.26} & \bestCellColor{30.25} & \secondBestCellColor{28.98} & \bestCellColor{29.50} & \secondBestCellColor{29.19} \\
& $\uparrow$ SSIM & \bestCellColor{0.893} & \bestCellColor{0.888} & \bestCellColor{0.896} & \bestCellColor{0.888} & \bestCellColor{0.885} & \secondBestCellColor{0.881} \\
& $\uparrow$ VMAF & \bestCellColor{72.05} & \bestCellColor{71.11} & \bestCellColor{75.40} & \bestCellColor{78.37} & \bestCellColor{76.96} & \bestCellColor{78.22} \\
\midrule
\multirow{4}{*}{Instant-NGP}& $\downarrow$ LPIPS & \bestCellColor{0.119} & \bestCellColor{0.125} & \bestCellColor{0.126} & \bestCellColor{0.128} & \bestCellColor{0.129} & \bestCellColor{0.128} \\
& $\uparrow$ PSNR & \secondBestCellColor{29.44} & \secondBestCellColor{29.22} & \secondBestCellColor{29.28} & \bestCellColor{29.23} & \secondBestCellColor{29.10} & \bestCellColor{29.32} \\
& $\uparrow$ SSIM & \secondBestCellColor{0.858} & \secondBestCellColor{0.881} & \secondBestCellColor{0.883} & \secondBestCellColor{0.881} & \secondBestCellColor{0.880} & \bestCellColor{0.883} \\
& $\uparrow$ VMAF & 62.37 & 60.88 & \secondBestCellColor{67.69} & \secondBestCellColor{68.24} & \secondBestCellColor{68.54} & \secondBestCellColor{70.39} \\
\midrule
\multirow{4}{*}{TiNeuVox}& $\downarrow$ LPIPS & 0.352 & 0.298 & 0.406 & 0.430 & 0.436 & 0.452 \\
& $\uparrow$ PSNR & 27.51 & 26.62 & 24.13 & 22.98 & 22.30 & 21.28 \\
& $\uparrow$ SSIM & 0.782 & 0.791 & 0.760 & 0.752 & 0.751 & 0.747 \\
& $\uparrow$ VMAF & 49.37 & 51.83 & 29.76 & 24.86 & 19.28 & 12.17 \\
\midrule
\multirow{4}{*}{NDVG}& $\downarrow$ LPIPS & 0.240 & 0.281 & 0.354 & 0.435 & 0.453 & 0.481 \\
& $\uparrow$ PSNR & 28.76 & 25.83 & 23.13 & 21.17 & 20.05 & 17.83 \\
& $\uparrow$ SSIM & 0.841 & 0.812 & 0.763 & 0.731 & 0.724 & 0.692 \\
& $\uparrow$ VMAF & 61.99 & 50.27 & 28.79 & 17.01 & 7.948 & 2.447 \\
\midrule
\multirow{4}{*}{HyperNeRF}& $\downarrow$ LPIPS & 0.233 & 0.250 & 0.275 & 0.322 & 0.374 & 0.388 \\
& $\uparrow$ PSNR & 25.75 & 26.53 & 25.96 & 24.85 & 23.29 & 23.04 \\
& $\uparrow$ SSIM & 0.827 & 0.818 & 0.800 & 0.777 & 0.758 & 0.761 \\
& $\uparrow$ VMAF & \secondBestCellColor{71.59} & \secondBestCellColor{69.29} & 58.71 & 49.67 & 33.25 & 33.88 \\
\midrule
\multirow{4}{*}{NeuralBody}& $\downarrow$ LPIPS & 0.288 & 0.333 & 0.354 & 0.368 & 0.396 & 0.429 \\
& $\uparrow$ PSNR & 27.51 & 25.88 & 27.18 & 25.30 & 24.81 & 25.68 \\
& $\uparrow$ SSIM & 0.804 & 0.777 & 0.739 & 0.762 & 0.745 & 0.668 \\
& $\uparrow$ VMAF & 42.89 & 42.13 & 34.00 & 33.25 & 26.65 & 21.11 \\
\midrule
\multirow{4}{*}{TAVA}& $\downarrow$ LPIPS & 0.261 & 0.303 & 0.341 & 0.410 & 0.467 & 0.504 \\
& $\uparrow$ PSNR & 28.47 & 26.93 & 25.83 & 24.28 & 23.13 & 22.21 \\
& $\uparrow$ SSIM & 0.820 & 0.801 & 0.782 & 0.749 & 0.721 & 0.704 \\
& $\uparrow$ VMAF & 60.16 & 55.27 & 46.70 & 29.98 & 15.05 & 6.436 \\
\bottomrule
\end{tabular}
}
\caption{Actor 3, Sequence 1}
\label{tab:frames-vs-quality-a3}
\end{table}

\begin{table}
 \centering
 \resizebox{\linewidth}{!}{
\begin{tabular}{clcccccc}
\multicolumn{1}{l}{Method} & Metric & 20 & 50 & 100 & 250 & 500 & 1000 \\
\midrule
\multirow{4}{*}{Ours}& $\downarrow$ LPIPS & \secondBestCellColor{0.110} & \secondBestCellColor{0.118} & \secondBestCellColor{0.114} & \secondBestCellColor{0.115} & \secondBestCellColor{0.114} & \secondBestCellColor{0.138} \\
& $\uparrow$ PSNR & \bestCellColor{27.52} & \bestCellColor{27.60} & \bestCellColor{27.55} & \bestCellColor{27.28} & \bestCellColor{27.04} & \secondBestCellColor{26.42} \\
& $\uparrow$ SSIM & \bestCellColor{0.851} & \bestCellColor{0.855} & \bestCellColor{0.860} & \bestCellColor{0.866} & \bestCellColor{0.870} & \secondBestCellColor{0.842} \\
& $\uparrow$ VMAF & \bestCellColor{90.93} & \bestCellColor{91.46} & \bestCellColor{90.47} & \bestCellColor{90.80} & \bestCellColor{90.27} & \bestCellColor{89.87} \\
\midrule
\multirow{4}{*}{Instant-NGP}& $\downarrow$ LPIPS & \bestCellColor{0.094} & \bestCellColor{0.087} & \bestCellColor{0.083} & \bestCellColor{0.082} & \bestCellColor{0.081} & \bestCellColor{0.084} \\
& $\uparrow$ PSNR & \secondBestCellColor{26.70} & \secondBestCellColor{25.66} & \secondBestCellColor{26.68} & \secondBestCellColor{26.48} & \secondBestCellColor{26.53} & \bestCellColor{26.53} \\
& $\uparrow$ SSIM & \secondBestCellColor{0.787} & \secondBestCellColor{0.807} & \secondBestCellColor{0.839} & \secondBestCellColor{0.855} & \secondBestCellColor{0.862} & \bestCellColor{0.856} \\
& $\uparrow$ VMAF & \secondBestCellColor{83.50} & \secondBestCellColor{87.28} & \secondBestCellColor{87.09} & \secondBestCellColor{87.17} & \secondBestCellColor{87.39} & \secondBestCellColor{87.59} \\
\midrule
\multirow{4}{*}{TiNeuVox}& $\downarrow$ LPIPS & 0.458 & 0.407 & 0.404 & 0.386 & 0.388 & 0.432 \\
& $\uparrow$ PSNR & 22.62 & 20.78 & 19.39 & 18.62 & 17.43 & 15.81 \\
& $\uparrow$ SSIM & 0.610 & 0.698 & 0.707 & 0.722 & 0.722 & 0.692 \\
& $\uparrow$ VMAF & 51.25 & 39.78 & 30.67 & 29.42 & 25.38 & 15.91 \\
\midrule
\multirow{4}{*}{NDVG}& $\downarrow$ LPIPS & 0.483 & 0.421 & 0.414 & 0.428 & 0.443 & 0.454 \\
& $\uparrow$ PSNR & 20.95 & 18.16 & 16.99 & 15.27 & 13.03 & 12.54 \\
& $\uparrow$ SSIM & 0.608 & 0.662 & 0.669 & 0.662 & 0.632 & 0.625 \\
& $\uparrow$ VMAF & 56.20 & 34.25 & 24.36 & 16.40 & 10.35 & 8.543 \\
\midrule
\multirow{4}{*}{HyperNeRF}& $\downarrow$ LPIPS & 0.394 & 0.297 & 0.309 & 0.322 & 0.335 & 0.366 \\
& $\uparrow$ PSNR & 23.50 & 23.44 & 22.13 & 21.16 & 20.17 & 19.53 \\
& $\uparrow$ SSIM & 0.651 & 0.739 & 0.742 & 0.747 & 0.745 & 0.727 \\
& $\uparrow$ VMAF & 78.24 & 71.99 & 66.48 & 55.13 & 42.85 & 40.66 \\
\midrule
\multirow{4}{*}{NeuralBody}& $\downarrow$ LPIPS & 0.454 & 0.377 & 0.372 & 0.367 & 0.392 & 0.423 \\
& $\uparrow$ PSNR & 25.02 & 23.10 & 24.20 & 22.82 & 22.62 & 22.05 \\
& $\uparrow$ SSIM & 0.639 & 0.722 & 0.729 & 0.740 & 0.723 & 0.703 \\
& $\uparrow$ VMAF & 54.05 & 50.99 & 41.15 & 43.07 & 27.27 & 24.65 \\
\midrule
\multirow{4}{*}{TAVA}& $\downarrow$ LPIPS & 0.431 & 0.359 & 0.375 & 0.416 & 0.452 & 0.500 \\
& $\uparrow$ PSNR & 24.97 & 23.22 & 22.36 & 21.01 & 19.84 & 18.87 \\
& $\uparrow$ SSIM & 0.644 & 0.724 & 0.727 & 0.712 & 0.683 & 0.640 \\
& $\uparrow$ VMAF & 72.20 & 56.97 & 48.24 & 29.85 & 12.28 & 3.305 \\
\bottomrule
\end{tabular}
}
\caption{Actor 4, Sequence 2}
\label{tab:frames-vs-quality-a4}
\end{table}

\begin{table}
 \centering
 \resizebox{\linewidth}{!}{
\begin{tabular}{clcccccc}
\multicolumn{1}{l}{Method} & Metric & 20 & 50 & 100 & 250 & 500 & 1000 \\
\midrule
\multirow{4}{*}{Ours}& $\downarrow$ LPIPS & \bestCellColor{0.067} & \bestCellColor{0.073} & \bestCellColor{0.073} & \bestCellColor{0.075} & \bestCellColor{0.082} & \bestCellColor{0.083} \\
& $\uparrow$ PSNR & \bestCellColor{32.01} & \bestCellColor{31.22} & \bestCellColor{31.04} & \bestCellColor{30.47} & \bestCellColor{30.18} & \bestCellColor{30.29} \\
& $\uparrow$ SSIM & \bestCellColor{0.950} & \bestCellColor{0.946} & \bestCellColor{0.946} & \bestCellColor{0.944} & \bestCellColor{0.939} & \bestCellColor{0.937} \\
& $\uparrow$ VMAF & \bestCellColor{78.99} & \bestCellColor{77.68} & \bestCellColor{79.22} & \bestCellColor{78.70} & \bestCellColor{79.40} & \bestCellColor{81.50} \\
\midrule
\multirow{4}{*}{Instant-NGP}& $\downarrow$ LPIPS & \secondBestCellColor{0.098} & \secondBestCellColor{0.090} & \secondBestCellColor{0.095} & \secondBestCellColor{0.091} & \secondBestCellColor{0.092} & \secondBestCellColor{0.090} \\
& $\uparrow$ PSNR & 29.47 & \secondBestCellColor{29.40} & \secondBestCellColor{29.55} & \secondBestCellColor{29.58} & \secondBestCellColor{29.62} & \secondBestCellColor{29.79} \\
& $\uparrow$ SSIM & \secondBestCellColor{0.903} & \secondBestCellColor{0.924} & \secondBestCellColor{0.919} & \secondBestCellColor{0.923} & \secondBestCellColor{0.921} & \secondBestCellColor{0.923} \\
& $\uparrow$ VMAF & 66.84 & 63.65 & 67.78 & \secondBestCellColor{69.37} & \secondBestCellColor{70.21} & \secondBestCellColor{71.07} \\
\midrule
\multirow{4}{*}{TiNeuVox}& $\downarrow$ LPIPS & 0.197 & 0.272 & 0.284 & 0.293 & 0.312 & 0.296 \\
& $\uparrow$ PSNR & 28.20 & 26.43 & 25.38 & 24.35 & 23.38 & 22.11 \\
& $\uparrow$ SSIM & 0.852 & 0.837 & 0.828 & 0.826 & 0.816 & 0.791 \\
& $\uparrow$ VMAF & 64.80 & 57.06 & 51.81 & 45.29 & 37.90 & 29.19 \\
\midrule
\multirow{4}{*}{NDVG}& $\downarrow$ LPIPS & 0.176 & 0.175 & 0.187 & 0.254 & 0.278 & 0.329 \\
& $\uparrow$ PSNR & \secondBestCellColor{29.80} & 27.89 & 27.41 & 23.50 & 22.03 & 19.14 \\
& $\uparrow$ SSIM & 0.894 & 0.892 & 0.878 & 0.829 & 0.806 & 0.762 \\
& $\uparrow$ VMAF & \secondBestCellColor{72.32} & 69.70 & 64.46 & 39.65 & 31.08 & 15.88 \\
\midrule
\multirow{4}{*}{HyperNeRF}& $\downarrow$ LPIPS & 0.237 & 0.234 & 0.223 & 0.223 & 0.253 & 0.278 \\
& $\uparrow$ PSNR & 25.29 & 25.14 & 25.74 & 25.40 & 24.87 & 23.97 \\
& $\uparrow$ SSIM & 0.839 & 0.838 & 0.845 & 0.846 & 0.828 & 0.816 \\
& $\uparrow$ VMAF & 69.36 & \secondBestCellColor{70.11} & \secondBestCellColor{69.02} & 66.22 & 57.84 & 47.33 \\
\midrule
\multirow{4}{*}{NeuralBody}& $\downarrow$ LPIPS & 0.289 & 0.283 & 0.270 & 0.291 & 0.316 & 0.336 \\
& $\uparrow$ PSNR & 26.91 & 26.10 & 28.18 & 25.27 & 24.77 & 26.77 \\
& $\uparrow$ SSIM & 0.826 & 0.823 & 0.836 & 0.820 & 0.811 & 0.802 \\
& $\uparrow$ VMAF & 39.81 & 37.83 & 40.60 & 36.36 & 25.49 & 29.24 \\
\midrule
\multirow{4}{*}{TAVA}& $\downarrow$ LPIPS & 0.208 & 0.213 & 0.227 & 0.269 & 0.324 & 0.373 \\
& $\uparrow$ PSNR & 28.13 & 27.03 & 26.83 & 25.76 & 24.33 & 23.15 \\
& $\uparrow$ SSIM & 0.857 & 0.852 & 0.849 & 0.835 & 0.808 & 0.782 \\
& $\uparrow$ VMAF & 64.42 & 63.31 & 59.15 & 49.24 & 36.51 & 24.45 \\
\bottomrule
\end{tabular}
}
\caption{Actor 5, Sequence 1}
\label{tab:frames-vs-quality-a5}
\end{table}

\begin{table}
 \centering
 \resizebox{\linewidth}{!}{
\begin{tabular}{clcccccc}
\multicolumn{1}{l}{Method} & Metric & 20 & 50 & 100 & 250 & 500 & 1000 \\
\midrule
\multirow{4}{*}{Ours}& $\downarrow$ LPIPS & \secondBestCellColor{0.133} & \secondBestCellColor{0.128} & \secondBestCellColor{0.123} & \secondBestCellColor{0.118} & \secondBestCellColor{0.117} & \secondBestCellColor{0.121} \\
& $\uparrow$ PSNR & \secondBestCellColor{27.26} & \secondBestCellColor{28.10} & \secondBestCellColor{27.80} & \bestCellColor{27.68} & \bestCellColor{27.69} & \secondBestCellColor{27.33} \\
& $\uparrow$ SSIM & \secondBestCellColor{0.899} & \secondBestCellColor{0.908} & \bestCellColor{0.913} & \bestCellColor{0.916} & \bestCellColor{0.915} & \bestCellColor{0.913} \\
& $\uparrow$ VMAF & \bestCellColor{85.85} & \bestCellColor{89.94} & \bestCellColor{88.05} & \bestCellColor{88.09} & \bestCellColor{88.18} & \bestCellColor{86.53} \\
\midrule
\multirow{4}{*}{Instant-NGP}& $\downarrow$ LPIPS & \bestCellColor{0.079} & \bestCellColor{0.090} & \bestCellColor{0.099} & \bestCellColor{0.093} & \bestCellColor{0.094} & \bestCellColor{0.096} \\
& $\uparrow$ PSNR & \bestCellColor{28.18} & \bestCellColor{28.38} & \bestCellColor{28.13} & \secondBestCellColor{27.57} & \secondBestCellColor{27.40} & \bestCellColor{27.53} \\
& $\uparrow$ SSIM & \bestCellColor{0.907} & \bestCellColor{0.914} & \secondBestCellColor{0.906} & \secondBestCellColor{0.908} & \secondBestCellColor{0.905} & \secondBestCellColor{0.904} \\
& $\uparrow$ VMAF & \secondBestCellColor{76.56} & \secondBestCellColor{76.17} & \secondBestCellColor{79.68} & \secondBestCellColor{79.01} & \secondBestCellColor{79.44} & \secondBestCellColor{79.31} \\
\midrule
\multirow{4}{*}{TiNeuVox}& $\downarrow$ LPIPS & 0.258 & 0.296 & 0.311 & 0.351 & 0.356 & 0.367 \\
& $\uparrow$ PSNR & 24.68 & 22.13 & 21.62 & 19.57 & 18.86 & 18.44 \\
& $\uparrow$ SSIM & 0.834 & 0.799 & 0.802 & 0.803 & 0.799 & 0.791 \\
& $\uparrow$ VMAF & 55.69 & 44.06 & 39.54 & 23.80 & 14.92 & 15.44 \\
\midrule
\multirow{4}{*}{NDVG}& $\downarrow$ LPIPS & 0.244 & 0.307 & 0.344 & 0.343 & 0.413 & 0.415 \\
& $\uparrow$ PSNR & 23.79 & 20.35 & 19.17 & 17.70 & 14.90 & 13.71 \\
& $\uparrow$ SSIM & 0.839 & 0.795 & 0.778 & 0.764 & 0.707 & 0.697 \\
& $\uparrow$ VMAF & 57.39 & 38.60 & 26.81 & 12.92 & 2.790 & 3.058 \\
\midrule
\multirow{4}{*}{HyperNeRF}& $\downarrow$ LPIPS & 0.197 & 0.243 & 0.262 & 0.276 & 0.321 & 0.327 \\
& $\uparrow$ PSNR & 25.45 & 23.61 & 23.01 & 22.32 & 20.52 & 20.54 \\
& $\uparrow$ SSIM & 0.866 & 0.843 & 0.839 & 0.835 & 0.811 & 0.807 \\
& $\uparrow$ VMAF & 73.33 & 73.32 & 68.97 & 55.20 & 39.83 & 36.24 \\
\midrule
\multirow{4}{*}{NeuralBody}& $\downarrow$ LPIPS & 0.278 & 0.330 & 0.342 & 0.333 & 0.366 & 0.403 \\
& $\uparrow$ PSNR & 25.18 & 23.35 & 25.81 & 23.58 & 23.20 & 25.21 \\
& $\uparrow$ SSIM & 0.847 & 0.807 & 0.814 & 0.820 & 0.801 & 0.770 \\
& $\uparrow$ VMAF & 51.62 & 46.30 & 46.14 & 38.76 & 32.29 & 19.65 \\
\midrule
\multirow{4}{*}{TAVA}& $\downarrow$ LPIPS & 0.258 & 0.330 & 0.325 & 0.374 & 0.436 & 0.460 \\
& $\uparrow$ PSNR & 25.27 & 23.35 & 23.35 & 21.98 & 20.19 & 19.86 \\
& $\uparrow$ SSIM & 0.849 & 0.807 & 0.824 & 0.797 & 0.751 & 0.731 \\
& $\uparrow$ VMAF & 65.48 & 46.30 & 56.80 & 36.29 & 12.33 & 6.512 \\
\bottomrule
\end{tabular}
}
\caption{Actor 6, Sequence 2}
\label{tab:frames-vs-quality-a6}
\end{table}

\begin{table}
 \centering
 \resizebox{\linewidth}{!}{
\begin{tabular}{clcccccc}
\multicolumn{1}{l}{Method} & Metric & 20 & 50 & 100 & 250 & 500 & 1000 \\
\midrule
\multirow{4}{*}{Ours}& $\downarrow$ LPIPS & \bestCellColor{0.093} & \bestCellColor{0.096} & \bestCellColor{0.089} & \bestCellColor{0.096} & \bestCellColor{0.102} & \bestCellColor{0.103} \\
& $\uparrow$ PSNR & \bestCellColor{31.76} & \bestCellColor{31.56} & \bestCellColor{31.04} & \bestCellColor{30.55} & \bestCellColor{30.69} & \secondBestCellColor{29.95} \\
& $\uparrow$ SSIM & \bestCellColor{0.940} & \bestCellColor{0.938} & \bestCellColor{0.943} & \bestCellColor{0.937} & \bestCellColor{0.935} & \bestCellColor{0.929} \\
& $\uparrow$ VMAF & \bestCellColor{82.38} & \bestCellColor{84.74} & \bestCellColor{85.12} & \bestCellColor{85.40} & \bestCellColor{86.37} & \bestCellColor{85.45} \\
\midrule
\multirow{4}{*}{Instant-NGP}& $\downarrow$ LPIPS & \secondBestCellColor{0.122} & \secondBestCellColor{0.108} & \secondBestCellColor{0.107} & \secondBestCellColor{0.108} & \secondBestCellColor{0.109} & \secondBestCellColor{0.107} \\
& $\uparrow$ PSNR & \secondBestCellColor{30.03} & \secondBestCellColor{30.38} & \secondBestCellColor{30.13} & \secondBestCellColor{30.14} & \secondBestCellColor{30.13} & \bestCellColor{30.31} \\
& $\uparrow$ SSIM & \secondBestCellColor{0.867} & \secondBestCellColor{0.909} & \secondBestCellColor{0.913} & \secondBestCellColor{0.915} & \secondBestCellColor{0.914} & \secondBestCellColor{0.916} \\
& $\uparrow$ VMAF & 73.15 & 69.42 & 74.06 & \secondBestCellColor{74.24} & \secondBestCellColor{74.28} & \secondBestCellColor{74.17} \\
\midrule
\multirow{4}{*}{TiNeuVox}& $\downarrow$ LPIPS & 0.358 & 0.246 & 0.314 & 0.321 & 0.289 & 0.356 \\
& $\uparrow$ PSNR & 26.89 & 26.84 & 25.16 & 24.48 & 24.07 & 22.34 \\
& $\uparrow$ SSIM & 0.800 & 0.830 & 0.819 & 0.818 & 0.808 & 0.799 \\
& $\uparrow$ VMAF & 57.94 & 65.20 & 49.75 & 42.40 & 39.87 & 27.73 \\
\midrule
\multirow{4}{*}{NDVG}& $\downarrow$ LPIPS & 0.243 & 0.222 & 0.232 & 0.282 & 0.304 & 0.334 \\
& $\uparrow$ PSNR & 29.37 & 27.94 & 26.29 & 22.52 & 21.38 & 19.11 \\
& $\uparrow$ SSIM & 0.861 & 0.868 & 0.854 & 0.811 & 0.797 & 0.766 \\
& $\uparrow$ VMAF & \secondBestCellColor{78.65} & 74.62 & 59.49 & 35.19 & 26.32 & 12.68 \\
\midrule
\multirow{4}{*}{HyperNeRF}& $\downarrow$ LPIPS & 0.255 & 0.228 & 0.240 & 0.259 & 0.280 & 0.319 \\
& $\uparrow$ PSNR & 26.88 & 26.92 & 26.00 & 24.94 & 24.47 & 22.17 \\
& $\uparrow$ SSIM & 0.828 & 0.851 & 0.841 & 0.827 & 0.812 & 0.792 \\
& $\uparrow$ VMAF & 77.46 & \secondBestCellColor{80.60} & \secondBestCellColor{76.25} & 63.25 & 55.51 & 33.48 \\
\midrule
\multirow{4}{*}{NeuralBody}& $\downarrow$ LPIPS & 0.328 & 0.312 & 0.294 & 0.309 & 0.325 & 0.355 \\
& $\uparrow$ PSNR & 25.84 & 25.37 & 28.50 & 25.17 & 25.69 & 27.44 \\
& $\uparrow$ SSIM & 0.793 & 0.811 & 0.829 & 0.818 & 0.808 & 0.796 \\
& $\uparrow$ VMAF & 41.02 & 42.56 & 47.78 & 32.95 & 34.04 & 31.75 \\
\midrule
\multirow{4}{*}{TAVA}& $\downarrow$ LPIPS & 0.253 & 0.252 & 0.263 & 0.314 & 0.349 & 0.392 \\
& $\uparrow$ PSNR & 28.60 & 27.57 & 26.55 & 25.22 & 24.57 & 23.46 \\
& $\uparrow$ SSIM & 0.835 & 0.843 & 0.837 & 0.814 & 0.796 & 0.773 \\
& $\uparrow$ VMAF & 69.11 & 64.59 & 59.55 & 45.76 & 36.82 & 22.89 \\
\bottomrule
\end{tabular}
}
\caption{Actor 7, Sequence 1}
\label{tab:frames-vs-quality-a7}
\end{table}

\begin{table}
 \centering
 \resizebox{\linewidth}{!}{
\begin{tabular}{clcccccc}
\multicolumn{1}{l}{Method} & Metric & 20 & 50 & 100 & 250 & 500 & 1000 \\
\midrule
\multirow{4}{*}{Ours}& $\downarrow$ LPIPS & \secondBestCellColor{0.107} & \secondBestCellColor{0.112} & \secondBestCellColor{0.108} & \secondBestCellColor{0.109} & \secondBestCellColor{0.109} & \secondBestCellColor{0.108} \\
& $\uparrow$ PSNR & \secondBestCellColor{30.62} & \secondBestCellColor{30.02} & \secondBestCellColor{29.94} & \secondBestCellColor{29.04} & \secondBestCellColor{29.59} & \secondBestCellColor{29.40} \\
& $\uparrow$ SSIM & \bestCellColor{0.917} & \secondBestCellColor{0.917} & \bestCellColor{0.920} & \bestCellColor{0.920} & \bestCellColor{0.921} & \secondBestCellColor{0.921} \\
& $\uparrow$ VMAF & \bestCellColor{90.80} & \bestCellColor{91.42} & \bestCellColor{91.87} & \bestCellColor{89.89} & \bestCellColor{90.39} & \bestCellColor{89.18} \\
\midrule
\multirow{4}{*}{Instant-NGP}& $\downarrow$ LPIPS & \bestCellColor{0.102} & \bestCellColor{0.088} & \bestCellColor{0.093} & \bestCellColor{0.089} & \bestCellColor{0.088} & \bestCellColor{0.087} \\
& $\uparrow$ PSNR & \bestCellColor{31.60} & \bestCellColor{30.31} & \bestCellColor{30.20} & \bestCellColor{29.79} & \bestCellColor{29.74} & \bestCellColor{29.90} \\
& $\uparrow$ SSIM & \secondBestCellColor{0.900} & \bestCellColor{0.925} & \secondBestCellColor{0.917} & \secondBestCellColor{0.920} & \secondBestCellColor{0.920} & \bestCellColor{0.922} \\
& $\uparrow$ VMAF & \secondBestCellColor{81.11} & \secondBestCellColor{77.64} & \secondBestCellColor{82.05} & \secondBestCellColor{80.89} & \secondBestCellColor{81.28} & \secondBestCellColor{80.35} \\
\midrule
\multirow{4}{*}{TiNeuVox}& $\downarrow$ LPIPS & 0.326 & 0.389 & 0.392 & 0.379 & 0.379 & 0.387 \\
& $\uparrow$ PSNR & 25.80 & 22.43 & 21.39 & 19.88 & 19.06 & 18.32 \\
& $\uparrow$ SSIM & 0.780 & 0.778 & 0.773 & 0.770 & 0.768 & 0.761 \\
& $\uparrow$ VMAF & 58.70 & 34.98 & 28.05 & 19.57 & 15.84 & 11.42 \\
\midrule
\multirow{4}{*}{NDVG}& $\downarrow$ LPIPS & 0.342 & 0.342 & 0.354 & 0.354 & 0.367 & 0.380 \\
& $\uparrow$ PSNR & 24.98 & 21.14 & 19.48 & 17.40 & 16.10 & 14.42 \\
& $\uparrow$ SSIM & 0.797 & 0.766 & 0.740 & 0.725 & 0.716 & 0.698 \\
& $\uparrow$ VMAF & 57.17 & 40.64 & 24.69 & 11.41 & 3.313 & 0.487 \\
\midrule
\multirow{4}{*}{HyperNeRF}& $\downarrow$ LPIPS & 0.275 & 0.259 & 0.271 & 0.288 & 0.312 & 0.328 \\
& $\uparrow$ PSNR & 27.88 & 25.73 & 24.63 & 23.21 & 21.67 & 20.57 \\
& $\uparrow$ SSIM & 0.833 & 0.831 & 0.821 & 0.807 & 0.790 & 0.782 \\
& $\uparrow$ VMAF & 81.03 & 71.27 & 65.61 & 52.05 & 40.14 & 30.99 \\
\midrule
\multirow{4}{*}{NeuralBody}& $\downarrow$ LPIPS & 0.363 & 0.342 & 0.358 & 0.364 & 0.372 & 0.388 \\
& $\uparrow$ PSNR & 29.31 & 25.81 & 26.99 & 24.38 & 23.71 & 24.99 \\
& $\uparrow$ SSIM & 0.817 & 0.813 & 0.801 & 0.792 & 0.787 & 0.771 \\
& $\uparrow$ VMAF & 49.76 & 49.72 & 43.29 & 34.87 & 30.58 & 22.76 \\
\midrule
\multirow{4}{*}{TAVA}& $\downarrow$ LPIPS & 0.321 & 0.311 & 0.338 & 0.383 & 0.409 & 0.438 \\
& $\uparrow$ PSNR & 27.99 & 25.23 & 24.30 & 22.10 & 20.95 & 20.07 \\
& $\uparrow$ SSIM & 0.815 & 0.816 & 0.803 & 0.771 & 0.747 & 0.722 \\
& $\uparrow$ VMAF & 70.40 & 61.50 & 51.62 & 30.88 & 18.23 & 8.532 \\
\bottomrule
\end{tabular}
}
\caption{Actor 8, Sequence 2}
\label{tab:frames-vs-quality-a8}
\end{table}

\clearpage
\newpage

\bibliographystyle{ACM-Reference-Format}